\documentclass[acmlarge]{acmart}

\usepackage{times}
\usepackage{soul}
\usepackage{url}
\usepackage{rotating}
\usepackage{algorithm}
\usepackage{algorithmic}
\usepackage{multirow}
\usepackage{xcolor}
\usepackage{tikz}
\usepackage{tabularx}
\usepackage[edges]{forest}
\usepackage{rotating}
\usepackage{natbib}
\usepackage{dirtytalk}
\usepackage{makecell}
\usepackage[switch]{lineno}
\usepackage[table]{xcolor}

\usepackage[utf8]{inputenc}

\usepackage[table]{xcolor}

\usetikzlibrary{arrows.meta,positioning}
\definecolor{hiddendraw}{HTML}{000000}    
\definecolor{mygreen}{RGB}{0,150,0}       

\tikzstyle{leaf}=[draw=hiddendraw,
    rounded corners,minimum height=1em,
    fill=mygreen!40,text opacity=1, align=center,
    fill opacity=.5,  text=black,align=left,font=\scriptsize,
    inner xsep=3pt,
    inner ysep=1pt,
    ]

\tikzstyle{middle}=[draw=hiddendraw,
    rounded corners,minimum height=1em,
    text opacity=1, align=center,
    fill opacity=.5,  text=black,align=left,font=\scriptsize,
    inner xsep=3pt,
    inner ysep=1pt,
    ]

\usepackage{wrapfig}
\usepackage{graphicx}

\usetikzlibrary{arrows.meta,positioning}
\definecolor{hiddendraw}{HTML}{000000}    
\definecolor{mygreen}{RGB}{0,150,0}       

\AtBeginDocument{%
  }

\setcopyright{acmlicensed}
\copyrightyear{2018}
\acmYear{2018}
\acmDOI{XXXXXXX.XXXXXXX}

\acmJournal{POMACS}
\acmVolume{37}
\acmNumber{4}
\acmArticle{111}
\acmMonth{8}

\begin{document}

\title{Positioning the Role of Mixture-of-Expert in Multimodal Learning: A Survey}

\author{Liangwei Zheng}
\email{liangwei.zheng@adelaide.edu.au}
\affiliation{%
  \institution{Adelaide University}
  \city{Adelaide}
  \state{South Australia}
  \country{Australia}
}

\author{Wenhao Liang}
\affiliation{%
  \institution{Adelaide University}
  \city{Adelaide}
  \state{South Australia}
  \country{Australia}
}
\email{wenhao.liang@adelaide.edu.au}

\author{Wei Emma Zhang}
\affiliation{%
  \institution{Adelaide University}
  \city{Adelaide}
  \state{South Australia}
  \country{Australia}
}

\author{Lin Yue}
\affiliation{%
  \institution{Adelaide University}
  \city{Adelaide}
  \state{South Australia}
  \country{Australia}
}

\author{Weitong Chen}
\affiliation{%
  \institution{Adelaide University}
  \city{Adelaide}
  \state{South Australia}
  \country{Australia}
}

\renewcommand{\shortauthors}{Trovato et al.}

\begin{abstract}

Multimodal learning faces heterogeneous, incomplete, and evolving inputs. This survey examines how Mixture-of-Experts (MoE) addresses these challenges through three perspectives: efficient multimodal computation, expert specialization for representation and cross-modal interaction, and adaptation to imperfect and evolving modalities. We review sparse backbones, upcycling, adaptive depth, and lightweight experts; analyze specialization in representation, alignment, and fusion; and examine robustness to missing, imbalanced, and shifting modalities. Across healthcare, Earth observation, and autonomous systems, we synthesize design considerations and trade-offs in specialization, sharing, efficiency, and robustness, highlighting open challenges in interpretable routing, temporal modality relevance, expert capacity management, and cross-expert coordination.

\end{abstract}



\received{20 February 2007}
\received[revised]{12 March 2009}
\received[accepted]{5 June 2009}

\maketitle

\section{Introduction}

Multimodal learning enables models to reason over information carried by different data sources, including images, text, audio, video, physiological signals, and structured records \cite{li2025uni,yuan2025survey,zheng2025understanding,zheng2024devil,tan2025paths}. Its central difficulty is that these modalities differ not only in representation, but also in granularity, reliability, and the kinds of evidence they provide. A model may need to align representations while preserving modality-specific information, capture useful cross-modal interactions, and make predictions when some sources are missing or degraded. These challenges arise across settings as varied as medical vision-language understanding, remote sensing, industrial diagnosis, and audio--visual recognition \citep{jiang2024med,lin2025remote,lin2026multisensorfault,kim2025mohave,tan2025hydrarag}.

Mixture-of-Experts (MoE) \cite{fedus2022switch} offers a flexible way to address this heterogeneity challenge. Rather than processing every input through an identical set of transformations, an MoE model distributes computation across experts and use a router to select or combine them according to the underlying structure of data \cite{zheng2025rethinking,emnlp/YuQJSML24}. In multimodal systems, the distributed computation design of MoE allow the model to learn modality-specific features\cite{nips/MustafaRPJH22, li2025uni}, integrate cross-modal interactions\cite{xin2025i2moe}, specialize on task or domain knowledge\cite{taskadaptive2026vlm}, and employ lightweight adaptations \cite{huai2025clmoe} to a pretrained model, which enables MoE to serve several distinct roles: (1) Increase model capacity through sparse activation, allocate computation path accordingly. (2) Preserve specialized and aligned representations cross modality heterogeneity structure. (3) Adapt dynamic and evolving changes in multimodal data. This advantages empower MoE model the potentials to address multimodal challenges that the dense models are struggling with.

In this survey, We first review the architecture of fundamental design of MoE architecture and provide comprehensive insights of each architecture. Next, we conduct comprehensive review on recent advancement of MoE for multimodal learning and answer the central research question: \textit{What is the role of MoE in multimodal learning?} by categorize MoE into three main stream (1) Efficient Mixture-of-Expert for Multimodal Model. These methods design multiple efficient architecture including sparse activation, upcycling experts and depth skipping MoE. (2) Expert Specialization across Representation, Modalities and Interactions. This section answer how the distributed architecture of MoE maintain the modality-specialization knowledge and their interactions in representation learning. (3) Adaptive MoE for Imperfect and Evolving Multimodal Data. This section discussed multiple fundamental challenges of multimodal learning that the modality structure, distribution and combination are generally imperfect and evolving over time in real-world observation. Then, we review the process on how MoE can accommodate these problems by asking how MoE architecture can adapt such changes. Figure~\ref{fig:taxonomy_of_MMoE} summarizes the taxonomy of multimodal MoE methods reviewed across these three technical perspectives. In addition, we further discuss the successful and potential applications of using MoE across various domain, including health care, environment science, industrial and autonomous systems, human-center intelligence and other emerging domain.


\textbf{Contributions}: This survey makes three primary contributions: (1) We provide a comprehensive review of Mixture-of-Experts (MoE) approaches for multimodal learning, covering recent methodological advances and applications across healthcare, remote sensing, industrial and autonomous systems, and other emerging domains. We examine how MoE addresses multimodal challenges beyond computational efficiency, including heterogeneous representations, cross-modal interactions, and imperfect data. (2) We develop a fine-grained taxonomy that characterizes the roles of MoE in multimodal learning from three perspectives: efficient computation, expert specialization for representation learning and cross-modal interaction, and adaptation to imperfect and evolving modalities. Our taxonomy emphasizes how expert design and routing mechanisms address specific multimodal learning challenges. (3) We establish connections among existing methods and analyze their impact on multimodal learning. We synthesize key design considerations regarding expert placement, specialization, knowledge sharing, and adaptive routing, while highlighting open challenges in interpretable routing, temporal modality relevance, expert capacity management, and cross-expert coordination.

\textbf{Paper Collection}: We collected research papers from major conferences, including but not limited to ICML, ICLR, NeurIPS, AAAI, IJCAI, CVPR, ICCV, ECCV, ACL, EMNLP, EACL, NAACL, KDD, WWW, ACM MM, MICCAI, and ICASSP, as well as journals such as TPAMI, TIP, TMI, TMM, JMLR, and TMLR. Our search employed combinations of keywords, including "Mixture-of-Experts", "Multimodal MoE", "Multimodal Learning", "Expert Specialization", "Multimodal Fusion", "Missing Modalities", and "Continual Learning". In total, we compiled 231 references, primarily covering studies published between 2022 and 2026, while including earlier foundational works. The literature search was last updated in October 2026.

\section{Recent Surveys}
\label{sec:related_surveys}

Recent surveys examine mixture-of-experts (MoE) and multimodal learning from complementary perspectives. Broad multimodal surveys cover representation, alignment, and fusion \citep{yuan2025survey}. The most closely related work, Xu et al. \citep{xu2026multimodalmoe}, frames multimodal MoE as a transition from sparse scaling toward conditional specialization, organizing its analysis around expert definition, routing granularity, sharing, insertion topology, and system realization. While this view explains how multimodal MoE models are constructed, our survey emphasizes how expert design and routing address specific multimodal learning challenges. Other surveys focus on particular domains or objectives. Cui et al. \citep{cui2026remotesensing} review MoE for remote-sensing perception, multisensor fusion, and temporal analysis. Xu et al. \citep{xu2024integration} examine MoE and multimodal generative AI in the Internet of Vehicles, emphasizing distributed execution, resource constraints, and cooperative intelligence. Our survey complements these domain-specific accounts through comparisons across healthcare, Earth observation, industrial and autonomous systems, human-centered intelligence, and information systems. Jin et al. \citep{jin2025efficient} consider MoE within a broader toolkit for efficient multimodal large language models, while Li and Tang \citep{li2026alignmentfusion} review multimodal alignment and fusion, treating MoE as one integration approach.

As shown in Table \ref{tab:survey_comparison}, existing surveys examine multimodal learning and mixture-of-experts (MoE) from complementary perspectives. Multimodal learning surveys typically emphasize representation, alignment, and fusion \citep{yuan2025survey,li2026alignmentfusion}, while other reviews focus on MoE architectures and systems, computational efficiency, or domain-specific applications. Among surveys focused specifically on multimodal MoE, Xu et al. \citep{xu2026multimodalmoe} emphasize conditional computation, expert design, routing, and system realization, including how modality heterogeneity affects routing and efficiency. Our survey adopts a learning-centric view, examining how MoE is positioned within multimodal pipelines to support computational efficiency, representation learning and cross-modal interaction, and adaptation to imperfect or evolving data. We connect these roles to architectural form, expert specialization, routing granularity, and data conditions, providing a framework for comparing both the multimodal challenges addressed and the expert designs and routing mechanisms used to address them.

\begin{table*}[ht]
\centering
\caption{Comparison with recent surveys related to multimodal learning and Mixture-of-Experts. A \checkmark indicates a major focus, while $\times$ indicates that the topic is not a primary focus.}
\label{tab:survey_comparison}
\scriptsize
\setlength{\tabcolsep}{3pt}
\resizebox{\textwidth}{!}{
\begin{tabular}{ccccccc}
\toprule
\textbf{Survey} & \textbf{Primary Perspective} &  \makecell[c]{\textbf{Multimodal} \\\textbf{Efficiency}} & \makecell[c]{\textbf{Representation}\\\textbf{\& Alignment}} & \makecell[c]{\textbf{Modality / Interaction}\\\textbf{Specialization}} & \makecell[c]{\textbf{Imperfect / Evolving}\\\textbf{Modalities}} & \makecell[c]{\textbf{Cross-domain}\\\textbf{Applications}} \\
\midrule
Xu et al. (2024) & MoE and multimodal generative AI for IoV & \checkmark & $\times$ & $\times$ & $\times$ & $\times$ \\
Jin et al. (2025) & Efficient multimodal large language models & \checkmark & $\times$ & $\times$ & $\times$ & \checkmark \\
Xu et al. (2026) & MoE design and system realization & \checkmark & $\times$ & \checkmark & $\times$ & $\times$ \\
Cui et al. (2026) & MoE for remote sensing & $\times$ & $\times$ & \checkmark & \checkmark & $\times$ \\
Li and Tang (2026) & Multimodal alignment and fusion & $\times$ & \checkmark & $\times$ & \checkmark & \checkmark \\
\textbf{Ours} & \textbf{Roles of MoE in multimodal learning} & \checkmark & \checkmark & \checkmark & \checkmark & \checkmark \\
\bottomrule
\end{tabular}}
\end{table*}

\textbf{Extension to previous work}: It extends our earlier IJCAI survey \citep{zheng2026tackling}, which identified three broad roles: efficient computation, representation learning, and adaptation to imperfect modalities. While retaining these three perspectives, the present survey substantially extends the earlier version in four respects. First, we broaden the literature coverage and refine the taxonomy within each perspective, incorporating large-scale multimodal MoE architectures, lightweight experts, finer distinctions in representation, alignment, and interaction specialization, and adaptive designs for modality imbalance, distribution shift, and continual learning. Second, we introduce dedicated discussions of how these designs affect multimodal learning and synthesize multimodal-specific design considerations, including expert placement and identity, routing constraints and signals, knowledge sharing, and adaptation to changing input conditions. Third, we add a cross-domain application review covering healthcare and life sciences, Earth and environmental science, industrial and autonomous systems, human-centered intelligence, information systems, and emerging domains. We compare specialization targets, routing signals, and operational requirements across these applications to identify reusable design patterns and domain-dependent choices. Fourth, we revisit open challenges in light of recent advances in adaptive experts and knowledge management, examining interpretable routing, expert capacity management, and cross-expert coordination, while explicitly distinguishing temporal changes in individual modality relevance from changes in cross-modal dependencies. Together, these extensions provide a more detailed account of existing methods and a broader synthesis of the design choices, application requirements, and unresolved problems shaping multimodal MoE research.

\section{Background}
\subsection{Multimodal Learning}

Multimodal learning aims to model information from heterogeneous data sources and their relationships. Its major objectives can be broadly categorized into interaction, alignment, and translation \citep{baltruvsaitis2018multimodal,guo2019deep}. Interaction focuses on how multiple modalities contribute complementary, redundant, or synergistic information for prediction. Recent works\citep{xin2025i2moe,liang2023quantifying} further characterizes these relationships through uniqueness, redundancy, and synergy, providing a more interpretable view of cross-modal interactions. Alignment maps heterogeneous modalities into compatible representation spaces so that semantically related information can be associated across modalities. Modern vision--language models\cite{zhang2024clipmoe,zhai2025mmes,wang2023image} increasingly rely on such alignment to support cross-modal understanding. Translation goes further by transferring information from one modality to another, such as generating images from textual descriptions\cite{xu2024integration, xue2023raphael, zhang2026adafuse}. As multimodal models scale to more modalities and tasks, dense architectures face increasing computational and optimization costs. MoE provides a flexible alternative by activating only a subset of specialized experts for each input, enabling modality-, task-, or semantic-specific computation while avoiding proportional growth in computational cost.

\subsection{Brief History of Mixture of Expert}

The Mixture-of-Experts (MoE) architecture has become an important alternative to dense Transformer models by replacing standard Feed-Forward Network (FFN) layers with sparsely activated experts. Although modern models such as DeepSeek-MoE \citep{dai2024deepseekmoe}, Qwen-MoE \citep{yang2025qwen3}, and Llama-MoE \citep{zhu2024llama} have greatly increased its popularity, the idea dates back to the original MoE proposed by \citep{jacobs1991adaptive} in 1991. Its key principle is modular computation, where only a subset of specialized components is activated for each input rather than using the entire network. MoE entered the large-scale Transformer era with the Switch Transformer \citep{fedus2022switch}, which integrated sparse expert layers into Transformer architectures. By routing each token to only a subset of experts through Top-$k$ gating, MoE substantially increases model capacity without proportionally increasing the $FLOPs$ required for each token. This property has made sparse expert architectures particularly attractive for scaling large language and multimodal models.

Recent work has further diversified MoE architectures. DeepSeek-MoE \citep{dai2024deepseekmoe} introduces fine-grained experts by decomposing large FFNs into smaller expert units, enabling more flexible combinations of specialized knowledge. Expert-Choice Routing \citep{puigcerver2023sparse} reverses conventional token-to-expert routing by allowing each expert to select its preferred tokens, improving expert utilization. Mixture-of-Depth \citep{raposo2024mixture} extends conditional computation to network depth by allowing less informative tokens to bypass selected layers, while MoE-LoRA \citep{luo2024moelora} introduces lightweight low-rank experts for parameter-efficient fine-tuning. Together, these developments extend MoE from sparse FFN scaling toward more flexible forms of conditional computation across experts, depth, and adaptation parameters.

\begin{figure}[ht]
    \centering
    \includegraphics[page=1,width=\textwidth]{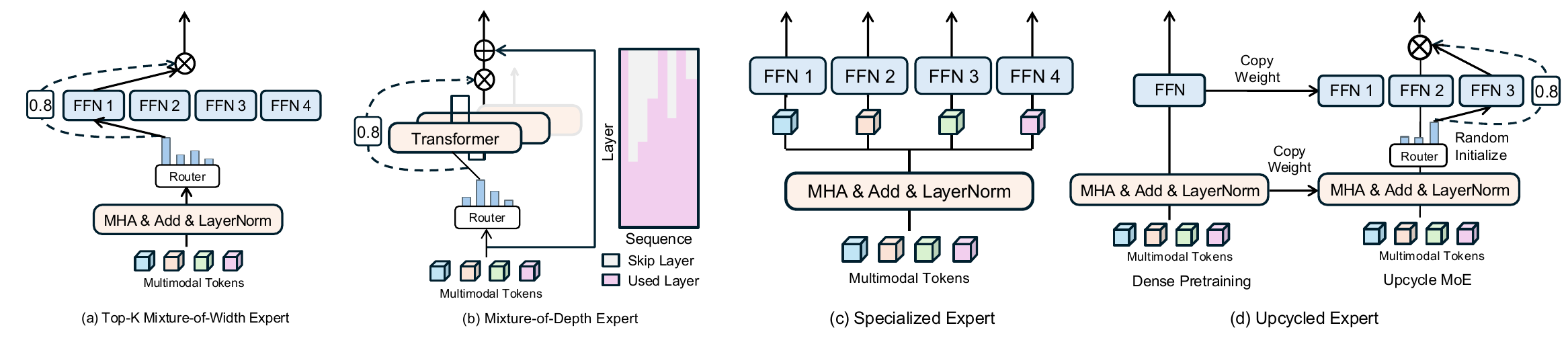}
    \caption{Types of Mixture-of-Experts for Multimodal Learning: (a) Top-K mixture-of-width routing dispatches each token to selected FFN experts after shared attention and combines their outputs using routing weights. (b) Mixture-of-depth routing selects tokens for Transformer computation at each layer, while unselected tokens bypass the block through the residual path. (c) Modality-specific experts apply separate FFN transformations to tokens from different modalities following shared attention. (d) MoE upcycling initializes multiple experts from pretrained dense FFN weights, retains shared attention, and introduces a randomly initialized router for sparse expert selection.
}
    \label{fig:types_moe}
\end{figure}

\section{Architecture of Mixture-of-Expert}

\subsection{Dense Mixture-of-Width Expert}

As shown in Fig. \ref{fig:types_moe}, one typical Mixture-of-Width (MoW), where each expert is activated during one forward pass, providing diverse opinions of experts for feature extraction. Given a data embedding $\mathbf{x} \in \mathbb{R}^d$ and an expert pool containing $N$ experts $\mathbf{E} = \{ E_i \}^N$. Dense Mixture-of-Width (DMoW) can be formulated as:

\begin{equation}
    \mathbf{E}(\mathbf{x}) = \mathbf{x} + \sum^N_{i=1} g_i E_i(\mathbf{x}) \label{eq:smoe}
\end{equation}

where $g_i$ is the gating score of corresponding expert $E_i$, producing by the router function $\mathcal{G} \in \mathbb{R}^{d \times N}$. While SoftMoE \citep{puigcerver2023sparse} reverses the dispatching method from the expert choice of typical MoW to the token choice operation. SoftMoE uses differentiable soft routing to construct expert inputs from weighted combinations of tokens. Each of the $N$ experts processes $p$ slots. Given input tokens $\mathbf{H} \in \mathbb{R}^{s \times d}$ and learnable parameters $\Phi \in \mathbb{R}^{d \times Np}$, SoftMoE can be formulated as:
\begin{equation}
    \mathbf{H}^{*}
    = \Psi_{Np}\mathbf{E}\bigl(\Psi_s^{\top}\mathbf{H}\bigr), 
    \qquad 
    \Psi_s = \operatorname{Softmax}_{s}(\mathbf{H}\Phi),
    \qquad
    \Psi_{Np} = \operatorname{Softmax}_{Np}(\mathbf{H}\Phi).
\label{eq:softmoe}
\end{equation}
Both matrices belong to $\mathbb{R}^{s \times Np}$. The dispatch matrix $\Psi_s$ is normalized over tokens, whereas the combine matrix $\Psi_{Np}$ is normalized over slots. Thus, $\Psi_s^{\top}\mathbf{H} \in \mathbb{R}^{Np \times d}$ contains the slot inputs, $\mathcal{E}$ applies each expert to its assigned $p$ slots, and $\Psi_{Np}$ combines the expert outputs into $\mathbf{H}^{*} \in \mathbb{R}^{s \times d}$.

\subsection{Sparse Mixture-of-Width Expert}
Although DMoW can provide diverse opinions from all experts, the computing efficiency can be a bottleneck to scaling up the model. In contrast to Dense activation, Sparse Mixture-of-Width (SMoW) introduces a sparse operation over the expert selection denoted by a sparse mask $\mathbf{m} \in \mathbb{R}^N$, such as Top-K function, where only one or a few experts are activated during forward. Such that, we can largely increase the trainable parameters without significantly enhancing the train/inference cost. We can show the formulation of SMoE as follows:

\begin{equation}
    \mathbf{E}(\mathbf{h}) = \mathbf{h} + \sum^N_{i=1} m_i g_i E_i(\mathbf{h}) \label{eq:smoe}
\end{equation}

where $g_i$ is the gating score of corresponding expert $E_i$, producing by the router function $\mathcal{G} \in \mathbb{R}^{d \times N}$. The sparsity of SMoE is enforced by a Top-K function, denoted as a binary mask $\mathbf{m}=\{0, 1\}^N$, which only those $g_i$ with lies in Top-K rank between other $g_i$. By contrast to FMoE, the differentiability of SMoE is partial due to the Top-K operation. This sparsity can potentially lead to the difficulty of optimization, such as expert collapse problem. However, SMoE is particularly computationally more efficient than FMoE due to the sparsity of selection. This trade-off is still a common concern in MoE community.

\subsection{Mixture-of-Depth Expert}

Mixture-of-Depth (MoD) is another architecture paradigm built upon a hypothesis: certain tokens require less processing than others \citep{raposo2024mixture}, indicating that not all tokens need to be passed through the entire model layers. In this case, a router instead learns to route the tokens to multiple experts, but to determine how many layers to skip, resulting in computational efficiency. Formally, For an input sequence $X^{l}=\{x_{i}^{l} |1 \le i \le S\}$ of length $S$ at a given layer $l$ , MoD first computes a scalar router weight $r_{i}^{l}=\omega_{\theta}x_{i}^{l}$ for each token using a linear projection. To maintain a static computation graph while reducing computational overhead, MoD introduces a sequence-level capacity constraint $C$, which defines the exact number of tokens processed by the block. This capacity establishes a dynamic threshold $P_{\beta}(R^{l})$, representing the $\beta$-th percentile of the set of router weights $R^{l}$, where $\beta=1-C/S$. The forward pass then functions as a conditional branch, evaluating whether a token's router weight exceeds the percentile threshold to dictate if it routes through the Transformer block $f(\cdot)$ or bypasses it via a residual connection:

\begin{equation}
    x_{i}^{l+1} = 
    \begin{cases} 
    r_{i}^{l} \cdot f(x_{i}^{l}) + x_{i}^{l}, & \text{if } r_{i}^{l} > P_{\beta}(R^{l}) \\ 
    x_{i}^{l}, & \text{if } r_{i}^{l} < P_{\beta}(R^{l}) 
    \end{cases}
    \label{eq:mod_routing}
\end{equation}

By multiplying the output of the computational path by the router weight, gradients can efficiently flow back through the router during training.

\section{Efficient MoE for Multimodal Learning}

\subsection{Sparse Experts for Large-Scale Multimodal Models}
\label{sec:moe_for_MLLM}

\begin{figure}[ht]
    \centering
    \includegraphics[page=1,width=\textwidth]{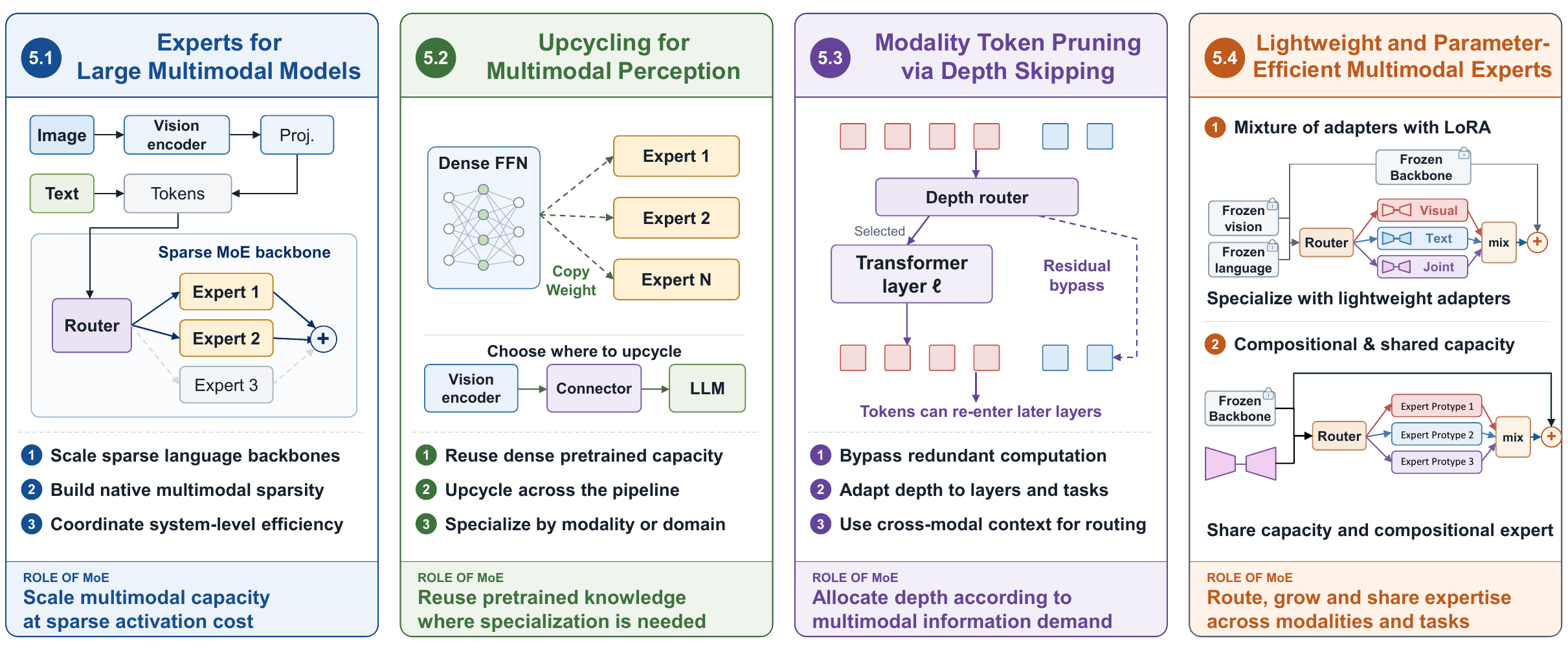}
    \caption{Efficiency-oriented multimodal Mixture-of-Experts designs. Section \ref{sec:moe_for_MLLM} discusses sparse expert backbones for large-scale multimodal models, examining modular extensions, native multimodal sparsity, and system-level efficiency. Section \ref{sec:upcycle_for_MLLM} focuses on multimodal upcycling, reviewing how pretrained dense parameters are converted into experts and where specialization is introduced across the multimodal pipeline. Section \ref{sec:efficient_mod} examines depth skipping, where routing selectively bypasses Transformer computation and adapts processing depth according to token redundancy, task requirements, and cross-modal context. Section \ref{sec:efficient_para} reviews lightweight and parameter-efficient experts, emphasizing how compact adaptations and parameter sharing balance specialized capacity with parameter cost.}
    \label{fig:section_5_overall_plot}
\end{figure}

Large-scale multimodal MoEs increasingly combine sparse capacity with decisions about where multimodal tokens enter the backbone and how long sequences are processed. Their development spans modular extensions of language MoE backbones, joint multimodal sparse training, and system-level efficiency, as summarized in Figure~\ref{fig:section_5_overall_plot}.

\subsubsection{Modular Extension via Language MoE Backbones.}
A common design retains a separate vision encoder while using sparse experts in the language backbone to process aligned visual and textual representations. DeepSeek-VL2\cite{wu2024deepseek}, Kimi K2.5\cite{kimi2026k25}, InternVL3.5\cite{internvl2025internvl35}, GLM-4.6V\cite{zai2025glm46v}, and Intern-S1-Pro\cite{interns1pro2026} exemplify this modular approach. Their designs differ beyond backbone scale: DeepSeek-VL2 uses shared and routed language experts; Kimi K2.5 jointly trains visual and textual capabilities; and Intern-S1-Pro employs grouped routing to support large-scale expert parallelism. InternVL3.5 additionally adapts visual resolution before language processing, although this resolution router is distinct from MoE expert routing. In these models, sparse experts primarily expand shared multimodal reasoning capacity after modality-specific encoding.

\subsubsection{Toward Native Multimodal Sparsity.}
Other models bring multimodal learning closer to sparse foundation-model training. Qwen3.5\cite{qwen2026qwen35} and Llama 4\cite{meta2025llama4} incorporate early text--vision fusion, while Seed1.6\cite{bytedance2025seed16} introduces visual data during mixed-modal continual pretraining. Qwen3-Omni\cite{xu2025qwen3} extends MoE to a Thinker--Talker architecture for multimodal perception and speech generation; LongCat-Flash-Omni\cite{longcat2025omni} combines shortcut-connected MoE with low-latency audio--visual interaction. ERNIE 5.0\cite{ernie2026technical} trains text, image, video, and audio under a unified objective with modality-agnostic expert routing and elastic expert capacity. Mistral Large 3\cite{mistral2025large3} and MiMo-V2.6-Pro-RL\cite{mimo2026v26pro} further demonstrate the scale of multimodal sparse backbones, although their publicly available descriptions provide less detail about modality-dependent expert behavior.

\subsubsection{From Model Sparsity to System-Level Efficiency.}
Sparse activation alone does not resolve the cost of long multimodal sequences. MiniMax-VL-01\cite{li2025minimax} and MiniMax-M3\cite{,minimax2026m3} combine MoE capacity with Lightning Attention and sparse attention respectively. Step-3 pairs sparse experts with Multi-Matrix Factorization Attention and Attention--FFN Disaggregation, while Nemotron 3\cite{stepfun2025step3} and Nano Omni\cite{nvidia2026nemotron3omni} integrates a multimodal MoE decoder with hybrid Mamba2--Transformer sequence modeling. Gemini 1.5\cite{team2024gemini} also uses MoE within a long-context multimodal system, though its expert organization remains undisclosed. Task-focused variants, including RynnBrain1.1\cite{rynnbrain2026}, Hy-Embodied-VLM-1.0\cite{hyembodied2026}, UI-Venus-1.5\cite{uivenus2026}, and Holo3.1\cite{holo2026}, extend sparse backbones to embodied and GUI tasks. Overall, large multimodal MoEs have progressed from adding sparse capacity after modality encoding toward combining expert computation with multimodal pretraining, sequence processing, and serving design. The distinction lies in where sparsity is introduced and which computational bottleneck it helps address.


\begin{figure*}[ht]
\centering
\begin{forest}
  for tree={
    forked edges,
    grow=east,
    reversed=true,
    align=left,
    anchor=west,
    parent anchor=east,
    child anchor=west,
    base=middle,
    font=\scriptsize,
    rectangle,
    line width=0.7pt,
    rounded corners,
    minimum width=2em,
    minimum height=1em,
    s sep=5pt,
    inner xsep=7pt,
    inner ysep=1pt,
    text ragged,
  },
  [Multimodal Mixture of Expert, middle,rotate=90,anchor=north
    [Efficient MoE, middle,align=left,text width=4.5em,minimum height=1em
        [Large Multimodal \\ Model, middle, text width=6em, align=left, minimum height=2em
            [{DeepSeek-VL2\cite{wu2024deepseek}, Kimi K2.5\cite{kimi2026k25}, InternVL3.5\cite{internvl2025internvl35}, GLM-4.6V\cite{zai2025glm46v}, Intern-S1-Pro\cite{interns1pro2026}, \\ Qwen3.5\cite{qwen2026qwen35}, Llama 4\cite{meta2025llama4}, Seed1.6\cite{bytedance2025seed16}, Qwen3-Omni\cite{xu2025qwen3}, LongCat-Flash-Omni\cite{longcat2025omni},\\ ERNIE 5.0\cite{ernie2026technical},  Mistral Large 3\cite{mistral2025large3}, MiMo-V2.6-Pro-RL\cite{mimo2026v26pro}, MiniMax-VL-01\cite{li2025minimax},\\ MiniMax-M3\cite{,minimax2026m3}, Nemotron 3\cite{stepfun2025step3}, Nano Omni\cite{nvidia2026nemotron3omni}, RynnBrain1.1\cite{rynnbrain2026}, \\ Hy-Embodied-VLM-1.0\cite{hyembodied2026}, UI-Venus-1.5\cite{uivenus2026}, Holo3.1\cite{holo2026}}, leaf, text width=28em]
        ]
        [Multimodal \\ Upcycling, middle, text width=6em,align=left
            [{MoE-LLaVA\cite{lin2024moe}, CuMo\cite{li2024cumo}, CLIP-MoE\cite{zhang2024clipmoe}, EvoMoE\cite{jing2026evomoe}, Eve\cite{rang2025eve}, Uni-MoE\cite{li2025uni}, \\ MoMa\cite{lin2024moma}, Med-MoE\cite{jiang2024med}}, leaf, text width=28em, align=left]
        ]
        [Modality Skipping, middle, text width=6em,align=left
            [{Gamma-MoD\cite{luo2024gamma}, Uni-MoD\cite{mao2025unimod}, VideoLLM-MoD \cite{wu2024videollmmod}, p-MoD\cite{zhang2025pmod}, Cat-MoD\cite{huang2026cat}}, leaf, text width=28em, align=left]
        ]
        [Multimodal \\ PEFT MoE, middle, text width=6em,align=left
            [{LaVIN\cite{luo2023cheap}, Omni-SMoLA\cite{wu2024omni}, AVMoE\cite{cheng2024mixtures}, MoME\cite{shen2024mome}, LiME\cite{kowsher2026lime}}, leaf, text width=28em, align=left]
        ]
    ]
    [Representation \\ Learning MoE, middle, align=left , text width=4.5em, minimum height=1em
        [Representation and \\ Alignment Learning, middle, text width=6em, align=left, minimum height=2em
            [{LIMoE ~\cite{nips/MustafaRPJH22}, MoVA ~\cite{zong2024mova}, Mono-InternVL \cite{luo2025monointernvl}, CLIP-FMoE\cite{tran2026clipfmoe}, S$^3$\cite{choi2026structural}, ERMoE\cite{cheng2026ermoe}, \\ MOON2.0\cite{nie2026moon}, R2-T2 ~\cite{li2025r2}}, leaf, align=left, text width=28em]
        ]
        [Modality-aware \\ Expert, middle, align=left, text width=6em,minimum height=1em
            [{Eagle\cite{iclr/ShiLWLRZHYSYSCT25}, ChartMoE\cite{xu2025chartmoe}, M$^3$-JEPA\cite{lei2024m3}, UNIALIGN\cite{zhou2025unialign}, Parrot\cite{sun2025parrot}, Med-MoE\cite{jiang2024med}, \\ MoMoK\cite{zhang2024multiple}, RAPHAEL\cite{xue2023raphael}}, leaf, align=left, text width=28em]
        ]
        [Multimodal \\ Interaction, middle, align=left, text width=6em,minimum height=1em
            [{MMoE\cite{emnlp/YuQJSML24}, I$^2$MoE\cite{xin2025i2moe}, MMCas\cite{jing2026modeling}, MERGE\cite{han2025massively}, AVMoE\cite{cheng2024mixtures}, MoAI\cite{lee2024moai}, \\ REMOTE\cite{lin2025remote}, SEATrack\cite{su2026seatrack}, QME\cite{zhu2025quality}, UGG-ReID\cite{wan2025ugg}, QA-MoE\cite{zhu2026qamoe}, ForceVLA\cite{yu2025forcevla}}, leaf, align=left, text width=28em]
        ]
    ]
    [Modality \\ Adaption MoE, middle, align=left, text width=4.5em, minimum height=1em
        [Flexible Modality, middle, align=left, text width=6em, minimum height=1em
            [{Task-Conditional MoE \cite{novosad2024task}, FuseMoE\cite{han2024fusemoe}, Flex-MoE\cite{yun2024flex}, MCMoE\cite{xu2026mcmoe}, ConfSMoE\cite{zheng2025rethinking},\\ MARS\cite{baek2026mars}}, leaf, align=left, text width=28em]
        ]
        [Modality Imbalance \\ and Shift, middle, align=left, text width=6em, minimum height=1em
            [{Moon2.0 \cite{nie2026moon}, Yin et al.\cite{yin2026imbalanced}, MiSTER-E\cite{dutta2026mixture}, CoPRIME\cite{naderi2026coprime}, MIME\cite{zhang2026mime},  QA-MoE\cite{zhu2026qamoe}, \\ Casual-MoE\cite{lan2026causalmoe}}, leaf, align=left, text width=28em]
        ]
        [LifeLong \\ Multimodal Experts, middle, align=left, text width=6em, minimum height=1em
            [Expert Expansion \\ and Isolation, middle, align=left, text width=5em, minimum height=1em
                [{MoE-Adapter\cite{yu2024moeadapters}, MoE-Adapters++\cite{yu2025moeadapters}, CoIN\cite{chen2024coin}, 
                CL-MoE\cite{huai2025clmoe}, \\ D-MoLE\cite{ge2025dmole}, BranchLoRA\cite{zhang2025branchlora}, 
                ProgLoRA\cite{yu2025proglora}, SMoLoRA\cite{wang2025smolora},\\ Incre-MoE\cite{cheng2026incremoe}, 
                Chee et al.\cite{chee2025multimodal}, Zhao et al.\cite{zhao2026taskshared}},
                leaf, align=left, text width=21em]
            ]
            [Routing Stability \\ and Expert Reuse, middle, align=left, text width=5em, minimum height=1em
                [{DyMoE\cite{zhao2026dymoe}, SAME\cite{xie2026same}, PASs-MoE\cite{hou2026passmoe}, 
                Pan et al.\cite{taskadaptive2026vlm},\\ MoE-Confidence\cite{alforhad2026moeconfidence}, 
                Calibrated Bayesian MoE\cite{mozaffari2026bayesianmoe}, CRAM\cite{tang2026cram}, \\
                AGO-MoE\cite{fa2026rank1}, MoBLoRA\cite{chen2026moblora}},
                leaf, align=left, text width=21em]
            ]
            [Expert Lifecycle \\ Management, middle, align=left, text width=5em, minimum height=1em
                [{LiveEdit\cite{chen2025liveedit}, Multi-MELO\cite{chen2025multimelo}, 
                TIME\cite{xu2026time}, SEUF\cite{zhuang2025seuf}, EXACT\cite{wang2027exact}},
                leaf, align=left, text width=21em]
            ]
        ] 
    ] 
  ] 
\end{forest}
\caption{Taxonomy of multimodal Mixture-of-Experts approaches reviewed in this survey. The taxonomy groups methods by their primary role: efficient MoE architectures scale multimodal models or allocate computation through sparse backbones, upcycling, modality skipping, and parameter-efficient experts; expert-specialization methods organize multimodal representations through representation learning, alignment, and interaction; and adaptive MoE methods respond to missing or flexible modalities, modality imbalance and distribution shift, and the evolution of expert knowledge. The subcategories further distinguish the main design strategies within each direction, offering an overview of how multimodal MoE supports efficient computation, structured expertise, and adaptation to changing data conditions.}
\label{fig:taxonomy_of_MMoE}
\end{figure*}

\subsection{Upcycling Experts for Multimodal Perception}
\label{sec:upcycle_for_MLLM}
\subsubsection{Dense-to-Sparse Expert Initialization.}
Upcycling~\citep{komatsuzaki2022sparse,he2024upcycling} initializes multiple experts from pretrained dense FFNs, preserving learned representations while enabling subsequent specialization. MoE-LLaVA~\cite{lin2024moe} applies this strategy after establishing visual--language alignment, thereby introducing sparse capacity without relearning the multimodal backbone from scratch. LLaVA-MoD~\cite{shu2024llavamod} further combines dense-to-sparse conversion with knowledge distillation, treating expert initialization as a mechanism for transferring teacher knowledge rather than merely replicating parameters. However, replicated experts may remain functionally similar; EvoMoE~\cite{jing2026evomoe} addresses this limitation by jointly encouraging expert diversification and token-dependent routing. These methods therefore move upcycling from simple parameter replication toward initialization schemes that preserve pretrained knowledge while promoting useful expert diversity.

\subsubsection{Pipeline-Aware Upcycling.}
Upcycling need not be restricted to the language backbone. CuMo~\cite{li2024cumo} co-upcycles the vision encoder and connector, placing sparse capacity closer to where visual representations are extracted and aligned. CLIP-MoE~\cite{zhang2024clipmoe} instead constructs a reusable sparse vision--language encoder from complementary dense CLIP models, decoupling upcycling from a particular language backbone. Eve~\cite{rang2025eve} follows a more selective strategy by preserving the original language FFN while adding visual-specific expert capacity. Across these designs, the key question shifts from whether to upcycle toward which multimodal component benefits most from additional sparse capacity, revealing a trade-off between broadly transforming the backbone and selectively specializing high-demand stages of the pipeline.

\subsubsection{Modality- and Domain-Aware Upcycling.}
More recent methods incorporate multimodal structure directly into expert initialization rather than relying on specialization to emerge after generic replication. Uni-MoE~\cite{li2025uni} pre-specializes experts for different modalities before integrating them into a shared sparse language backbone, while MoMa~\cite{lin2024moma} preserves modality structure through modality-specific expert groups within an early-fusion architecture. Med-MoE~\citep{jiang2024med} extends the same principle to domain specialization by organizing experts around heterogeneous medical imaging domains. Compared with generic upcycling, these approaches introduce stronger inductive biases during initialization, trading some flexibility for more explicit modality- or domain-specific expertise. Overall, multimodal upcycling evolves along three coupled decisions: how pretrained capacity is converted into experts, where sparse capacity is introduced, and what structure should guide expert specialization. This reframes upcycling from a generic dense-to-sparse conversion technique into a multimodal-aware mechanism for allocating pretrained knowledge across heterogeneous components, modalities, and domains.

\subsection{Modality Token Pruning via Depth Skipping}
\label{sec:efficient_mod}

Mixture-of-Depths (MoD) extends sparse multimodal computation from selecting which expert processes a token to deciding whether and how deeply the token should be processed. Multimodal MoD has consequently progressed from redundancy-based skipping toward increasingly adaptive and context-aware allocation of computational depth.

\subsubsection{Multimodal Depth Sparsity.}
MoD~\cite{raposo2024mixture} introduces vertical conditional computation by allowing selected tokens to bypass Transformer layers while remaining available through residual paths, providing a recoverable alternative to irreversible token pruning. This property is particularly relevant to multimodal models, where visual inputs often contain substantially more redundant tokens than text. VideoLLM-MoD~\cite{wu2024videollmmod} applies this principle to long video-language sequences by selectively bypassing redundant visual tokens, while $\gamma$-MoD~\cite{luo2024gamma} converts pretrained dense MLLMs into depth-sparse models by identifying redundant layers and jointly routing vision and language tokens. Compared with permanent token removal, these methods shift the efficiency objective toward preserving information while selectively reducing the amount of computation it receives.

\subsubsection{Multimodal-Adaptive Depth Allocation.}
Once depth sparsity is available, the problem shifts from whether to skip computation toward how much computation different tokens should receive. p-MoD~\cite{zhang2025pmod} adapts the active-token ratio across network depth, exploiting increasing visual redundancy in deeper layers, whereas UniMoD~\cite{mao2025unimod} conditions depth allocation on task requirements in unified multimodal understanding and generation. Compared with $\gamma$-MoD~\cite{luo2024gamma}, which primarily identifies where depth sparsity should be introduced, these methods make the computational budget itself adaptive to layer position or task demand. Cat-MoD~\cite{huang2026cat} extends this idea further by using textual semantics to guide the allocation of computation to visual tokens during multimodal alignment, showing that the processing demand of one modality can depend on evidence from another. The progression therefore moves from redundancy-aware skipping toward budget allocation conditioned on network depth, task requirements, and cross-modal context.

Overall, multimodal MoD evolves from recoverable redundancy reduction toward adaptive and context-dependent computation allocation. Its central question shifts from which tokens can be skipped to how computational depth should be distributed according to redundancy, task demand, and cross-modal relevance.

\subsection{Lightweight and Parameter-Efficient Multimodal Experts}
\label{sec:efficient_para}

Unlike sparse backbones or depth skipping, which primarily reduce activated computation, lightweight multimodal MoE targets the parameter cost of specialization itself by introducing compact experts into largely frozen pretrained models. LaVIN~\cite{luo2023cheap} represents an early form of conditional adaptation, routing multimodal instructions through lightweight modality adapters, whereas Omni-SMoLA~\cite{wu2024omni} replaces a small discrete adapter set with soft mixtures of low-rank residual experts, enabling finer-grained specialization across visual, textual, and multimodal representations. AVMoE~\cite{cheng2024mixtures} takes a complementary approach by structuring experts according to function, explicitly separating unimodal processing from cross-modal interaction. As the number of tasks and modalities grows, however, independently instantiated experts can still accumulate substantial parameter overhead. MoME~\cite{shen2024mome} addresses this by distributing specialized capacity across visual and language pathways, while LiME~\cite{kowsher2026lime} pushes sharing further by coupling a common parameter-efficient transformation with lightweight expert-specific prototypes. Across these works, the design trend therefore shifts from making each expert cheaper toward deciding which parts of expertise should remain specialized and which can be shared. This exposes a central trade-off: stronger sharing reduces the marginal cost of adding expertise but may constrain expert diversity, whereas more function- or modality-specific experts provide richer specialization at greater parameter cost. Lightweight multimodal MoE should therefore be viewed not merely as parameter reduction, but as a problem of allocating a limited adaptation budget across shared and specialized multimodal functions.

\subsection{Discussion}

\textbf{Impact of Different MoE Architectures on Multimodal Learning.}
Different MoE architectures address different sources of inefficiency in multimodal learning. Sparse MoE backbones increase model capacity while keeping the activated computation constrained, making them useful when a model needs greater representational capacity without proportionally increasing the computation per input; their multimodal benefit also depends on whether experts are introduced after modalities enter a shared representation space or earlier within modality-specific processing \cite{wu2024deepseek,team2025kimi}. Upcycling converts pretrained capacity into specialized experts at selected stages, such as the vision encoder, multimodal connector, or language backbone, helping improve modality-specific perception, cross-modal alignment, or shared multimodal reasoning according to where specialization is needed; it is particularly suitable when strong pretrained components are already available \cite{li2024cumo,zhang2024clipmoe,rang2025eve}. Mixture-of-Depth allocates different amounts of computation to multimodal tokens, reducing unnecessary processing when modalities contain long or redundant sequences, and is therefore useful when token-level computation is a major efficiency bottleneck; however, overly aggressive skipping may remove information needed for subsequent cross-modal reasoning \cite{wu2024videollmmod,luo2024gamma,zhang2025pmod,mao2025unimod}. Lightweight experts reduce the parameter cost of modality- or function-specific specialization, allowing new expertise to be added under a limited parameter budget and making them useful when the pretrained backbone should remain largely frozen; this comes with a trade-off between greater expert diversity and more parameter sharing \cite{luo2023cheap,wu2024omni,cheng2024mixtures,shen2024mome,kowsher2026lime}. Thus, selecting an MoE architecture involves matching its capacity, placement, computation allocation, and parameter cost to the main bottleneck and specialization needs of the multimodal system.

\textbf{Multimodal-Specific MoE Design Considerations.}
Beyond these general efficiency mechanisms, multimodal learning introduces several MoE design considerations that are less prominent in unimodal settings. (1) Expert placement becomes a multi-stage decision because multimodal models typically contain modality encoders, alignment or connector modules, and shared reasoning components. Introducing experts at different stages therefore changes whether specialization occurs before, during, or after cross-modal alignment \cite{li2024cumo,zhang2024clipmoe,rang2025eve}. (2) Computational demand is inherently asymmetric across modalities. Visual and video streams can contain substantially more tokens and redundancy than language, motivating modality-dependent allocation of depth and computation rather than a uniform computational budget \cite{wu2024videollmmod,luo2024gamma,zhang2025pmod}. (3) Routing can be cross-modally conditioned, such that the amount of computation required by one modality depends on information provided by another modality. For example, textual information can guide the allocation of computation to visual tokens \cite{huang2026cat}. (4) Expert specialization can be organized around multimodal functions rather than only semantic content, allowing separate expertise for modality-specific processing, shared representations, and cross-modal interactions \cite{cheng2024mixtures,shen2024mome,kowsher2026lime}. These properties distinguish multimodal MoE design from conventional sparse scaling, where experts and computational budgets are often treated as more homogeneous components.

\section{Expert Specialization across Representations, Modalities, and Interactions}

\begin{figure}[ht]
    \centering
    \includegraphics[page=1,width=\textwidth]{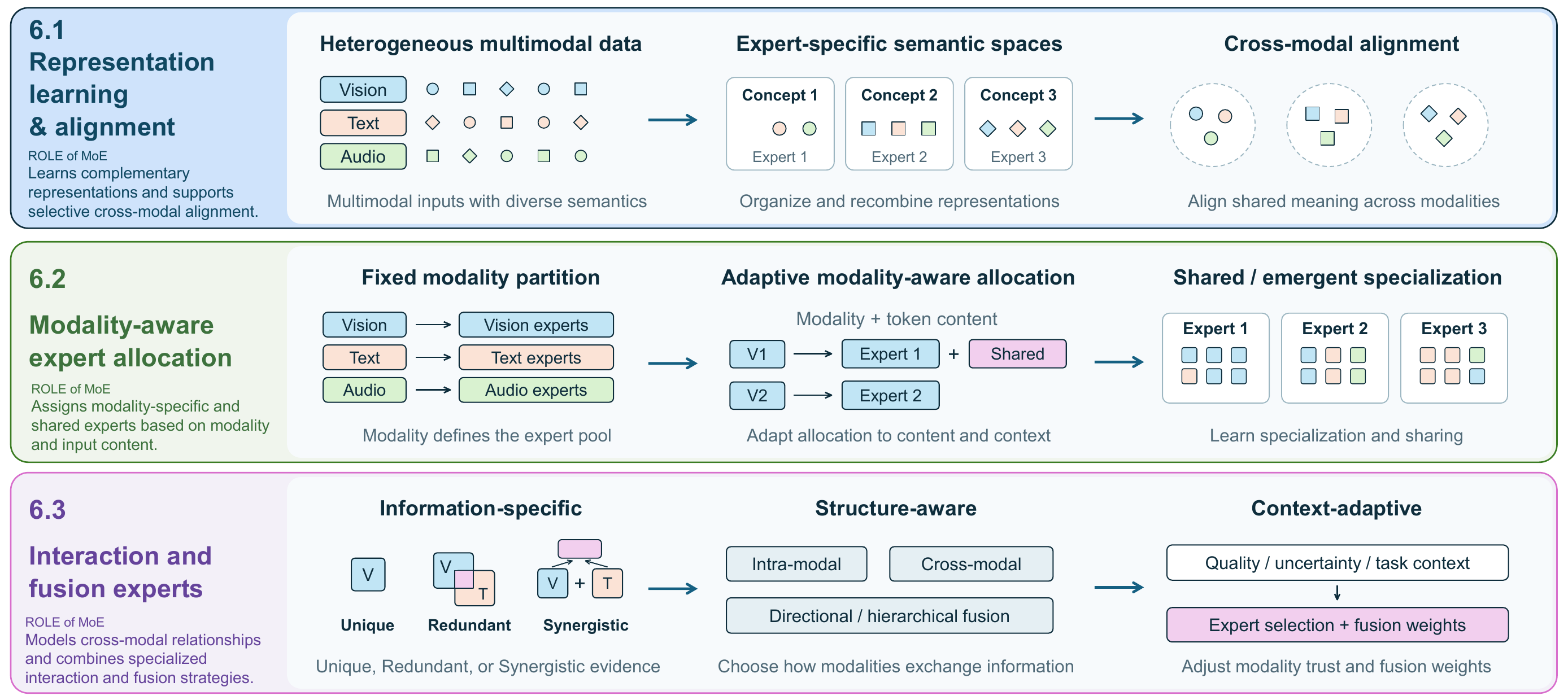}
    \caption{Expert Specialization across Representation, Modalities and Interactions: Section~\ref{sec:MoE_repre_and_align} discusses how MoE structures and aligns heterogeneous representation spaces through semantic specialization, expert selection, and alignment mappings. Section~\ref{sec:specific_moe} focuses on modality-specific experts, where shared dense transformations are progressively replaced or complemented by dedicated, adaptively routed, or emergently specialized expert pathways. Section~\ref{sec:moe_inter_fuse} then shifts the specialization target from individual representations to the relationships between modalities, examining information-specific, structure-aware, and context-adaptive experts for more flexible multimodal interaction and fusion.}
    \label{fig:section_6_overall_plot}
\end{figure}

Multimodal learning involves heterogeneous representations with distinct statistical and semantic structures, making fully shared processing prone to interference and insufficient specialization. MoE addresses this limitation by decomposing shared capacity into specialized computational pathways that can preserve modality-specific structures while selectively modeling cross-modal dependencies. Depending on the target of specialization, existing methods can be organized around three levels: representation and alignment, modality-aware expert allocation, and cross-modal interaction and fusion. This section reviews how expert specialization evolves across these three levels. Figure~\ref{fig:section_6_overall_plot} summarizes these three levels of expert specialization and their progression.

\subsection{Multimodal Representation and Alignment Experts}
\label{sec:MoE_repre_and_align}

\subsubsection{Multimodal Representation Learning.}
Representation learning aims to capture the intrinsic structure of data, while the heterogeneous representation spaces of multimodal inputs naturally motivate MoE models to distribute different semantic patterns across specialized experts. Early studies mainly rely on emergent specialization within a shared sparse representation space. LIMoE~\cite{nips/MustafaRPJH22} demonstrates this paradigm in large-scale image--text contrastive learning using a single sparse Transformer, where modality-aware entropy regularization prevents one modality from dominating the expert pool while allowing experts to develop modality and semantic preferences. IMP-MoE~\cite{akbari2023alternating} retains the shared expert space but addresses the optimization conflicts that arise when heterogeneous modalities and objectives jointly update the model, using alternating gradient updates to improve the stability of multimodal representation learning. These approaches show that useful specialization can emerge without explicitly partitioning the representation space, but also expose the limitations of relying solely on routing dynamics to resolve multimodal heterogeneity. Subsequent works move toward more structured representation specialization by explicitly separating heterogeneous representation sources and semantic subspaces. MoVA~\cite{zong2024mova} and Mono-InternVL~\cite{luo2025monointernvl} both reduce reliance on a fully shared representation space, either by dynamically selecting complementary pretrained vision experts or by introducing dedicated visual experts within the language model. CLIP-FMoE~\cite{tran2026clipfmoe} and S$^3$~\cite{choi2026structural} further make this specialization explicit at the semantic level, organizing representations into constrained semantic regions or concept-level expert components. These methods therefore shift multimodal representation learning from emergent expert preferences toward explicitly structured representation spaces. As representation spaces become more specialized, routing increasingly serves to match inputs with suitable representations. ERMoE~\cite{cheng2026ermoe} directly routes tokens according to their similarity with expert-specific subspaces, while MOON2.0~\cite{nie2026moon} adapts expert utilization according to the modality composition of each input. Together, these methods couple expert selection more closely with the structure and imbalance of multimodal representations. Finally, representation selection can remain suboptimal even when both the experts and their routing policies are well trained. R2-T2~\cite{li2025r2} extends representation-aware expert selection to inference by adjusting expert routing weights at test time, allowing existing expert representations to be recombined according to individual test instances without retraining the underlying representation encoder. Overall, multimodal MoE representation learning evolves from emergent specialization toward explicitly structured and adaptively selected representation spaces. Across these methods, the main design trade-off lies between the flexibility of emergent expert specialization and the stronger inductive bias introduced by explicitly organized semantic or modality-specific representations.

\subsubsection{Multimodal Alignment.}
Multimodal alignment uses MoE to replace a single universal mapping with specialized mappings across heterogeneous semantic spaces. EAGLE~\cite{iclr/ShiLWLRZHYSYSCT25} combines complementary pretrained vision experts before aligning them with the language space, while ChartMoE~\cite{xu2025chartmoe} places specialization directly in the connector through multiple alignment experts. Later methods move from specialized mappings toward structuring the alignment space itself. M$^3$-JEPA~\cite{lei2024m3} separates modality-specific and shared information in a latent space, UNIALIGN~\cite{zhou2025unialign} extends alignment across multiple modalities through modality-aware experts and soft binding, and Parrot~\cite{sun2025parrot} conditions visual alignment on linguistic context. Other works introduce domain-specific alignment priors: Med-MoE~\cite{chopra2025medmoe} combines shared and modality/domain-specific experts for heterogeneous medical imaging, MoMoK~\cite{zhang2024multiple} separates modality-specific knowledge in multimodal knowledge graphs, and RAPHAEL~\cite{xue2023raphael} specializes experts by spatial concepts and diffusion stages for text-to-image generation. Overall, multimodal MoE alignment progresses from combining specialized representations toward learning context-, modality-, and domain-dependent alignment spaces. Compared with representation learning, where experts mainly organize what features are represented, alignment experts increasingly determine how heterogeneous representations should be mapped and connected, revealing a broader shift from fixed correspondence toward conditional alignment.

\subsection{Modality-aware Experts}
\label{sec:specific_moe}

Modality-aware experts exploit multimodal heterogeneity by assigning modality-dependent computation to different expert pathways. The central question is therefore not simply whether modalities should use different experts, but how strongly modality identity should constrain expert specialization. Existing approaches form a continuum from explicit modality-specific partitioning, through adaptive modality-aware routing, to shared or emergent specialization where modality identity becomes a softer inductive bias.

\subsubsection{Static Modality-Specific Expert Partitioning.}
Early approaches impose a direct correspondence between modality identity and model parameters. VLMo~\cite{bao2022vlmo} and VL-MoE~\cite{shen2023scaling} separate vision- and language-specific FFNs while retaining shared attention, whereas BEiT-3~\cite{wang2023image} extends this private--shared structure with modality-specific and shared components at foundation-model scale. In pretrained LLMs, Mono-InternVL~\cite{luo2025monointernvl} and Eve~\cite{rang2025eve} preserve the original language pathway while adding visual-specific expert capacity, thereby reducing interference with pretrained linguistic knowledge. MoT~\cite{liang2024mixture} pushes specialization beyond FFNs by assigning modality-specific Transformer parameters while preserving global self-attention. Across these designs, the key distinction is between separating modality-specific transformations and preserving cross-modal communication: stronger parameter separation can protect heterogeneous representations, but interaction pathways must remain shared enough to support multimodal reasoning. Overall, static partitioning establishes modality identity as an explicit architectural prior, progressing from FFN-level separation toward broader modality-specific computation while retaining shared mechanisms for cross-modal interaction.

\subsubsection{Adaptive Modality-Aware Expert Routing.}
Static partitioning can be overly restrictive because tokens within the same modality may require different expertise. MoMa~\cite{lin2024moma} therefore uses modality identity to define coarse expert pools while allowing token-level routing within each pool. EVE~\cite{chen2024eve} further varies this constraint across depth, using stronger modality-specific routing in shallow layers and softer sharing in deeper layers. MoHAVE~\cite{kim2025mohave} introduces hierarchical routing, first balancing contributions across modality-specific expert groups and then selecting experts within each group according to sample-dependent reliability. EvoMoE~\cite{jing2026evomoe} relaxes modality constraints further by conditioning routing jointly on modality and token content. These methods progressively transform modality identity from a hard routing boundary into a flexible routing prior that interacts with layer depth, semantic content, and input reliability. Overall, adaptive routing preserves modality structure without enforcing fixed expert assignments, revealing a trade-off between strong modality priors and finer-grained content-dependent specialization.

\subsubsection{Shared and Emergent Modality Specialization.}
As routing becomes more flexible, a further question is whether experts need explicit modality identities at all. MoIIE~\cite{wang2025moiie} and MoST~\cite{lou2026most} retain private--shared structures, combining modality-specific experts with shared experts for cross-modal semantics. SMAR~\cite{xia2025smar} weakens this correspondence by encouraging modality-dependent routing preferences without assigning fixed expert identities, while SMoES~\cite{bo2026smoes} makes the strength of modality guidance layer-dependent. At the other extreme, Aria~\cite{li2024aria} develops modality-preferred experts despite modality-agnostic routing, and Shukor et al.~\cite{shukor2025scaling} show that such emergent specialization can outperform deterministic modality-aware routing. Their analyses further suggest that specialization varies with depth, with stronger modality dependence in some layers and greater sharing in others. Overall, modality specialization evolves from explicit private--shared structures toward soft guidance and ultimately emergent specialization. This suggests that modality identity need not always be encoded as a fixed architectural constraint; instead, the appropriate balance between specialization and sharing may itself be learned from token semantics, network depth, and input context.

\subsection{Multimodal Interaction and Fusion Experts} \label{sec:moe_inter_fuse}

Expert specialization can target not only individual modality representations but also the relationships through which modalities exchange and integrate information. Interaction and fusion experts model complementary, redundant, directional, or context-dependent cross-modal relationships. Research in this area has progressed from identifying what information is shared between modalities, to determining how it should be exchanged, and more recently to adapting when and to what extent modalities should interact.

\subsubsection{Information-Specific Interaction Experts.}
Early interaction-aware MoEs mainly specialize experts according to the types of information shared across modalities. Recent studies \citep{liang2023quantifying,liang2023multimodal} use Partial Information Decomposition (PID) \citep{williams2010nonnegative} to characterize multimodal interaction through uniqueness, redundancy, and synergy, providing an information-theoretic basis for decomposing cross-modal relationships. MMoE \citep{emnlp/YuQJSML24} operationalizes this idea by assigning samples to experts specialized in these interaction categories, shifting fusion from a universal mechanism toward interaction-dependent processing. However, its interaction labels are fixed pseudo-labels derived from pretrained model predictions, which may limit the approach in more complex multimodal settings. I$^2$MoE \citep{xin2025i2moe} relaxes this reliance on predefined categories by learning interactions through specialized fusion experts and a reweighting mechanism. MMCas \cite{jing2026modeling} follows a similar information-theoretic approach while extending uniqueness, redundancy, and synergy modeling to a larger collection of heterogeneous modalities. Moving beyond static information categories, MERGE \cite{han2025massively} incorporates temporal dependencies between modality pairs into expert routing, allowing experts to capture interaction patterns that change over time. Overall, this line of work progresses from predefined interaction categories toward learned, input-dependent, and temporally adaptive modeling. Accordingly, MoE's role shifts from separating fixed forms of multimodal information to identifying which interaction patterns are relevant for different inputs and temporal contexts.

\subsubsection{Structure-Aware Fusion Experts.}
While information-specific experts characterize the semantic nature of multimodal relationships, structure-aware fusion experts focus on how these relationships are computationally realized. AVMoE \cite{cheng2024mixtures} separates unimodal and cross-modal processing by treating unimodal and cross-modal adapters as different experts, allowing a router to adjust their relative use across tasks. MoAI \citep{lee2024moai} adds structure to cross-modal information flow through attention experts for pairwise interactions among visual, language, and auxiliary representations, replacing a single fusion operator with directional pathways. REMOTE \cite{lin2025remote} moves from modality-pair interactions to relation-dependent fusion by selecting features for different intra- and inter-modal relational triplets. SEATrack \cite{su2026seatrack} extends this structural specialization to hierarchical cross-modal relationships, modeling global relations across sensing modalities. Fusion specialization can also move beyond representations: QME \cite{zhu2025quality} formulates multimodal integration as a mixture of score-fusion experts that combine modality-specific similarity distributions. Together, these methods trace a progression from separating unimodal and cross-modal processing to organizing information flow through directional, relational, and hierarchical pathways, and ultimately to decision-level fusion. MoE thus provides specialized routes for different forms and levels of multimodal interaction.

\subsubsection{Context-Adaptive Fusion Experts.}
More recent studies recognize that the appropriate interaction strategy depends not only on modality content and structure but also on input reliability and context. UGG-ReID \cite{wan2025ugg} introduces uncertainty-guided expert routing, using sample-level modality uncertainty to select experts and reduce the influence of unreliable information during fusion. QA-MoE \cite{zhu2026qamoe} extends this adaptation by modeling modality quality along a continuous reliability spectrum, allowing routing to respond to varying levels of input degradation rather than only discrete missing-modality conditions. Beyond signal quality, ForceVLA \cite{yu2025forcevla} shows that interaction can also depend on task context, dynamically integrating vision-language representations with real-time force feedback during action decoding. These methods mark a shift from content- and structure-dependent fusion toward condition-dependent interaction, in which routing also reflects reliability, uncertainty, and physical context. The role of interaction experts consequently expands from selecting which cross-modal representation to construct to adapting whether a modality should be trusted, how strongly it should contribute, and which form of interaction suits the current input.

\subsection{Discussion}

\textbf{Impact of Different Expert Specialization Designs on Multimodal Learning.}
Different forms of expert specialization affect different stages of multimodal learning. Experts in representation learning determine how heterogeneous modality features are organized before they are integrated, allowing different semantic structures to occupy specialized subspaces rather than forcing all modalities into a uniformly shared representation \cite{nips/MustafaRPJH22,tran2026clipfmoe,choi2026structural,cheng2026ermoe}. Alignment experts improve how representations from different modalities are mapped and connected, allowing correspondences to adapt to modality, domain, or linguistic context \cite{iclr/ShiLWLRZHYSYSCT25,xu2025chartmoe,zhou2025unialign,sun2025parrot}. Modality-aware routing helps preserve modality-specific structures and reduce interference where modalities differ, while supporting expert sharing when their semantic or computational requirements overlap \cite{bao2022vlmo,shen2023scaling,lin2024moma,chen2024eve,shukor2025scaling}. Interaction and fusion experts further benefit multimodal learning by specializing the relationships between modalities: they help distinguish unimodal processing from cross-modal exchange, and adapt the form and strength of that exchange to the input or task \cite{cheng2024mixtures}. Together, these designs let MoE adapt task-relevant computation, feature organization, cross-modal alignment, and information exchange to the demands of each input, rather than relying on a single fixed pathway across multimodal tasks.

\textbf{Multimodal-Specific Expert Design Considerations.}
These designs reveal several properties that distinguish expert specialization in multimodal MoE from conventional expert decomposition. (1) Expert identity does not need to correspond directly to modality identity. Experts can instead represent semantic subspaces, modality-specific transformations, shared multimodal knowledge, or particular forms of cross-modal interaction \cite{tran2026clipfmoe,choi2026structural,wang2025moiie,lou2026most,cheng2024mixtures}. (2) Modality identity can constrain routing at different strengths. It may define a fixed expert partition, restrict the candidate expert pool, provide a soft routing prior, or disappear entirely while modality-preferred specialization emerges from training \cite{lin2024moma,chen2024eve,xia2025smar,bo2026smoes,li2024aria,shukor2025scaling}. (3) The appropriate degree of expert sharing is not constant across the multimodal pipeline. Stronger modality-specific processing can be beneficial where heterogeneous structures need to be preserved, whereas greater sharing becomes important when representations must support common semantics and cross-modal reasoning \cite{chen2024eve,wang2025moiie,lou2026most}. (4) Multimodal MoE can specialize not only the processing of individual modalities but also the computation between them. Experts may therefore represent different interaction or fusion functions, making cross-modal relationships themselves a target of specialization rather than treating fusion as a single shared operation \cite{cheng2024mixtures,xin2025i2moe,han2025massively,yu2026pathmoe}. The central design problem is consequently not simply how to divide parameters among experts, but what should define expert identity, how strongly modality identity should constrain routing, where knowledge should remain shared, and which cross-modal relationships require specialized computational pathways.

\section{Adaptive MoE for Imperfect and Evolving Multimodal Data}

\begin{figure}[ht]
    \centering
    \includegraphics[page=1,width=\textwidth]{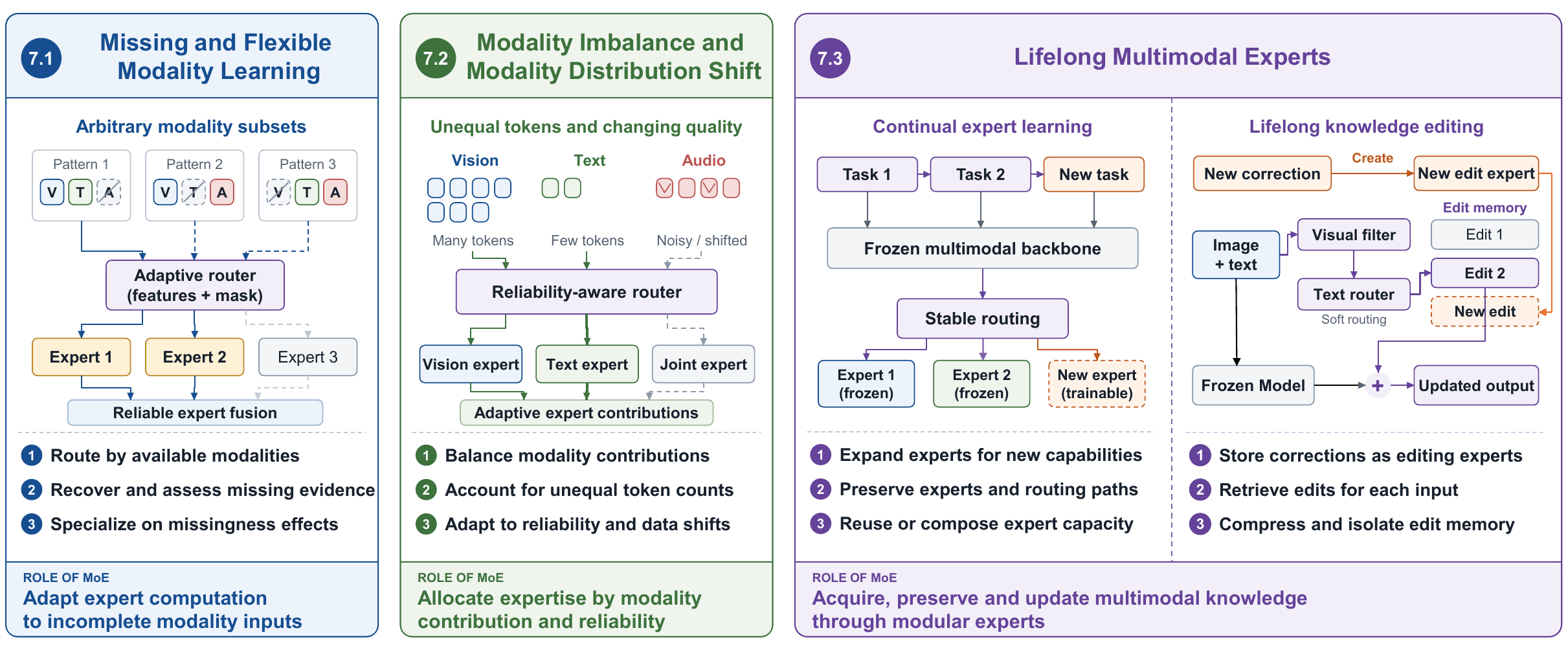}
    \caption{Adaptive MoE for Imperfect and Evolving Multimodal Data: Section~\ref{sec:missing_flexible} discusses missing and flexible modality learning, where experts adapt to incomplete and varying modality inputs. Section~\ref{sec:modality_imbalance_shift} focuses on modality imbalance and distribution shift, emphasizing adaptive expert allocation under unequal modality contributions and changing reliability. Section~\ref{sec:lifelong_experts} further extends adaptation over time, examining how experts can be expanded, preserved, reused, edited, and selectively removed as multimodal knowledge evolves.}
    \label{fig:section_7_overall_plot}
\end{figure}

A fundamental challenge of multimodal learning is that multimodal information is rarely complete, equally reliable, or stationary over time. Modalities may be missing, imbalanced, corrupted, or affected by distribution shifts, while the knowledge required by multimodal models may also continuously evolve after deployment. Dense multimodal models typically accommodate these variations within largely fixed parameters and computational pathways, which can make them sensitive to changing input conditions and introduce interference when adapting to new knowledge. MoE provides a more flexible alternative by dynamically allocating specialized computation according to the availability, reliability, and evolution of multimodal information. Recent research shows a clear trend from adapting experts to predefined modality configurations toward more dynamic forms of adaptation based on information quality, distributional variation, and long-term knowledge evolution. In this chapter, we review this progression from three perspectives. Figure~\ref{fig:section_7_overall_plot} summarizes the progression from input-level adaptation to modality reliability and lifelong expertise.

\subsection{Missing and Flexible Modality Learning} \label{sec:missing_flexible}
Missing and flexible modality learning aims to maintain effective inference when only a subset of modalities is available, requiring computation to adapt to changing input configurations. Early MoE approaches address this by specializing expert pathways for different modality patterns. Task-Conditional MoE~\cite{novosad2024task} treats modality configurations as related tasks with different parameter subsets, reducing interference between complete and incomplete inputs. FuseMoE~\cite{han2024fusemoe} uses specialized gating to integrate varying numbers of modalities and handle irregular, sparse observations. Flex-MoE~\cite{yun2024flex} makes availability explicit in routing by combining a general router learned from complete observations with specialized routing for observed-modality combinations. Together, these methods mark a shift from generic modality dropout toward routing adapted to the available modalities. Later work extends this configuration-aware routing to recovering and assessing information lost through missingness. MCMoE~\cite{xu2026mcmoe} reconstructs missing representations with an adaptive gated modality generator and combines modality experts for unimodal and cross-modal fusion. ConfSMoE~\cite{zheng2025rethinking} adds confidence-guided routing and two-stage imputation, allowing expert selection to reflect prediction reliability as well as modality identity. This moves adaptation toward estimating which expertise remains reliable and useful under incomplete inputs. MARS~\cite{baek2026mars} further shifts specialization from predefined modality combinations to the representational effects of missingness: it uses the residual between complete- and incomplete-modality representations to guide expert specialization and learns a deployable router for incomplete inputs. Overall, missing-modality MoE progresses from configuration-aware routing, through reconstruction and reliability-aware fusion, toward specialization based on how missing evidence changes the learned representation.

\subsection{Modality Imbalance and Modality Distribution Shift} \label{sec:modality_imbalance_shift}
Unequal modality contributions can allow dominant modalities to suppress weaker ones, while differences in token quantity, signal quality, and data distributions can further bias routing and fusion. MOON2.0~\cite{nie2026moon} adapts expert computation to each input's modality composition. Yin et al.~\cite{yin2026imbalanced} show that imbalance can also occur within a modality, and use common and private experts to disentangle modality-invariant and modality-specific information and regulate their contributions. MiSTER-E~\cite{dutta2026mixture} moves balancing to the sample level by dynamically weighting speech, text, and multimodal experts, allowing a strong unimodal expert to dominate when cross-modal evidence is less informative. These methods shift modality balancing from global compensation toward allocation based on input composition and sample-level informativeness. Imbalance can also arise from unequal token populations, which bias expert utilization independently of representation quality. CoPRIME~\cite{naderi2026coprime} addresses this routing-level imbalance with probabilistic routing and ELBO- and entropy-based regularization, encouraging stable specialization without requiring uniform expert use. More recent methods adapt routing to changing modality reliability and distribution. MIME~\cite{zhang2026mime} uses text as a relatively reliable semantic anchor to suppress audio or visual experts whose outputs are corrupted or inconsistent with it. QA-MoE~\cite{zhu2026qamoe} extends reliability-aware routing to a continuous spectrum of modality quality and missingness, using uncertainty estimates to guide expert selection. Causal-MoE~\cite{lan2026causalmoe} further accounts for latent environmental, subject-specific, and sensor-related confounders, combining perturbation-based backdoor adjustment with expert-specific fusion to improve robustness to distribution shifts. Overall, this research progresses from balancing modality contributions, to accounting for unequal token distributions, and then to reliability- and shift-aware specialization that adapts computation to changing input conditions.

\subsection{Lifelong Multimodal Experts}
\label{sec:lifelong_experts}

Multimodal models deployed in dynamic environments must continually acquire new capabilities while preserving previously learned knowledge. MoE provides a natural architecture for such adaptation because knowledge can be localized into modular experts whose parameters and routing behaviors evolve selectively. Existing studies reveal three increasingly important questions: when new expertise should be introduced, how previously acquired expertise should be preserved and reused, and how expert knowledge should be updated throughout its lifecycle.

\subsubsection{Expert Expansion and Knowledge Isolation.}
Early lifelong MoE approaches primarily use expert modularity to isolate newly acquired knowledge from existing capabilities. MoE-Adapters~\cite{yu2024moeadapters} and MoE-Adapters++~\cite{yu2025moeadapters} dynamically introduce expert adapters as tasks evolve, while CoIN~\cite{chen2024coin} uses multiple LoRA experts to preserve instruction-specific adaptation during continual multimodal instruction tuning. Subsequent methods make expert allocation more adaptive: CL-MoE~\cite{huai2025clmoe} and D-MoLE~\cite{ge2025dmole} determine expert usage according to task-, instance-, or layer-specific adaptation demands, while LLaVA-CMoE~\cite{zhao2025llavacmoe} introduces new expert capacity only when the existing specialization is insufficient. BranchLoRA~\cite{zhang2025branchlora} and ProgLoRA~\cite{yu2025proglora} further organize expandable LoRA experts into reusable branches or progressive expert pools, enabling new tasks to selectively introduce and reuse adaptation capacity. Similar principles extend beyond generic multimodal instruction tuning. SMoLoRA~\cite{wang2025smolora} separates expertise associated with visual understanding and instruction following, while Incre-MoE~\cite{cheng2026incremoe} adaptively expands experts as new detection domains or classes emerge. Chee et al.~\cite{chee2025multimodal} place an MoE structure within cross-modality adapters to continually specialize different multimodal interactions, whereas Zhao et al.~\cite{zhao2026taskshared} combine task-shared and task-specific experts to preserve transferable and task-dependent knowledge in generative multimodal models. Collectively, these methods shift lifelong multimodal MoE from static task isolation toward selectively structuring expert capacity according to evolving tasks, modalities, interactions, and domains.

\subsubsection{Routing Stability and Scalable Expert Reuse.}
Expanding experts alone does not guarantee knowledge preservation because changes in routing may redirect previously learned inputs toward inappropriate experts. Recent work therefore moves from parameter preservation toward functional preservation. DyMoE~\cite{zhao2026dymoe}, SAME~\cite{xie2026same}, and PASs-MoE~\cite{hou2026passmoe} characterize continual degradation through router drift, expert drift, or their coupled evolution, and constrain expert updates and routing pathways to maintain alignment between stored capabilities and their activation patterns. Other approaches reduce dependence on fixed task identities through prototype-, confidence-, or probabilistic expert selection, including Pan et al.~\cite{taskadaptive2026vlm}, MoE-Confidence~\cite{alforhad2026moeconfidence}, and Calibrated Bayesian MoE~\cite{mozaffari2026bayesianmoe}. A related challenge is uncontrolled capacity growth when new experts are continually added. CRAM~\cite{tang2026cram}, Rank-1 Expert Pool~\cite{fa2026rank1}, and MoBLoRA~\cite{chen2026moblora} instead emphasize selective reuse and compositional expertise, allocating new capacity only when existing experts are insufficient or reconstructing task-specific adaptations from reusable expert components. Lifelong MoE therefore increasingly requires a joint decision over whether new knowledge should instantiate additional capacity, reuse existing expertise, or be represented through a composition of previously learned components.

\subsubsection{Toward Lifecycle Management of Expertise.}
Recent work further extends expert modularity from task acquisition toward persistent knowledge maintenance. LiveEdit~\cite{chen2025liveedit} and Multi-MELO~\cite{chen2025multimelo} represent individual knowledge updates as dynamically retrieved low-rank experts, while TIME~\cite{xu2026time} introduces more compact factorized editing experts to reduce the capacity growth caused by accumulating edits. In this view, experts no longer correspond only to tasks or modalities, but can also act as modular memory units that are selectively created and retrieved as knowledge changes. The opposite operation remains substantially less explored in multimodal MoE. Studies in language-only MoE suggest that causal or attribution-based localization can identify which experts actually store target knowledge before performing selective modification. SEUF~\cite{zhuang2025seuf} and EXACT~\cite{wang2027exact} provide representative examples of this expert-level unlearning direction. Extending this principle to multimodal models would enable expert lifecycle management in which knowledge can be introduced, reused, updated, consolidated, and eventually retired without globally modifying the model. Overall, lifelong multimodal MoE is evolving from task-wise parameter isolation toward the management of persistent expert capabilities. The central challenge is no longer simply how to add experts for new tasks, but how to coordinate expert creation, routing stability, capacity reuse, and knowledge maintenance as multimodal knowledge evolves over time.

\subsection{Discussion}

\textbf{Impact of Adaptive MoE Designs on Multimodal Learning.}
Different adaptive MoE designs address distinct forms of variation in multimodal learning. Missing- and flexible-modality methods adapt expert computation to the modalities observed in each input, allowing different modality subsets to use suitable expert pathways and, when supported by reconstruction, to recover information associated with missing modalities. These methods are useful when the available input modalities vary across samples \cite{novosad2024task,han2024fusemoe,yun2024flex,xu2026mcmoe,zheng2025rethinking,baek2026mars}. Even when all modalities are present, their contributions, token distributions, signal quality, or cross-modal relationships may be uneven or change over time. Modality-imbalance and distribution-shift methods adapt expert utilization to modality composition, relative informativeness, uncertainty, and distributional change, helping models make more effective use of weak, noisy, or shifted inputs \cite{nie2026moon,yin2026imbalanced,naderi2026coprime,zhang2026mime,zhu2026qamoe,lan2026causalmoe}. Lifelong multimodal MoE extends adaptation from the current input to the expert system itself: it can add capacity for new tasks or domains, preserve previously learned pathways, reuse or compose existing expertise, and update stored knowledge as the model evolves. These mechanisms support continued learning while helping retain useful capabilities across changing tasks, modalities, and domains \cite{yu2024moeadapters,chen2024coin,huai2025clmoe,zhang2025branchlora,wang2025smolora,zhao2026dymoe,xie2026same,tang2026cram,chen2025liveedit,xu2026time}. Together, these approaches support multimodal robustness at different levels: input-level adaptation determines which evidence can be incorporated, reliability- and shift-aware adaptation adjusts how that evidence contributes, and lifelong adaptation maintains the expert capacity needed over time.

\textbf{Multimodal-Specific Adaptive MoE Design Considerations.}
These adaptive designs reveal several properties that distinguish multimodal MoE from models that adapt to a single source of input variation. (1) Modality availability defines a combinatorial space of possible inputs, and expert specialization can reflect not only observed modality subsets but also the representational effects of missing evidence \cite{novosad2024task,han2024fusemoe,yun2024flex,xu2026mcmoe,zheng2025rethinking,baek2026mars}. (2) Modality presence and modality usefulness are distinct: an available source may still be unreliable or weak for a particular input, making expert activation sensitive to confidence, signal quality, or task-specific contribution as well as availability \cite{zheng2025rethinking,zhang2026mime,zhu2026qamoe,lan2026causalmoe}. (3) Modality imbalance can arise from both unequal information contributions and unequal token populations. Expert routing therefore governs not only which evidence is emphasized, but also how computation is distributed across modalities with different token demands \cite{nie2026moon,yin2026imbalanced,naderi2026coprime}. (4) Distribution shift can reflect changes in environmental, subject-specific, or sensor conditions, extending adaptive routing beyond responses to explicit corruption or missingness \cite{zhang2026mime,zhu2026qamoe,lan2026causalmoe}. (5) Lifelong adaptation makes the expert pool and its routing pathways part of the adaptation process: expertise may be expanded, reused, composed, or updated as tasks and domains evolve, while previously learned capabilities depend on maintaining their functional routing paths \cite{yu2024moeadapters,chen2024coin,huai2025clmoe,zhao2026dymoe,xie2026same,tang2026cram,fa2026rank1,chen2026moblora}. The central design problem is therefore how to coordinate adaptation to available inputs, changing evidence quality, modality-dependent computation, distributional variation, and evolving expertise within a single multimodal expert system.

\begin{figure*}[ht]
\centering
\begin{forest}
  for tree={
    forked edges,
    grow=east,
    reversed=true,
    align=left,
    anchor=west,
    parent anchor=east,
    child anchor=west,
    base=middle,
    font=\tiny,
    rectangle,
    line width=0.5pt,
    rounded corners,
    minimum width=2em,
    minimum height=1em,
    s sep=2pt,
    inner xsep=7pt,
    inner ysep=1pt,
    text ragged,
  },
  [ApplicationsMultimodal Mixture of Expert, middle,rotate=90,anchor=north
        [Healthcare, middle, text width=5em, align=left,  minimum height=1em
            [Adaptive \\ Clinical Models, middle, text width=5em, align=left, minimum height=1em
                [{Med-MoE\cite{jiang2024med}, MedMoE\cite{chopra2025medmoe}, M$^{4}$oE\cite{jiang2024m4oe}, MOME\cite{luo2025breastmome}, CoMed-TR\citep{chen2026comedtr}, M4oE\cite{wu2025m4oe},\\ Abdullakutty et al.\cite{abdullakutty2026breastmoe}, Liao et al.\cite{liao2026mgmtmoe}, NeuroMoE\cite{raza2025neuromoe}}, leaf, align=left, text width=27em]
            ]
            [Computational \\ Pathology, middle, text width=5em, align=left,  minimum height=1em
            [{Hashimoto et al.\cite{hashimoto2024lymphoma}, MoME\cite{xiong2024mome}, SurMoE.\cite{zhang2026surmoe}, Saueressig et al.\cite{saueressig2026glioma}, scMM\cite{minoura2021scmm}}, leaf, align=left, text width=27em]
            ]
        ]
        [Environment \\ Science, middle, text width=5em, align=left,  minimum height=1em
            [Multisensor Earth \\ Observation, middle, text width=5em, align=left,  minimum height=1em
            [{RingMoE\cite{bi2026ringmoe}, SM3Det\citep{li2026sm3det}, RS-MoE\cite{lin2025rsmoe}, SkyMoE\citep{liu2026skymoe},dual-MoE\cite{kong2025dualmoe}, MixtureRS\citep{liu2025mixturers},\\ MultiMoE\citep{qi2026multimoe}}, leaf, align=left, text width=27em]
            ]
            [Environment\\ Science, middle, text width=5em, align=left,  minimum height=1em
            [{PMFN\cite{liu2025priorenhanced}, M$^{2}$CD\cite{liu2025m2cd}, DMoE-AttU-Net\cite{khankeshizadeh2026dmoe}, UniRoute\cite{shu2026uniroute}, MaMOL\cite{gao2025mamol}, DAMoE\cite{liu2026damoe}, \\ LITE\cite{li2024lite}, Alizadehsani et al.\cite{alizadehsani2026ecological}, Pest-MoE\cite{li2026pestmoe}, Incre-MoE\cite{cheng2026incremoe}, BuildFunc-MoE\cite{wang2026buildfunc}}, leaf, align=left, text width=27em]
            ]
        ]
        [Industrial System, middle, text width=5em, align=left,  minimum height=1em
            [Anomaly \\ Detection, middle, text width=5em, align=left,  minimum height=1em
            [{Lin et al.\cite{lin2026multisensorfault}, Guo et al.\cite{guo2026qwindustry},  Willibald et al.\cite{willibald2025anomaly}, Li et al. \cite{li2026textguided}}, leaf, align=left, text width=27em]
            ]
            [Autonomous \\ Machine, middle, text width=5em, align=left,  minimum height=1em
            [{MoME\cite{park2025mome}, PRISM-Occ\citep{zhang2026prismocc}, AW-MoE\citep{lin2026awmoe}, DriveMoE\citep{yang2026drivemoe}, ForceVLA\cite{yu2025forcevla}, \\ ForceVLA2\cite{li2026forcevla2}, MoDE-VLA\cite{tang2026modevla}}, leaf, align=left, text width=27em]
            ]
        ]
        [Human Centered \\ Intelligence, middle, text width=5em, align=left,  minimum height=1em
            [Communication \\ Understanding, middle, text width=5em, align=left,  minimum height=1em
                [{MoHAVE\cite{kim2025mohave}, MoME\cite{cappellazzo2025momeavsr},MiSTER-E\cite{dutta2026mixture}, EUAR\cite{gao2024euar}, QA-MoE\cite{zhu2026qamoe}, C$^{2}$MOE\cite{shou2026c2moe},\\ PAMoE-MSA\citep{huang2025pamoe}, MIME\cite{zhang2026mime}, Wang et al. \cite{wang2026clipmoe}}, leaf, align=left, text width=27em]
            ]
            [Human \\ Recognition, middle, text width=5em, align=left,  minimum height=1em
            [{QME\cite{zhu2025quality}, SemGait\cite{xi2026semgait}, FlexiReID\cite{sun2025flexireid}, MFRNet\cite{feng2025mfrnet}, DeMo\cite{wang2025demo}, MoSCE-ReID\cite{ren2025mosce}, \\ UGG-ReID\citep{wan2025ugg}, MCMoE\cite{xu2026mcmoe}, OneTrackerV2\citep{hong2026dualmoe}}, leaf, align=left, text width=27em]
            ]
        ]
        [Information\\ Retrieval, middle, text width=5em, align=left,  minimum height=1em
            [Structural \\ Information \\ Understanding, middle, text width=5em, align=left,  minimum height=1em
                [{MoMoK\cite{zhang2024multiple}, MMoE\cite{hu2025multilevel}, REMOTE\cite{lin2025remote}, MMCas\cite{jing2026modeling}, MIMoE-FND\cite{liu2025mimoefnd}, ChartMoE\cite{xu2025chartmoe}, \\ DocRouter\cite{zhang2025docrouter}, visual-prompt MoE\cite{xu2023visualpromptmoe}}, leaf, align=left, text width=27em]
            ]
            [Information \\ Retrieval, middle, text width=5em, align=left,  minimum height=1em
                [{AMC\citep{zhu2023amc}, MMES\citep{zhai2025mmes}, ADaFuSE\citep{zhang2026adafuse}, DA-MoE\citep{li2026damoecrossmodal}, MOON2.0\cite{nie2026moon}, M3SRec\citep{bian2023m3srec},\\ HM4SR\citep{zhang2025hm4sr}, M3Rec\citep{guo2025m3rec}, XSMoE\citep{qu2025xsmoe}, MaMoE4Rec\citep{zheng2026mamoe4rec}, CAMMSR\citep{xu2026cammsr}}, leaf, align=left, text width=27em
                ]
            ]
        ]
        [Other Emerging \\ Domains, middle, text width=5em, align=left,  minimum height=1em
            [{MAMoE\cite{liang2026mamoe}, ISAC\citep{zhang2026isac},CaSemCom\cite{liu2025casemcom},TiMi\cite{lin2026timi}, Expert Modulation \citep{zhang2026modulated}, MSE-ITT\cite{koval2025mseitt}, LLMoE\citep{liu2025llmoetrading},\\ MoL-MoE\citep{soares2025molmoe}, MDM-DTA\cite{dai2026mdmdta}, MF-DTA\cite{kang2025mfdta}, MMPG\cite{wang2026mmpg}, EssLM-MoE\cite{zeng2026esslmmoe}, MERA\cite{wu2026mera}}, leaf, align=left, text width=33.8em
            ]
        ]
  ]
\end{forest}
\caption{Taxonomy of application domains for multimodal Mixture-of-Experts. The taxonomy groups methods by domain and task, covering healthcare and life sciences, Earth and environmental science, industrial and autonomous systems, human-centered intelligence, information and retrieval systems, and emerging applications in communication, forecasting, and scientific prediction.}
\label{fig:taxonomy_of_applications}
\end{figure*}

\section{Applications of Multimodal Mixture-of-Experts}
\label{sec:application}

While previous sections examine multimodal MoE from the perspectives of efficiency, expert specialization, interaction, and adaptation, these developments are ultimately driven by application-specific requirements. Real-world multimodal systems combine information sources with different semantics, acquisition processes, reliability, and operational constraints, motivating experts to specialize around domain-specific structures and routers to dynamically select the most relevant information or capability. This section revisits multimodal MoE from an application perspective, covering healthcare and life sciences, Earth and environmental intelligence, industrial and autonomous systems, human-centered intelligence, information and document systems, and other emerging domains. Figure~\ref{fig:taxonomy_of_applications} maps the application areas and representative multimodal MoE methods reviewed in this section.

\subsection{Healthcare and Life Sciences}
\label{sec:app_healthcare}

Healthcare decisions often integrate medical images, physiological signals, clinical records, and molecular measurements. However, these modalities may be incomplete, noisy, or differently informative across patients and tasks. Multimodal MoE addresses this heterogeneity by preserving modality-, domain-, or task-specific knowledge in different experts and dynamically routing each case to the most relevant expertise.

\subsubsection{Unified and Adaptive Clinical Models.}
Multimodal MoE enables medical models to share a unified backbone while preserving specialized computation across domains, tasks, patients, and clinical conditions. Med-MoE\cite{jiang2024med} and MedMoE\cite{chopra2025medmoe} organize specialized experts across medical domains and instructions, while M$^{4}$oE\cite{jiang2024m4oe} and MOME\cite{luo2025breastmome} extend this principle to multimodal medical imaging and multi-task prediction. In medical VQA, CoMed-TR\citep{chen2026comedtr} routes questions between large medical vision--language models and task-specific small models. Beyond task-level specialization, M4oE\cite{wu2025m4oe} jointly models patient variation, modality heterogeneity, and task dependencies, while disease-specific approaches\cite{abdullakutty2026breastmoe,liao2026mgmtmoe,raza2025neuromoe} dynamically combine pathology, structured clinical data, and different MRI sequences for breast cancer, glioblastoma, and neurological-disorder prediction. Collectively, these methods show that medical MoE is evolving from domain- and task-specific expert allocation toward more patient- and condition-adaptive computation within unified clinical models.

\subsubsection{Computational Pathology and Multi-Omics Integration.} MoE models combine whole-slide images with flow cytometry for lymphoma classification, or route between pathology, radiology, and genomic representations for cancer survival and glioma subtyping \citep{hashimoto2024lymphoma,xiong2024mome,zhang2026surmoe,saueressig2026glioma}. At the molecular level, scMM uses multimodal experts to integrate single-cell omics and support cross-modal generation \citep{minoura2021scmm}. MoE therefore coordinates evidence describing disease at different spatial and biological scales.

\subsection{Earth and Environment Science}
\label{sec:app_earth}

Earth and environmental intelligence combines heterogeneous observations from optical, SAR, LiDAR, hyperspectral or multispectral imagery, together with geographic and climate records. These sources capture different physical properties and vary in reliability with clouds, illumination, terrain, season, and sensor availability. Multimodal MoE is therefore particularly suitable for this domain because expert specialization can preserve sensor-specific representations, while routing adapts computation to task requirements and changing observation conditions.

\subsubsection{Multisensor Earth Observation.}
A major role of MoE in Earth observation is to unify heterogeneous sensors and tasks without forcing them to share identical representations. RingMoE\cite{bi2026ringmoe} organizes modality-specialized, collaborative, and shared experts across multiple remote-sensing sensors and tasks, while SM3Det\citep{li2026sm3det} uses sparse experts to unify object detection across heterogeneous sensor modalities and annotation formats. RS-MoE\cite{lin2025rsmoe} and SkyMoE\citep{liu2026skymoe} further extend expert specialization to image captioning, visual question answering, and geospatial interpretation through task- and instruction-aware routing. At the perception level, dual-MoE\cite{kong2025dualmoe} and MixtureRS\citep{liu2025mixturers} select complementary spectral, spatial, and elevation fusion strategies for hyperspectral--LiDAR learning, while MultiMoE\citep{qi2026multimoe} introduces lightweight modality-aware experts for semantic segmentation. These methods show how MoE can combine broad capability sharing with sensor- and task-specific computation across heterogeneous Earth-observation settings.

\subsubsection{Environmental Decision Support.}
Beyond unifying sensors, MoE increasingly adapts expert allocation to observation quality, sensing conditions, and downstream context. Prior-enhanced MoE dynamically weights fusion experts according to environmental conditions and multimodal reliability \cite{liu2025priorenhanced}, while M$^{2}$CD\cite{liu2025m2cd}, DMoE-AttU-Net\cite{khankeshizadeh2026dmoe}, and UniRoute\cite{shu2026uniroute} specialize feature extraction or difference operators for heterogeneous optical--SAR change detection. MaMOL\cite{gao2025mamol} routes different missing-sensor patterns through shared, modality-specific, and pattern-specific LoRA experts, whereas DAMoE\cite{liu2026damoe} uses sensing priors to select experts for pansharpening, super-resolution, and SAR-assisted optical reconstruction. Similar principles extend to environmental decision support: LITE\cite{li2024lite} addresses incomplete environmental variables; multimodal experts incorporate seasonal knowledge or text for ecological and agricultural monitoring\cite{alizadehsani2026ecological,li2026pestmoe}; Incre-MoE\cite{cheng2026incremoe} expands expertise as forestry domains and object classes evolve; and BuildFunc-MoE\cite{wang2026buildfunc} integrates imagery, night-time light, DEM, and POI information for urban-function mapping. Collectively, these applications shift Earth-observation MoE from static multisensor fusion toward condition-aware sensing, where routing determines which observations, processing strategies, and expert capacities remain most informative under the current environment and task.

\subsection{Industrial and Autonomous Systems}
\label{sec:app_industry}

Industrial and autonomous systems integrate heterogeneous sensors that capture machine states, surrounding environments, and physical interactions. Their usefulness varies with operating conditions: faults may appear differently across vibration and acoustic signals, camera and LiDAR reliability may change with weather or spatial context, and force or tactile feedback becomes critical during contact. Multimodal MoE is therefore well suited to these settings because expert routing can adapt both sensor fusion and task-specific computation to the current physical state.

\subsubsection{Industrial Diagnosis and Anomaly Monitoring.}
In industrial monitoring, MoE enables specialized processing of heterogeneous sensing signals while retaining cross-sensor reasoning. For rotating-machinery diagnosis, multisensor MoE routes between experts for individual signals and their interactions, while knowledge-guided experts support defect detection and fault reasoning across heterogeneous industrial inputs \citep{lin2026multisensorfault,guo2026qwindustry}. In anomaly detection and inspection, expert fusion combines visual observations with robot force and motion signals, while text-guided top-$K$ MoE integrates RGB appearance and 3D geometry \citep{willibald2025anomaly,li2026textguided}. These methods use routing to determine whether local sensor evidence, cross-modal interactions, or contextual knowledge is most informative for diagnosing the current machine or operational state.

\subsubsection{Condition-Adaptive Perception and Action.}
In autonomous systems, MoE increasingly extends from sensor fusion toward condition-dependent perception and control. MoME\cite{park2025mome} selects camera, LiDAR, or joint detection experts for individual object queries, while PRISM-Occ\citep{zhang2026prismocc} routes experts within and across modalities for BEV occupancy prediction. AW-MoE\citep{lin2026awmoe} further uses image-guided weather information to activate specialized LiDAR--radar detection experts, making perception responsive to local sensing reliability. This adaptive specialization also extends to action generation: DriveMoE\citep{yang2026drivemoe} routes between camera views and driving-behavior experts, while ForceVLA\cite{yu2025forcevla} and ForceVLA2\cite{li2026forcevla2} incorporate force-aware experts into vision--language--action models, and MoDE-VLA\cite{tang2026modevla} additionally integrates tactile and force inputs for dexterous behavior. Collectively, these systems show a progression from multimodal perception toward condition-aware allocation of physical expertise, where routing determines which observations and capabilities should influence diagnosis, perception, or action under the current operating state.

\subsection{Human-Centered Multimodal Intelligence}
\label{sec:app_human}

Human-centered applications integrate speech, language, visual behavior, biometric cues, and other signals to understand communication, affect, identity, and action. These signals can vary substantially across individuals and environments. Multimodal MoE is therefore useful for dynamically adjusting which evidence and specialized interpretation should be trusted for each sample.

\subsubsection{Communication and Affective Understanding.}
In speech and affective understanding, MoE adapts multimodal processing to the reliability and semantic relevance of human signals. MoHAVE\cite{kim2025mohave} adjusts the contribution of audio and visual expert groups according to modality reliability, while MoME\cite{cappellazzo2025momeavsr} employs sparse experts across different audio--visual representation scales. For emotion and sentiment understanding, MiSTER-E\cite{dutta2026mixture} combines speech, text, and joint experts, while EUAR\cite{gao2024euar} and QA-MoE\cite{zhu2026qamoe} incorporate uncertainty or signal quality into routing. C$^{2}$MOE\cite{shou2026c2moe} further addresses incomplete observations by modeling consistent and complementary multimodal evidence, while other approaches \citep{huang2025pamoe,zhang2026mime,wang2026clipmoe} specialize experts according to sentiment polarity, semantic representation, or individual input characteristics. Collectively, these methods replace a fixed balance among human signals with sample-dependent routing that determines which cues and cross-modal interactions are most informative.

\subsubsection{Adaptive Human Recognition and Tracking.}
MoE also supports human recognition by adapting expert selection to biometric quality, modality availability, and semantic context. QME\cite{zhu2025quality} dynamically fuses face, gait, and body recognition according to signal quality, while SemGait\cite{xi2026semgait} combines visual, textual, and gait-attribute information through attribute-driven experts. In person re-identification, FlexiReID\cite{sun2025flexireid} supports flexible retrieval across RGB, infrared, sketch, and text inputs, whereas MFRNet\cite{feng2025mfrnet} and DeMo\cite{wang2025demo} balance shared and modality-specific identity representations through expert specialization. MoSCE-ReID\cite{ren2025mosce} introduces attribute-group LoRA experts for text-guided visual recognition, while UGG-ReID\citep{wan2025ugg} uses uncertainty-aware routing to strengthen multimodal interaction under local noise. Similar adaptive mechanisms extend to incomplete action-quality assessment through MCMoE\cite{xu2026mcmoe} and to temporal and RGB+X tracking through OneTrackerV2\citep{hong2026dualmoe}. These applications show that human-centered MoE increasingly treats expert routing as a mechanism for adapting interpretation to the reliability, availability, and semantic relevance of person-specific evidence.

\subsection{Information, Document, and Online Systems}
\label{sec:app_information}

Information-centric applications combine visual and textual content with relational structure, document layout, propagation patterns, retrieval context, and user behavior. In these settings, expert specialization often depends less on modality identity itself than on the semantic role that each source plays in the current task. Multimodal MoE therefore provides a mechanism for routing information according to the corresponding semantic source.

\subsubsection{Structural Information Understanding.}
MoE is increasingly used to specialize computation according to the structural role and interaction pattern of multimodal information. For multimodal knowledge graphs and entity linking, MoMoK\cite{zhang2024multiple} and MMoE\cite{hu2025multilevel} select expertise according to entity relations and cross-modal context, while REMOTE\cite{lin2025remote} routes among intra- and inter-modal interactions for relation extraction. MoRe\citep{wang2022multimodalretrieval} further combines text- and image-based retrieval with expert prediction for multimodal named-entity and relation extraction. Similar interaction-aware specialization appears in online content modeling: MMCas\cite{jing2026modeling} assigns experts to unique, redundant, and synergistic multimodal information for cascade prediction, while MIMoE-FND\cite{liu2025mimoefnd} models image--text interactions for fake-news detection. In structured documents, ChartMoE\cite{xu2025chartmoe} and DocRouter\cite{zhang2025docrouter} use expert connectors to adapt visual representations to chart and document structure, while visual-prompt MoE\cite{xu2023visualpromptmoe} specializes multimodal information extraction. MoRE\citep{peng2026retrievalexperts} further routes among text, image, and table retrieval experts under reasoning guidance. Collectively, these methods shift expert specialization from modality identity toward relation-, interaction-, and structure-specific processing.

\subsubsection{Information Retrieval}
In retrieval and recommendation, the appropriate expert depends on query composition, item evidence, and evolving user preference. AMC\citep{zhu2023amc} adapts experts for text-guided image retrieval, MMES\citep{zhai2025mmes} handles composed queries combining a reference image with modification text, and ADaFuSE\citep{zhang2026adafuse} supports interactive image--text retrieval, while DA-MoE\citep{li2026damoecrossmodal} uses heterogeneous experts for image--text alignment and retrieval. For recommendation, MOON2.0\cite{nie2026moon}, M3SRec\citep{bian2023m3srec}, HM4SR\citep{zhang2025hm4sr}, and M3Rec\citep{guo2025m3rec} model multimodal item features and sequential preferences, whereas XSMoE\citep{qu2025xsmoe}, MaMoE4Rec\citep{zheng2026mamoe4rec}, and CAMMSR\citep{xu2026cammsr} further adapt expert allocation to evolving feedback, graph structure, or category-specific preference. These systems illustrate a broader role of MoE as a context-dependent information selector, where routing determines not only which modality contributes, but which representation, retrieval source, or preference pattern is most relevant to the current information need.

\subsection{Emerging Domain-Specific Applications}
\label{sec:app_emerging}
In networked sensing and communication, MAMoE\cite{liang2026mamoe} coordinates expertise for 3D reconstruction and transmission, while ISAC \citep{zhang2026isac} weights visual, radar, LiDAR, and positional experts according to their reliability. CaSemCom\cite{liu2025casemcom} routes high-impact semantic features for wireless transmission, and CoGMoE\cite{li2026cogmoe} coordinates graph-based experts across camera and LiDAR observations. For multimodal forecasting, TiMi\cite{lin2026timi} and Expert Modulation\citep{zhang2026modulated} use textual context to guide time-series modeling, while MSE-ITT\cite{koval2025mseitt} and LLMoE\citep{liu2025llmoetrading} route experts across financial signals and news. Similar mechanisms appear in environmental prediction, where multimodal solar forecasting\cite{erni2026solar} adapt computation to changing observations and source relevance. In scientific prediction, MoL-MoE\citep{soares2025molmoe} combines SMILES, SELFIES, and graph views, while MDM-DTA\cite{dai2026mdmdta} and MF-DTA\cite{kang2025mfdta} use expert fusion for drug--target affinity prediction. MMPG\cite{wang2026mmpg}, EssLM-MoE\cite{zeng2026esslmmoe}, and MERA\cite{wu2026mera} further specialize across physical, chemical, geometric, sequence, structural, and functional protein representations.

\subsection{Reflections from Applications}

Comparing healthcare, Earth observation, industrial monitoring, and autonomous systems reveals four patterns in how MoE expertise is organized and used. (1) The target of expert specialization differs across application domains. Healthcare models commonly organize experts around medical domains, prediction tasks, or patient variation, while Earth-observation models combine sensor-specific expertise with shared experts for cross-sensor and task-level processing \cite{jiang2024med,wu2025m4oe,chen2026comedtr,bi2026ringmoe,lin2025rsmoe}. Industrial and autonomous systems more often specialize around individual signals, sensor interactions, operating states, or physical capabilities such as force-aware action \cite{lin2026multisensorfault,park2025mome,yu2025forcevla,tang2026modevla}. In information-centric applications, specialization can instead follow relations, document structure, or query composition \cite{zhang2024multiple,xu2025chartmoe,peng2026retrievalexperts}. (2) The conditions that trigger adaptive routing also differ by domain. Patient and modality variation can change which clinical evidence is informative; sensor availability, weather, or observation quality can change the value of Earth-observation inputs; and machine state or physical contact can alter which signals matter to industrial diagnosis or control \cite{wu2025m4oe,xu2026mcmoe,liu2025priorenhanced,gao2025mamol,lin2026awmoe,yu2025forcevla}. (3) MoE contributes at different points in the application workflow. In healthcare and remote sensing, it often combines heterogeneous evidence while selecting expertise relevant to a diagnosis, prediction, or sensing task. In industrial and autonomous systems, expert selection can also condition downstream perception or action on the current operating state \cite{jiang2024m4oe,bi2026ringmoe,liu2025priorenhanced,park2025mome,tang2026modevla}. (4) The extent to which MoE designs transfer across domains depends on the structure of the data and task. Shared capacity combined with specialized experts and context-dependent routing recurs across clinical, sensing, and autonomous applications. The routing signals and expertise assigned to individual experts, however, reflect domain-specific needs such as patient-level clinical evidence, sensor-specific physical properties, or state-dependent interactions in robotic systems. Structure-aware routing in document and retrieval systems further shows how expert identity can follow semantic relations or query structure \cite{wu2025m4oe,bi2026ringmoe,lin2026awmoe,zhang2025docrouter,zheng2026mamoe4rec}. Together, these comparisons suggest that multimodal MoE provides a reusable pattern for allocating specialized computation, while the specialization targets and routing signals remain closely tied to each application's data structure and operational needs.

\section{Open Challenges} 
\label{sec:challenges}

\textbf{Principled and interpretable expert routing.}
Multimodal inputs differ substantially in representation complexity, redundancy, token quantity, and information content~\cite{luo2024gamma}. Existing approaches have already moved beyond purely fixed routing: Gamma-MoD~\cite{luo2024gamma} and UniMoD~\cite{mao2025unimod}, for example, adapt computation according to input complexity or task demand. However, these mechanisms still provide limited explanation of why a particular input requires a particular amount of expert capacity. More generally, routing scores do not necessarily correspond to the marginal utility of activating an additional expert. An important open problem is therefore to develop criteria that relate expert activation to measurable quantities such as modality complexity, uncertainty, information gain, or expected performance improvement. Such criteria would make routing both more adaptive and more interpretable than selecting experts solely according to learned gating scores.

\textbf{Temporal modality relevance drift.}
The task-specific importance of a modality may change as the task context, system state, or underlying process evolves. Prior studies have observed time-varying modality reliability, context-dependent modality weighting, and modality contribution drift~\cite{gurban2008dynamic,dehghanghanatkaman2026dynamic,zhang2026regularizing}. Importantly, such changes can occur even when a modality remains available and reliable. Most multimodal MoE models adapt routing according to the current representation, availability, or instantaneous quality of each modality, but do not explicitly track how its predictive utility evolves over a longer trajectory. Future work could therefore incorporate historical or state information into routing and distinguish persistent relevance changes from short-term fluctuations or corruption.

\textbf{Expert capacity allocation, consolidation, and pruning.}
Recent multimodal MoE methods already support expert expansion, reuse, and compositional adaptation, reducing the need to allocate a completely new expert for every task~\cite{tang2026cram,fa2026rank1,chen2026moblora,xu2026time}. However, these mechanisms do not fully answer how much expert capacity different modalities, layers, or interaction functions actually require. Multimodal models may contain experts with highly unequal utility and substantial functional overlap, while rarely activated experts may still preserve important specialized knowledge. Expert management therefore requires criteria that go beyond activation frequency and determine when expertise should be expanded, reused, merged, compressed, or removed according to its functional contribution.

\textbf{Cross-expert communication and conflict resolution.}
Existing multimodal MoE models increasingly specialize experts for modality-specific information, shared representations, and different forms of cross-modal interaction~\cite{emnlp/YuQJSML24,shen2024mome}. However, interaction-aware experts and expert-to-expert communication are not equivalent. Most models still coordinate independently specialized experts primarily through router weights or output aggregation. When experts provide complementary or conflicting evidence, it remains unclear when they should exchange information, what information should be communicated, and how disagreement should be resolved without eliminating useful specialization. Future work could therefore investigate explicit cross-expert coordination mechanisms that preserve expert diversity while supporting joint multimodal reasoning.

\textbf{Non-stationary modality dependencies.}
Multimodal performance depends not only on the contribution of individual modalities but also on how modalities interact through redundancy, complementarity, and synergy. Prior work shows that different modalities can generalize and overfit at different rates, allowing dominant modalities to suppress weaker but complementary information~\cite{wang2020makes}. Existing interaction-aware MoE methods can model different or temporally dependent interaction patterns, but longer-term changes in these relationships remain less explored. A modality pair may shift from complementary to redundant or even conflicting as the environment or task evolves. This problem differs from temporal modality relevance drift: the latter concerns the utility of an individual modality, whereas dependency drift concerns how the relationship between modalities changes. Future MoE models should therefore adapt not only expert activation but also the interaction structure represented by the experts.

\section{Conclusion}

This survey reviews the role of Mixture-of-Experts in multimodal learning from three perspectives: efficient computation, expert specialization across representations and modalities, and adaptation to imperfect or evolving multimodal data. Across these directions, MoE is shifting from a sparse scaling mechanism toward a more general framework for allocating specialized computation according to modality structure, context, and changing information demands. A recurring design tension is the balance between specialization and sharing: stronger specialization better preserves heterogeneous modality structures, while greater sharing improves efficiency and knowledge transfer. Future multimodal MoE research should therefore focus on more interpretable and adaptive routing, efficient expert lifecycle management, and mechanisms that can respond to evolving modality relevance and dependencies.

\bibliographystyle{ACM-Reference-Format}
\bibliography{sample-base}

@misc{xu2026multimodalmoe,
  author       = {Cheng Xu and Jiacheng Liu and Xinfeng Xia and Haozhe Wei and
                  Wenfeng Wang and Xiaofeng Hou and Chao Li and Minyi Guo},
  title        = {Multimodal Mixture-of-Experts: From Sparse Scaling to
                  Conditional Specialization},
  year         = {2026},
  howpublished = {Available at SSRN 7245038},
  url          = {https://ssrn.com/abstract=7245038}
}

@article{cui2026remotesensing,
  author  = {Yongchuan Cui and Peng Liu and Lajiao Chen},
  title   = {Mixture-of-Experts in Remote Sensing: A Survey},
  journal = {Journal of Geoscience and Earth Observation},
  volume  = {1},
  number  = {1},
  pages   = {4--38},
  year    = {2026},
  doi     = {10.62762/JGEO.2025.140654},
  url     = {https://doi.org/10.62762/JGEO.2025.140654}
}

@article{xu2024integration,
  author        = {Minrui Xu and Dusit Niyato and Jiawen Kang and Zehui Xiong and
                   Abbas Jamalipour and Yuguang Fang and Dong In Kim and Xuemin Shen},
  title         = {Integration of Mixture of Experts and Multimodal Generative {AI}
                   in Internet of Vehicles: A Survey},
  journal       = {arXiv preprint arXiv:2404.16356},
  year          = {2024},
  eprint        = {2404.16356},
  archiveprefix = {arXiv},
  primaryclass  = {cs.NI},
  url           = {https://arxiv.org/abs/2404.16356}
}

@article{jin2025efficient,
  author  = {Yizhang Jin and Jian Li and Tianjun Gu and Yexin Liu and
             Bo Zhao and Jinxiang Lai and Zhenye Gan and Yabiao Wang and
             Chengjie Wang and Xin Tan and Lizhuang Ma},
  title   = {Efficient Multimodal Large Language Models: A Survey},
  journal = {Visual Intelligence},
  volume  = {3},
  number  = {27},
  year    = {2025},
  doi     = {10.1007/s44267-025-00099-6},
  url     = {https://doi.org/10.1007/s44267-025-00099-6}
}

@article{li2026alignmentfusion,
  author  = {Songtao Li and Hao Tang},
  title   = {Multimodal Alignment and Fusion: A Survey},
  journal = {International Journal of Computer Vision},
  volume  = {134},
  eid     = {103},
  year    = {2026},
  doi     = {10.1007/s11263-025-02667-1},
  url     = {https://doi.org/10.1007/s11263-025-02667-1}
}

@inproceedings{zheng2026tackling,
  author    = {Liangwei Zheng and Wei Emma Zhang and Olaf Maennel and
               Lin Yue and Weitong Chen},
  title     = {Tackling Multimodal Learning Challenges with Mixture-of-Expert:
               A Survey},
  publisher = {Proceedings of the Thirty-Fifth International Joint Conference
               on Artificial Intelligence},
  pages     = {8116--8124},
  year      = {2026},
  doi       = {10.24963/ijcai.2026/900},
  url       = {https://doi.org/10.24963/ijcai.2026/900}
}

@inproceedings{tan2025paths,
  title={Paths-over-graph: Knowledge graph empowered large language model reasoning},
  author={Tan, Xingyu and Wang, Xiaoyang and Liu, Qing and Xu, Xiwei and Yuan, Xin and Zhang, Wenjie},
  booktitle={Proceedings of the ACM on Web Conference 2025},
  pages={3505--3522},
  year={2025}
}

@inproceedings{tan2025hydrarag,
  title={Hydrarag: Structured cross-source enhanced large language model reasoning},
  author={Tan, Xingyu and Wang, Xiaoyang and Liu, Qing and Xu, Xiwei and Yuan, Xin and Zhu, Liming and Zhang, Wenjie},
  booktitle={Proceedings of the 2025 Conference on Empirical Methods in Natural Language Processing},
  pages={14442--14470},
  year={2025}
}

@inproceedings{dai2024deepseekmoe,
  title={Deepseekmoe: Towards ultimate expert specialization in mixture-of-experts language models},
  author={Dai, Damai and Deng, Chengqi and Zhao, Chenggang and Xu, RX and Gao, Huazuo and Chen, Deli and Li, Jiashi and Zeng, Wangding and Yu, Xingkai and Wu, Yu and others},
  booktitle={Proceedings of the 62nd Annual Meeting of the Association for Computational Linguistics (Volume 1: Long Papers)},
  pages={1280--1297},
  year={2024}
}

@article{luo2024moelora,
  title={Moelora: Contrastive learning guided mixture of experts on parameter-efficient fine-tuning for large language models},
  author={Luo, Tongxu and Lei, Jiahe and Lei, Fangyu and Liu, Weihao and He, Shizhu and Zhao, Jun and Liu, Kang},
  journal={arXiv preprint arXiv:2402.12851},
  year={2024}
}

@article{li2025minimax,
  title={Minimax-01: Scaling foundation models with lightning attention},
  author={Li, Aonian and Gong, Bangwei and Yang, Bo and Shan, Boji and Liu, Chang and Zhu, Cheng and Zhang, Chunhao and Guo, Congchao and Chen, Da and Li, Dong and others},
  journal={arXiv preprint arXiv:2501.08313},
  year={2025}
}

@article{xu2025qwen3,
  title={Qwen3-omni technical report},
  author={Xu, Jin and Guo, Zhifang and Hu, Hangrui and Chu, Yunfei and Wang, Xiong and He, Jinzheng and Wang, Yuxuan and Shi, Xian and He, Ting and Zhu, Xinfa and others},
  journal={arXiv preprint arXiv:2509.17765},
  year={2025}
}

@article{yang2025qwen3,
  title={Qwen3 technical report},
  author={Yang, An and Li, Anfeng and Yang, Baosong and Zhang, Beichen and Hui, Binyuan and Zheng, Bo and Yu, Bowen and Gao, Chang and Huang, Chengen and Lv, Chenxu and others},
  journal={arXiv preprint arXiv:2505.09388},
  year={2025}
}

@article{wu2024deepseek,
  title={Deepseek-vl2: Mixture-of-experts vision-language models for advanced multimodal understanding},
  author={Wu, Zhiyu and Chen, Xiaokang and Pan, Zizheng and Liu, Xingchao and Liu, Wen and Dai, Damai and Gao, Huazuo and Ma, Yiyang and Wu, Chengyue and Wang, Bingxuan and others},
  journal={arXiv preprint arXiv:2412.10302},
  year={2024}
}

@article{team2025kimi,
  title={Kimi-vl technical report},
  author={Team, Kimi and Du, Angang and Yin, Bohong and Xing, Bowei and Qu, Bowen and Wang, Bowen and Chen, Cheng and Zhang, Chenlin and Du, Chenzhuang and Wei, Chu and others},
  journal={arXiv preprint arXiv:2504.07491},
  year={2025}
}

@article{team2024gemini,
  title={Gemini 1.5: Unlocking multimodal understanding across millions of tokens of context},
  author={Team, Gemini and Georgiev, Petko and Lei, Ving Ian and Burnell, Ryan and Bai, Libin and Gulati, Anmol and Tanzer, Garrett and Vincent, Damien and Pan, Zhufeng and Wang, Shibo and others},
  journal={arXiv preprint arXiv:2403.05530},
  year={2024}
}

@inproceedings{zhu2024llama,
  title={Llama-moe: Building mixture-of-experts from llama with continual pre-training},
  author={Zhu, Tong and Qu, Xiaoye and Dong, Daize and Ruan, Jiacheng and Tong, Jingqi and He, Conghui and Cheng, Yu},
  booktitle={Proceedings of the 2024 conference on empirical methods in natural language processing},
  pages={15913--15923},
  year={2024}
}

@article{baltruvsaitis2018multimodal,
  title={Multimodal machine learning: A survey and taxonomy},
  author={Baltru{\v{s}}aitis, Tadas and Ahuja, Chaitanya and Morency, Louis-Philippe},
  journal={IEEE transactions on pattern analysis and machine intelligence},
  volume={41},
  number={2},
  pages={423--443},
  year={2018},
  publisher={IEEE}
}

@article{guo2019deep,
  title={Deep multimodal representation learning: A survey},
  author={Guo, Wenzhong and Wang, Jianwen and Wang, Shiping},
  journal={Ieee Access},
  volume={7},
  pages={63373--63394},
  year={2019},
  publisher={IEEE}
}

@article{yuan2025survey,
  title={A survey of multimodal learning: Methods, applications, and future},
  author={Yuan, Yuan and Li, Zhaojian and Zhao, Bin},
  journal={ACM Computing Surveys},
  volume={57},
  number={7},
  pages={1--34},
  year={2025},
  publisher={ACM New York, NY}
}

@article{fedus2022switch,
  title={Switch transformers: Scaling to trillion parameter models with simple and efficient sparsity},
  author={Fedus, William and Zoph, Barret and Shazeer, Noam},
  journal={Journal of Machine Learning Research},
  volume={23},
  number={120},
  pages={1--39},
  year={2022}
}

@inproceedings{zheng2024devil,
  title={Devil in the tail: A multi-modal framework for drug-drug interaction prediction in long tail distinction},
  author={Zheng, Liangwei Nathan and Dong, Chang George and Zhang, Wei Emma and Chen, Xin and Yue, Lin and Chen, Weitong},
  booktitle={Proceedings of the 33rd ACM International Conference on Information and Knowledge Management},
  pages={3395--3404},
  year={2024}
}

@misc{kimi2026k25,
  title={Kimi K2.5: Visual Agentic Intelligence},
  author={{Kimi Team}},
  year={2026},
  eprint={2602.02276},
  archivePrefix={arXiv},
  url={https://arxiv.org/abs/2602.02276}
}

@misc{internvl2025internvl35,
  title={InternVL3.5: Advancing Open-Source Multimodal Models in Versatility, Reasoning, and Efficiency},
  author={Wang, Weiyun and others},
  year={2025},
  eprint={2508.18265},
  archivePrefix={arXiv},
  url={https://arxiv.org/abs/2508.18265}
}

@misc{zai2025glm46v,
  title={GLM-4.6V: Open Source Multimodal Models with Native Tool Use},
  author={{Z.ai}},
  year={2025},
  url={https://z.ai/blog/glm-4.6v}
}

@misc{interns1pro2026,
  title={Intern-S1-Pro: Scientific Multimodal Foundation Model at Trillion Scale},
  author={{InternLM Team}},
  year={2026},
  eprint={2603.25040},
  archivePrefix={arXiv},
  url={https://arxiv.org/abs/2603.25040}
}

@misc{qwen2026qwen35,
  title={Qwen3.5: Towards Native Multimodal Agents},
  author={{Qwen Team}},
  year={2026},
  url={https://qwen.ai/blog?id=qwen3.5}
}

@misc{bytedance2025seed16,
  title={Introduction to Techniques Used in Seed1.6},
  author={{ByteDance Seed Team}},
  year={2025},
  url={https://seed.bytedance.com/en/blog/introduction-to-techniques-used-in-seed1-6}
}

@misc{longcat2025omni,
  title={LongCat-Flash-Omni Technical Report},
  author={{Meituan LongCat Team}},
  year={2025},
  eprint={2511.00279},
  archivePrefix={arXiv},
  url={https://arxiv.org/abs/2511.00279}
}

@misc{ernie2026technical,
  title={ERNIE 5.0 Technical Report},
  author={Wang, Haifeng and others},
  year={2026},
  eprint={2602.04705},
  archivePrefix={arXiv},
  url={https://arxiv.org/abs/2602.04705}
}

@misc{mistral2025large3,
  title={Mistral Large 3},
  author={{Mistral AI}},
  year={2025},
  url={https://docs.mistral.ai/models/mistral-large-3-25-12}
}

@misc{mimo2026v26pro,
  title={MiMo-V2.6-Pro-RL},
  author={{Xiaomi MiMo Team}},
  year={2026},
  url={https://huggingface.co/XiaomiMiMo/MiMo-V2.6-Pro-RL}
}

@misc{minimax2026m3,
  title={MiniMax-M3},
  author={{MiniMax AI}},
  year={2026},
  url={https://huggingface.co/MiniMaxAI/MiniMax-M3}
}

@misc{stepfun2025step3,
  title={Step-3 is Large yet Affordable: Model-System Co-Design for Cost-Effective Decoding},
  author={{StepFun Team}},
  year={2025},
  eprint={2507.19427},
  archivePrefix={arXiv},
  url={https://arxiv.org/abs/2507.19427}
}

@misc{nvidia2026nemotron3omni,
  title={Nemotron 3 Nano Omni: Efficient and Open Multimodal Intelligence},
  author={{NVIDIA}},
  year={2026},
  eprint={2604.24954},
  archivePrefix={arXiv},
  url={https://arxiv.org/abs/2604.24954}
}

@misc{rynnbrain2026,
  title={RynnBrain 1.1: Towards More Capable and Generalizable Embodied Foundation Model},
  author={{Alibaba DAMO Academy}},
  year={2026},
  eprint={2607.17977},
  archivePrefix={arXiv},
  url={https://arxiv.org/abs/2607.17977}
}

@misc{hyembodied2026,
  title={Hy-Embodied-VLM-1.0: Efficient Physical-World Agents},
  author={Wang, Ziyi and others},
  year={2026},
  eprint={2607.12894},
  archivePrefix={arXiv},
  url={https://arxiv.org/abs/2607.12894}
}

@misc{uivenus2026,
  title={UI-Venus-1.5 Technical Report},
  author={{Venus Team}},
  year={2026},
  eprint={2602.09082},
  archivePrefix={arXiv},
  url={https://arxiv.org/abs/2602.09082}
}

@misc{holo2026,
  title={Holo3.1: Fast and Local Computer Use Agents},
  author={{H Company}},
  year={2026},
  url={https://huggingface.co/Hcompany/Holo-3.1-35B-A3B}
}

@article{li2024cumo,
  title={Cumo: Scaling multimodal llm with co-upcycled mixture-of-experts},
  author={Li, Jiachen and Wang, Xinyao and Zhu, Sijie and Kuo, Chia-Wen and Xu, Lu and Chen, Fan and Jain, Jitesh and Shi, Humphrey and Wen, Longyin},
  journal={Advances in Neural Information Processing Systems},
  volume={37},
  pages={131224--131246},
  year={2024}
}

@inproceedings{nips/MustafaRPJH22,
  author       = {Basil Mustafa and
                  Carlos Riquelme and
                  Joan Puigcerver and
                  Rodolphe Jenatton and
                  Neil Houlsby},
  editor       = {Sanmi Koyejo and
                  S. Mohamed and
                  A. Agarwal and
                  Danielle Belgrave and
                  K. Cho and
                  A. Oh},
  title        = {Multimodal Contrastive Learning with LIMoE: the Language-Image Mixture
                  of Experts},
  booktitle    = {Advances in Neural Information Processing Systems 35: Annual Conference
                  on Neural Information Processing Systems 2022, NeurIPS 2022, New Orleans,
                  LA, USA, November 28 - December 9, 2022},
  year         = {2022}
}

@article{liang2024mixture,
  title={Mixture-of-transformers: A sparse and scalable architecture for multi-modal foundation models},
  author={Liang, Weixin and Yu, Lili and Luo, Liang and Iyer, Srinivasan and Dong, Ning and Zhou, Chunting and Ghosh, Gargi and Lewis, Mike and Yih, Wen-tau and Zettlemoyer, Luke and others},
  journal={arXiv preprint arXiv:2411.04996},
  year={2024}
}

@inproceedings{iclr/ShiLWLRZHYSYSCT25,
  author       = {Min Shi and
                  Fuxiao Liu and
                  Shihao Wang and
                  Shijia Liao and
                  Subhashree Radhakrishnan and
                  Yilin Zhao and
                  De{-}An Huang and
                  Hongxu Yin and
                  Karan Sapra and
                  Yaser Yacoob and
                  Humphrey Shi and
                  Bryan Catanzaro and
                  Andrew Tao and
                  Jan Kautz and
                  Zhiding Yu and
                  Guilin Liu},
  title        = {Eagle: Exploring The Design Space for Multimodal LLMs with Mixture
                  of Encoders},
  booktitle    = {The Thirteenth International Conference on Learning Representations,
                2025, Singapore},
  year         = {2025},

}

@article{komatsuzaki2022sparse,
  title={Sparse upcycling: Training mixture-of-experts from dense checkpoints},
  author={Komatsuzaki, Aran and Puigcerver, Joan and Lee-Thorp, James and Ruiz, Carlos Riquelme and Mustafa, Basil and Ainslie, Joshua and Tay, Yi and Dehghani, Mostafa and Houlsby, Neil},
  journal={arXiv preprint arXiv:2212.05055},
  year={2022}
}

@article{han2025massively,
  title={Massively multimodal foundation models: A framework for capturing interactions with specialized mixture-of-experts},
  author={Han, Xing and Chung, Hsing-Huan and Ghosh, Joydeep and Liang, Paul Pu and Saria, Suchi},
  journal={arXiv preprint arXiv:2509.25678},
  year={2025}
}

@inproceedings{jing2026modeling,
  title={Modeling Multimodal Information Cascade on Social Media with Interpretable Mixture of Experts},
  author={Jing, Xin and Shi, Zeyu and Cheng, Zhangtao and Jing, Yichen and Lu, Yuhuan and Deng, Bangchao and Yang, Dingqi},
  booktitle={Proceedings of the ACM Web Conference 2026},
  pages={9757--9767},
  year={2026}
}

@inproceedings{emnlp/YuQJSML24,
  author       = {Haofei Yu and
                  Zhengyang Qi and
                  Lawrence Jang and
                  Russ Salakhutdinov and
                  Louis{-}Philippe Morency and
                  Paul Pu Liang},
  title        = {MMoE: Enhancing Multimodal Models with Mixtures of Multimodal Interaction
                  Experts},
  booktitle    = {Proceedings of the 2024 Conference on Empirical Methods in Natural
                  Language Processing, Miami, FL, USA, 
                  2024},
  publisher    = {Association for Computational Linguistics},
  year         = {2024},
}

@article{xin2025i2moe,
  title={I2MoE: Interpretable Multimodal Interaction-aware Mixture-of-Experts},
  author={Xin, Jiayi and Yun, Sukwon and Peng, Jie and Choi, Inyoung and Ballard, Jenna L and Chen, Tianlong and Long, Qi},
  journal={arXiv preprint arXiv:2505.19190},
  year={2025}
}

@inproceedings{jiang2024m4oe,
  title={M4oe: A foundation model for medical multimodal image segmentation with mixture of experts},
  author={Jiang, Yufeng and Shen, Yiqing},
  booktitle={international conference on medical image computing and computer-assisted intervention},
  pages={621--631},
  year={2024},
  organization={Springer}
}

@article{lin2024moe,
  title={Moe-llava: Mixture of experts for large vision-language models},
  author={Lin, Bin and Tang, Zhenyu and Ye, Yang and Cui, Jiaxi and Zhu, Bin and Jin, Peng and Huang, Jinfa and Zhang, Junwu and Pang, Yatian and Ning, Munan and others},
  journal={arXiv preprint arXiv:2401.15947},
  year={2024}
}

@article{akbari2023alternating,
  title={Alternating gradient descent and mixture-of-experts for integrated multimodal perception},
  author={Akbari, Hassan and Kondratyuk, Dan and Cui, Yin and Hornung, Rachel and Wang, Huisheng and Adam, Hartwig},
  journal={Advances in Neural Information Processing Systems},
  volume={36},
  pages={79142--79154},
  year={2023}
}

@article{he2024upcycling,
  title={Upcycling large language models into mixture of experts},
  author={He, Ethan and Khattar, Abhinav and Prenger, Ryan and Korthikanti, Vijay and Yan, Zijie and Liu, Tong and Fan, Shiqing and Aithal, Ashwath and Shoeybi, Mohammad and Catanzaro, Bryan},
  journal={arXiv preprint arXiv:2410.07524},
  year={2024}
}

@article{chopra2025medmoe,
  title={MedMoE: Modality-Specialized Mixture of Experts for Medical Vision-Language Understanding},
  author={Chopra, Shivang and Sanchez-Rodriguez, Gabriela and Mao, Lingchao and Feola, Andrew J and Li, Jing and Kira, Zsolt},
  journal={arXiv preprint arXiv:2506.08356},
  year={2025}
}

@article{jiang2024med,
  title={Med-moe: Mixture of domain-specific experts for lightweight medical vision-language models},
  author={Jiang, Songtao and Zheng, Tuo and Zhang, Yan and Jin, Yeying and Yuan, Li and Liu, Zuozhu},
  journal={arXiv preprint arXiv:2404.10237},
  year={2024}
}

@inproceedings{luo2025monointernvl,
  title={{Mono-InternVL}: Pushing the Boundaries of Monolithic Multimodal Large Language Models with Endogenous Visual Pre-training},
  author={Luo, Gen and Yang, Xue and Dou, Wenhan and Wang, Zhaokai and Liu, Jiawen and Dai, Jifeng and Qiao, Yu and Zhu, Xizhou},
  booktitle={Proceedings of the IEEE/CVF Conference on Computer Vision and Pattern Recognition},
  pages={24960--24971},
  year={2025}
}

@inproceedings{tran2026clipfmoe,
  title={{CLIP-FMoE}: Scalable CLIP via Fused Mixture-of-Experts with Enforced Specialization},
  author={Tran, Luong and Nguyen, Lan-Cuong and Nguyen, Huynh Dang and Dat, Nguyen Cong and Le, Dung D. and Nguyen, Van},
  booktitle={International Conference on Learning Representations},
  year={2026}
}

@inproceedings{choi2026structural,
  title={Toward Structural Multimodal Representations: Specialization, Selection, and Sparsification via Mixture-of-Experts},
  author={Choi, Hahyeon and Kwak, Nojun},
  booktitle={Forty-third International Conference on Machine Learning},
  year={2026}
}

@inproceedings{cheng2026ermoe,
  title={{ERMoE}: Eigen-Reparameterized Mixture-of-Experts for Stable Routing and Interpretable Specialization},
  author={Cheng, Anzhe and Duan, Shukai and Li, Shixuan and Yin, Chenzhong and Cheng, Mingxi and Ping, Heng and Chattopadhyay, Tamoghna and Thomopoulos, Sophia I. and Nazarian, Shahin and Thompson, Paul M. and Bogdan, Paul},
  booktitle={Proceedings of the IEEE/CVF Conference on Computer Vision and Pattern Recognition},
  pages={12997--13006},
  year={2026}
}

@inproceedings{nie2026moon,
  title={{MOON2.0}: Dynamic Modality-balanced Multimodal Representation Learning for E-commerce Product Understanding},
  author={Nie, Zhanheng and Fu, Chenghan and Zhang, Daoze and Wu, Junxian and Guan, Wanxian and Wang, Pengjie and Xu, Jian and Zheng, Bo},
  booktitle={Proceedings of the IEEE/CVF Conference on Computer Vision and Pattern Recognition},
  pages={22975--22985},
  year={2026}
}

@article{zong2024mova,
  title={Mova: Adapting mixture of vision experts to multimodal context},
  author={Zong, Zhuofan and Ma, Bingqi and Shen, Dazhong and Song, Guanglu and Shao, Hao and Jiang, Dongzhi and Li, Hongsheng and Liu, Yu},
  journal={Advances in Neural Information Processing Systems},
  volume={37},
  pages={103305--103333},
  year={2024}
}

@article{lin2024moma,
  title={Moma: Efficient early-fusion pre-training with mixture of modality-aware experts},
  author={Lin, Xi Victoria and Shrivastava, Akshat and Luo, Liang and Iyer, Srinivasan and Lewis, Mike and Ghosh, Gargi and Zettlemoyer, Luke and Aghajanyan, Armen},
  journal={arXiv preprint arXiv:2407.21770},
  year={2024}
}

@article{zhang2024multiple,
  title={Multiple heads are better than one: Mixture of modality knowledge experts for entity representation learning},
  author={Zhang, Yichi and Chen, Zhuo and Guo, Lingbing and Xu, Yajing and Hu, Binbin and Liu, Ziqi and Zhang, Wen and Chen, Huajun},
  journal={arXiv preprint arXiv:2405.16869},
  year={2024}
}

@article{puigcerver2023sparse,
  title={From sparse to soft mixtures of experts},
  author={Puigcerver, Joan and Riquelme, Carlos and Mustafa, Basil and Houlsby, Neil},
  journal={arXiv preprint arXiv:2308.00951},
  year={2023}
}

@article{liang2023quantifying,
  title={Quantifying \& modeling multimodal interactions: An information decomposition framework},
  author={Liang, Paul Pu and Cheng, Yun and Fan, Xiang and Ling, Chun Kai and Nie, Suzanne and Chen, Richard and Deng, Zihao and Allen, Nicholas and Auerbach, Randy and Mahmood, Faisal and others},
  journal={Advances in Neural Information Processing Systems},
  volume={36},
  pages={27351--27393},
  year={2023}
}

@inproceedings{liang2023multimodal,
  title={Multimodal fusion interactions: A study of human and automatic quantification},
  author={Liang, Paul Pu and Cheng, Yun and Salakhutdinov, Ruslan and Morency, Louis-Philippe},
  booktitle={Proceedings of the 25th International Conference on Multimodal Interaction},
  pages={425--435},
  year={2023}
}

@article{williams2010nonnegative,
  title={Nonnegative decomposition of multivariate information},
  author={Williams, Paul L and Beer, Randall D},
  journal={arXiv preprint arXiv:1004.2515},
  year={2010}
}

@article{jacobs1991adaptive,
  title={Adaptive mixtures of local experts},
  author={Jacobs, Robert A and Jordan, Michael I and Nowlan, Steven J and Hinton, Geoffrey E},
  journal={Neural computation},
  volume={3},
  number={1},
  pages={79--87},
  year={1991},
  publisher={MIT Press}
}

@inproceedings{wu2024omni,
  title={Omni-smola: Boosting generalist multimodal models with soft mixture of low-rank experts},
  author={Wu, Jialin and Hu, Xia and Wang, Yaqing and Pang, Bo and Soricut, Radu},
  booktitle={Proceedings of the IEEE/CVF Conference on Computer Vision and Pattern Recognition},
  pages={14205--14215},
  year={2024}
}

@inproceedings{lee2024moai,
  title={Moai: Mixture of all intelligence for large language and vision models},
  author={Lee, Byung-Kwan and Park, Beomchan and Won Kim, Chae and Man Ro, Yong},
  booktitle={European Conference on Computer Vision},
  pages={273--302},
  year={2024},
  organization={Springer}
}

@article{xue2023raphael,
  title={Raphael: Text-to-image generation via large mixture of diffusion paths},
  author={Xue, Zeyue and Song, Guanglu and Guo, Qiushan and Liu, Boxiao and Zong, Zhuofan and Liu, Yu and Luo, Ping},
  journal={Advances in Neural Information Processing Systems},
  volume={36},
  pages={41693--41706},
  year={2023}
}

@article{mao2025unimod,
  title={Unimod: Efficient unified multimodal transformers with mixture-of-depths},
  author={Mao, Weijia and Yang, Zhenheng and Shou, Mike Zheng},
  journal={arXiv preprint arXiv:2502.06474},
  year={2025}
}

@article{li2025uni,
  title={Uni-moe: Scaling unified multimodal llms with mixture of experts},
  author={Li, Yunxin and Jiang, Shenyuan and Hu, Baotian and Wang, Longyue and Zhong, Wanqi and Luo, Wenhan and Ma, Lin and Zhang, Min},
  journal={IEEE Transactions on Pattern Analysis and Machine Intelligence},
  year={2025},
  publisher={IEEE}
}

@article{raposo2024mixture,
  title={Mixture-of-depths: Dynamically allocating compute in transformer-based language models},
  author={Raposo, David and Ritter, Sam and Richards, Blake and Lillicrap, Timothy and Humphreys, Peter Conway and Santoro, Adam},
  journal={arXiv preprint arXiv:2404.02258},
  year={2024}
}

@article{shen2024mome,
  title={Mome: Mixture of multimodal experts for generalist multimodal large language models},
  author={Shen, Leyang and Chen, Gongwei and Shao, Rui and Guan, Weili and Nie, Liqiang},
  journal={Advances in neural information processing systems},
  volume={37},
  pages={42048--42070},
  year={2024}
}

@article{lei2024m3,
  title={M3-Jepa: Multimodal Alignment via Multi-directional MoE based on the JEPA framework},
  author={Lei, Hongyang and Cheng, Xiaolong and Wang, Dan and Fan, Kun and Qin, Qi and Huang, Huazhen and Wu, Yetao and Gu, Qingqing and Jiang, Zhonglin and Chen, Yong and others},
  journal={arXiv preprint arXiv:2409.05929},
  year={2024}
}

@article{han2024fusemoe,
  title={Fusemoe: Mixture-of-experts transformers for fleximodal fusion},
  author={Han, Xing and Nguyen, Huy and Harris, Carl and Ho, Nhat and Saria, Suchi},
  journal={Advances in Neural Information Processing Systems},
  volume={37},
  pages={67850--67900},
  year={2024}
}

@article{yun2024flex,
  title={Flex-moe: Modeling arbitrary modality combination via the flexible mixture-of-experts},
  author={Yun, Sukwon and Choi, Inyoung and Peng, Jie and Wu, Yangfan and Bao, Jingxuan and Zhang, Qiyiwen and Xin, Jiayi and Long, Qi and Chen, Tianlong},
  journal={Advances in Neural Information Processing Systems},
  volume={37},
  pages={98782--98805},
  year={2024}
}

@inproceedings{zheng2025understanding,
  title={Understanding Why Large Language Models Can Be Ineffective in Time Series Analysis: The Impact of Modality Alignment},
  author={Zheng, Liangwei Nathan and Dong, Chang and Zhang, Wei Emma and Yue, Lin and Xu, Miao and Maennel, Olaf and Chen, Weitong},
  booktitle={Proceedings of the 31st ACM SIGKDD Conference on Knowledge Discovery and Data Mining V. 2},
  pages={4026--4037},
  year={2025}
}

@article{zheng2025rethinking,
  title={Rethinking Gating Mechanism in Sparse MoE: Handling Arbitrary Modality Inputs with Confidence-Guided Gate},
  author={Zheng, Liangwei Nathan and Zhang, Wei Emma and Guo, Mingyu and Xu, Miao and Maennel, Olaf and Chen, Weitong},
  journal={arXiv preprint arXiv:2505.19525},
  year={2025}
}

@article{li2025r2,
  title={R2-T2: Re-Routing in Test-Time for Multimodal Mixture-of-Experts},
  author={Li, Zhongyang and Li, Ziyue and Zhou, Tianyi},
  journal={arXiv preprint arXiv:2502.20395},
  year={2025}
}

@article{luo2024gamma,
  title={$\gamma-$ MoD: Exploring Mixture-of-Depth Adaptation for Multimodal Large Language Models},
  author={Luo, Yaxin and Luo, Gen and Ji, Jiayi and Zhou, Yiyi and Sun, Xiaoshuai and Shen, Zhiqiang and Ji, Rongrong},
  journal={arXiv preprint arXiv:2410.13859},
  year={2024}
}

@inproceedings{wang2020makes,
  title={What makes training multi-modal classification networks hard?},
  author={Wang, Weiyao and Tran, Du and Feiszli, Matt},
  booktitle={Proceedings of the IEEE/CVF conference on computer vision and pattern recognition},
  pages={12695--12705},
  year={2020}
}

@article{zhang2024clipmoe,
  title={{CLIP-MoE}: Towards Building Mixture of Experts for {CLIP} with Diversified Multiplet Upcycling},
  author={Zhang, Jihai and Qu, Xiaoye and Zhu, Tong and Cheng, Yu},
  journal={arXiv preprint arXiv:2409.19291},
  year={2024},
  url={https://arxiv.org/abs/2409.19291}
}

@article{shu2024llavamod,
  title={{LLaVA-MoD}: Making {LLaVA} Tiny via {MoE} Knowledge Distillation},
  author={Shu, Fangxun and Liao, Yue and Zhuo, Le and Xu, Chenning and Zhang, Lei and Zhang, Guanghao and Shi, Haonan and Chen, Long and Zhong, Tao and He, Wanggui and Li, Siming and Li, Haoyuan and Li, Bolin and Yu, Zhelun and Liu, Si and Li, Hongsheng and Jiang, Hao},
  journal={arXiv preprint arXiv:2408.15881},
  year={2024},
  url={https://arxiv.org/abs/2408.15881}
}

@article{rang2025eve,
  title={{Eve}: Efficient Multimodal Vision Language Models with Elastic Visual Experts},
  author={Rang, Miao and Bi, Zhenni and Liu, Chuanjian and Tang, Yehui and Han, Kai and Wang, Yunhe},
  journal={arXiv preprint arXiv:2501.04322},
  year={2025},
  url={https://arxiv.org/abs/2501.04322}
}

@inproceedings{chen2024eve,
  title={Eve: Efficient vision-language pre-training with masked prediction and modality-aware moe},
  author={Chen, Junyi and Guo, Longteng and Sun, Jia and Shao, Shuai and Yuan, Zehuan and Lin, Liang and Zhang, Dongyu},
  booktitle={Proceedings of the AAAI Conference on Artificial Intelligence},
  volume={38},
  number={2},
  pages={1110--1119},
  year={2024}
}

@inproceedings{wu2024videollmmod,
  title={{VideoLLM-MoD}: Efficient Video-Language Streaming with Mixture-of-Depths Vision Computation},
  author={Wu, Shiwei and Chen, Joya and Lin, Kevin Qinghong and Wang, Qimeng and Gao, Yan and Xu, Qianli and Xu, Tong and Hu, Yao and Chen, Enhong and Shou, Mike Zheng},
  booktitle={Advances in Neural Information Processing Systems},
  volume={37},
  year={2024},
  url={https://proceedings.neurips.cc/paper_files/paper/2024/hash/c6a79e139ec4f371701ea8cc9e06018e-Abstract-Conference.html}
}

@inproceedings{zhang2025pmod,
  title={{p-MoD}: Building Mixture-of-Depths {MLLMs} via Progressive Ratio Decay},
  author={Zhang, Jun and Meng, Desen and Zhang, Zhengming and Huang, Zhenpeng and Wu, Tao and Wang, Limin},
  booktitle={Proceedings of the IEEE/CVF International Conference on Computer Vision},
  year={2025},
  pages={3705--3715},
  url={https://openaccess.thecvf.com/content/ICCV2025/html/Zhang_p-MoD_Building_Mixture-of-Depths_MLLMs_via_Progressive_Ratio_Decay_ICCV_2025_paper.html}
}

@inproceedings{huang2026cat,
  title={Cat-{M}o{D}: Accelerating Multimodal Alignment via Caption Token Guided Asymmetric Mixture-of-Depths},
  author={Huang, YiJie and Yang, Xiaocui and Feng, Shi and Zhang, Wen and Song, Kaisong and Zhang, Yifei and Wang, Daling},
  booktitle={Proceedings of the 64th Annual Meeting of the Association for Computational Linguistics (Volume 1: Long Papers)},
  year={2026},
  month=jul,
  pages={26354--26376},
  publisher={Association for Computational Linguistics},
  doi={10.18653/v1/2026.acl-long.1213},
  url={https://aclanthology.org/2026.acl-long.1213/}
}

@inproceedings{luo2023cheap,
  title     = {Cheap and Quick: Efficient Vision-Language Instruction Tuning for Large Language Models},
  author    = {Luo, Gen and Zhou, Yiyi and Ren, Tianhe and Chen, Shengxin and Sun, Xiaoshuai and Ji, Rongrong},
  booktitle = {Advances in Neural Information Processing Systems},
  volume    = {36},
  pages     = {29615--29627},
  year      = {2023}
}

@article{kowsher2026lime,
  title   = {LiME: Lightweight Mixture of Experts for Efficient Multimodal Multi-task Learning},
  author  = {Kowsher, Md and Mansoor, Haris and Prottasha, Nusrat Jahan and Garibay, Ozlem and Zhu, Victor and Ji, Zhengping and Chen, Chen},
  journal = {arXiv preprint arXiv:2604.02338},
  year    = {2026},
  url     = {https://arxiv.org/abs/2604.02338}
}

@article{yu2024moeadapters,
  title   = {Boosting Continual Learning of Vision-Language Models via Mixture-of-Experts Adapters},
  author  = {Yu, Jiazuo and Zhuge, Yunzhi and Zhang, Lu and Hu, Ping and Wang, Dong and Lu, Huchuan and He, You},
  journal = {arXiv preprint arXiv:2403.11549},
  year    = {2024},
  url     = {https://arxiv.org/abs/2403.11549}
}

@inproceedings{huai2025clmoe,
  title     = {CL-MoE: Enhancing Multimodal Large Language Model with Dual Momentum Mixture-of-Experts for Continual Visual Question Answering},
  author    = {Huai, Tianyu and Zhou, Jie and Wu, Xingjiao and Chen, Qin and Bai, Qingchun and Zhou, Ze and He, Liang},
  booktitle = {Proceedings of the IEEE/CVF Conference on Computer Vision and Pattern Recognition},
  pages     = {19608--19617},
  year      = {2025},
  doi       = {10.1109/CVPR52734.2025.01826}
}

@article{zhao2025llavacmoe,
  title   = {LLaVA-CMoE: Towards Continual Mixture of Experts for Large Vision-Language Models},
  author  = {Zhao, Hengyuan and Wang, Ziqin and Sun, Qixin and Song, Kaiyou and Li, Yilin and Hu, Xiaolin and Guo, Qingpei and Liu, Si},
  journal = {arXiv preprint arXiv:2503.21227},
  year    = {2025},
  url     = {https://arxiv.org/abs/2503.21227}
}

@misc{meta2025llama4,
  author       = {{Meta AI}},
  title        = {The Llama 4 Herd: The Beginning of a New Era of Natively Multimodal AI Innovation},
  year         = {2025},
  month        = apr,
  howpublished = {\url{https://ai.meta.com/blog/llama-4-multimodal-intelligence/}},
  note         = {Accessed: 2026-09-07}
}

@article{bao2022vlmo,
  title={Vlmo: Unified vision-language pre-training with mixture-of-modality-experts},
  author={Bao, Hangbo and Wang, Wenhui and Dong, Li and Liu, Qiang and Mohammed, Owais Khan and Aggarwal, Kriti and Som, Subhojit and Piao, Songhao and Wei, Furu},
  journal={Advances in neural information processing systems},
  volume={35},
  pages={32897--32912},
  year={2022}
}

@inproceedings{shen2023scaling,
  title={Scaling vision-language models with sparse mixture of experts},
  author={Shen, Sheng and Yao, Zhewei and Li, Chunyuan and Darrell, Trevor and Keutzer, Kurt and He, Yuxiong},
  booktitle={Findings of the Association for Computational Linguistics: EMNLP 2023},
  pages={11329--11344},
  year={2023}
}

@inproceedings{wang2023image,
  title={Image as a foreign language: Beit pretraining for vision and vision-language tasks},
  author={Wang, Wenhui and Bao, Hangbo and Dong, Li and Bjorck, Johan and Peng, Zhiliang and Liu, Qiang and Aggarwal, Kriti and Mohammed, Owais Khan and Singhal, Saksham and Som, Subhojit and others},
  booktitle={2023 IEEE/CVF Conference on Computer Vision and Pattern Recognition (CVPR)},
  pages={19175--19186},
  year={2023},
  organization={IEEE}
}

@article{yu2026pathmoe,
  title={PathMoE: Interpretable Multimodal Interaction Experts for Pediatric Brain Tumor Classification},
  author={Yu, Jian and Nguyen, Joakim and Fang, Jinrui and Naeem, Awais and Cao, Zeyuan and Krishnan, Sanjay and Konz, Nicholas and Chen, Tianlong and Krishnan, Chandra and Wang, Hairong and others},
  journal={arXiv preprint arXiv:2603.01547},
  year={2026}
}

@inproceedings{xu2025chartmoe,
  title     = {{ChartMoE}: Mixture of Diversely Aligned Expert Connector for Chart Understanding},
  author    = {Xu, Zhengzhuo and Qu, Bowen and Qi, Yiyan and Du, Sinan and Xu, Chengjin and Yuan, Chun and Guo, Jian},
  booktitle = {International Conference on Learning Representations},
  year      = {2025}
}

@inproceedings{zhou2025unialign,
  title     = {{UNIALIGN}: Scaling Multimodal Alignment within One Unified Model},
  author    = {Zhou, Bo and Li, Liulei and Wang, Yujia and Liu, Huafeng and Yao, Yazhou and Wang, Wenguan},
  booktitle = {Proceedings of the IEEE/CVF Conference on Computer Vision and Pattern Recognition},
  pages     = {29644--29655},
  year      = {2025},
  doi       = {10.1109/CVPR52734.2025.02760}
}

@inproceedings{sun2025parrot,
  title     = {Parrot: Multilingual Visual Instruction Tuning},
  author    = {Sun, Hai-Long and Zhou, Da-Wei and Li, Yang and Lu, Shiyin and Yi, Chao and Chen, Qing-Guo and Xu, Zhao and Luo, Weihua and Zhang, Kaifu and Zhan, De-Chuan and Ye, Han-Jia},
  booktitle = {Proceedings of the 42nd International Conference on Machine Learning},
  pages     = {57984--58007},
  year      = {2025},
  volume    = {267},
  series    = {Proceedings of Machine Learning Research},
  publisher = {PMLR}
}

@inproceedings{kim2025mohave,
  author    = {Sungnyun Kim and
               Kangwook Jang and
               Sangmin Bae and
               Sungwoo Cho and
               Se-Young Yun},
  title     = {MoHAVE: Mixture of Hierarchical Audio-Visual Experts for Robust Speech Recognition},
  booktitle = {International Conference on Machine Learning},
  year      = {2025}
}

@inproceedings{jing2026evomoe,
  author    = {Linglin Jing and
               Yuting Gao and
               Zhigang Wang and
               Wang Lan and
               Yiwen Tang and
               Weiyun Wang and
               Wenhai Wang and
               Qingpei Guo},
  title     = {EvoMoE: Expert Evolution in Mixture of Experts for Multimodal Large Language Models},
  booktitle = {Proceedings of the AAAI Conference on Artificial Intelligence},
  pages     = {22463--22471},
  year      = {2026}
}

@article{wang2025moiie,
  author  = {Dianyi Wang and
             Siyuan Wang and
             Zejun Li and
             Yikun Wang and
             Yitong Li and
             Duyu Tang and
             Xiaoyu Shen and
             Xuanjing Huang and
             Zhongyu Wei},
  title   = {MoIIE: Mixture of Intra- and Inter-Modality Experts for Large Vision Language Models},
  journal = {CoRR},
  volume  = {abs/2508.09779},
  year    = {2025},
  doi     = {10.48550/ARXIV.2508.09779}
}

@article{lou2026most,
  author  = {Yuxuan Lou and
             Kai Yang and
             Yang You},
  title   = {MoST: Mixing Speech and Text with Modality-Aware Mixture of Experts},
  journal = {CoRR},
  volume  = {abs/2601.10272},
  year    = {2026},
  doi     = {10.48550/ARXIV.2601.10272}
}

@article{xia2025smar,
  author  = {Guoyang Xia and
             Yifeng Ding and
             Fengfa Li and
             Lei Ren and
             Wei Chen and
             Fangxiang Feng and
             Xiaojie Wang},
  title   = {SMAR: Soft Modality-Aware Routing Strategy for MoE-based Multimodal Large Language Models Preserving Language Capabilities},
  journal = {CoRR},
  volume  = {abs/2506.06406},
  year    = {2025},
  doi     = {10.48550/ARXIV.2506.06406}
}

@article{bo2026smoes,
  author  = {Zi-Hao Bo and
             Yaqian Li and
             Anzhou Hou and
             Rinyoichi Takezoe and
             Ertao Zhao and
             Tianxiang Pan and
             Jiale Yan and
             Mo Guang and
             Kaiwen Long},
  title   = {SMoES: Soft Modality-Guided Expert Specialization in MoE-VLMs},
  journal = {CoRR},
  volume  = {abs/2604.23996},
  year    = {2026},
  doi     = {10.48550/ARXIV.2604.23996}
}

@article{li2024aria,
  author  = {Dongxu Li and
             Yudong Liu and
             Haoning Wu and
             Yue Wang and
             Zhiqi Shen and
             Bowen Qu and
             Xinyao Niu and
             Guoyin Wang and
             Bei Chen and
             Junnan Li},
  title   = {Aria: An Open Multimodal Native Mixture-of-Experts Model},
  journal = {CoRR},
  volume  = {abs/2410.05993},
  year    = {2024},
  doi     = {10.48550/ARXIV.2410.05993}
}

@inproceedings{shukor2025scaling,
  author    = {Mustafa Shukor and
               Enrico Fini and
               Victor Guilherme Turrisi da Costa and
               Matthieu Cord and
               Joshua M. Susskind and
               Alaaeldin El-Nouby},
  title     = {Scaling Laws for Native Multimodal Models},
  booktitle = {IEEE/CVF International Conference on Computer Vision},
  pages     = {12--23},
  year      = {2025},
  doi       = {10.1109/ICCV51701.2025.00009}
}

@inproceedings{cheng2024mixtures,
  author    = {Ying Cheng and
               Yang Li and
               Junjie He and
               Rui Feng},
  title     = {Mixtures of Experts for Audio-Visual Learning},
  booktitle = {Advances in Neural Information Processing Systems},
  year      = {2024}
}

@inproceedings{lin2025remote,
  author    = {Xinkui Lin and
               Yongxiu Xu and
               Minghao Tang and
               Shilong Zhang and
               Hongbo Xu and
               Hao Xu and
               Yubin Wang},
  title     = {REMOTE: A Unified Multimodal Relation Extraction Framework with Multilevel Optimal Transport and Mixture-of-Experts},
  booktitle = {ACM International Conference on Multimedia},
  pages     = {121--130},
  year      = {2025},
  doi       = {10.1145/3746027.3754868}
}

@inproceedings{wan2025ugg,
  author    = {Xixi Wan and
               Aihua Zheng and
               Bo Jiang and
               Beibei Wang and
               Chenglong Li and
               Jin Tang},
  title     = {UGG-ReID: Uncertainty-Guided Graph Model for Multi-Modal Object Re-Identification},
  booktitle = {Advances in Neural Information Processing Systems},
  year      = {2025}
}

@article{su2026seatrack,
  author  = {Junbin Su and
             Ziteng Xue and
             Shihui Zhang and
             Kun Chen and
             Weiming Hu and
             Zhipeng Zhang},
  title   = {SEATrack: Simple, Efficient, and Adaptive Multimodal Tracker},
  journal = {CoRR},
  volume  = {abs/2604.12502},
  year    = {2026}
}

@inproceedings{yu2025forcevla,
  author    = {Jiawen Yu and
               Hairuo Liu and
               Qiaojun Yu and
               Jieji Ren and
               Ce Hao and
               Haitong Ding and
               Guangyu Huang and
               Guofan Huang and
               Yan Song and
               Panpan Cai and
               Wenqiang Zhang and
               Cewu Lu},
  title     = {ForceVLA: Enhancing VLA Models with a Force-aware MoE for Contact-rich Manipulation},
  booktitle = {Advances in Neural Information Processing Systems},
  year      = {2025}
}

@inproceedings{xu2026mcmoe,
  title={MCMoE: Completing Missing Modalities with Mixture of Experts for Incomplete Multimodal Action Quality Assessment},
  author={Xu, Huangbiao and Wu, Huanqi and Ke, Xiao and Wu, Junyi and Xu, Rui and Xu, Jinglin},
  booktitle={Proceedings of the AAAI Conference on Artificial Intelligence},
  volume={40},
  number={13},
  pages={11241--11249},
  year={2026},
  doi={10.1609/aaai.v40i13.38104}
}

@inproceedings{novosad2024task,
  title={A Task-Conditional Mixture-of-Experts Model for Missing Modality Segmentation},
  author={Novosad, Philip and Carano, Richard A. D. and Krishnan, Anitha Priya},
  booktitle={Medical Image Computing and Computer Assisted Intervention -- MICCAI 2024},
  pages={34--43},
  year={2024},
  publisher={Springer Nature Switzerland},
  doi={10.1007/978-3-031-72114-4_4}
}

@article{baek2026mars,
  title={Residual-Guided Expert Specialization for Incomplete Multimodal Learning},
  author={Baek, Seunghun and Park, Jihwan and Sim, Jaeyoon and Jeong, Minjae and Lee, Hoseok and Kim, Won Hwa},
  journal={arXiv preprint arXiv:2606.30355},
  year={2026}
}

@article{yin2026imbalanced,
  title={Exploring imbalanced multimodal learning from intra-modality information learning perspective},
  author={Yin, Hang and Xu, Yue and Zhao, Ji and Zhang, Xuehe},
  journal={Knowledge-Based Systems},
  volume={351},
  pages={116881},
  year={2026},
  publisher={Elsevier},
  doi={10.1016/j.knosys.2026.116881}
}

@inproceedings{naderi2026coprime,
  title={From Token Imbalance to Balanced Routing: An ELBO-Regularized Probabilistic Framework for Contrastive Multimodal Learning},
  author={Naderi, Habibeh and Soleimani, Behrouz Haji and Matwin, Stan},
  booktitle={Proceedings of the 29th International Conference on Artificial Intelligence and Statistics},
  pages={2413--2421},
  year={2026},
  volume={300},
  series={Proceedings of Machine Learning Research},
  publisher={PMLR}
}

@article{zhang2026mime,
  title={MIME: Mutual information-guided mixture of experts for robust multimodal sentiment analysis},
  author={Zhang, Dehai and Xiao, Qiang and He, Xu and Zhang, Jiahao},
  journal={Journal of King Saud University Computer and Information Sciences},
  volume={38},
  pages={286},
  year={2026},
  publisher={Springer},
  doi={10.1007/s44443-026-00587-9}
}

@inproceedings{zhu2026qamoe,
  title={QA-MoE: Towards a Continuous Reliability Spectrum with Quality-Aware Mixture of Experts for Robust Multimodal Sentiment Analysis},
  author={Zhu, Yitong and Jiang, Yuxuan and Jiang, Guanxuan and Hou, Bojing and Zhou, Peng Yuan and Lin, Ge and Wang, Yuyang},
  booktitle={Proceedings of the 64th Annual Meeting of the Association for Computational Linguistics (Volume 1: Long Papers)},
  pages={31689--31703},
  year={2026},
  publisher={Association for Computational Linguistics},
  doi={10.18653/v1/2026.acl-long.1461}
}

@article{lan2026causalmoe,
  title={Causal mixture-of-experts for robust multimodal fusion},
  author={Lan, Yang and Long, Xin and Zeng, Yaoyuan and Chen, Huiling and Xiao, Yu and Yang, Jungang},
  journal={Pattern Recognition},
  volume={180},
  pages={114306},
  year={2026},
  publisher={Elsevier},
  doi={10.1016/j.patcog.2026.114306}
}

@article{yu2025moeadapters,
  title={MoE-Adapters++: Toward More Efficient Continual Learning of Vision-Language Models Via Dynamic Mixture-of-Experts Adapters},
  author={Yu, Jiazuo and Huang, Zichen and Zhuge, Yunzhi and Zhang, Lu and Hu, Ping and Wang, Dong and Lu, Huchuan and He, You},
  journal={IEEE Transactions on Pattern Analysis and Machine Intelligence},
  volume={47},
  number={12},
  pages={11912--11928},
  year={2025},
  doi={10.1109/TPAMI.2025.3597942}
}

@inproceedings{chen2024coin,
  title={CoIN: A Benchmark of Continual Instruction Tuning for Multimodel Large Language Models},
  author={Chen, Cheng and Zhu, Junchen and Luo, Xu and Shen, Hengtao and Song, Jingkuan and Gao, Lianli},
  booktitle={Advances in Neural Information Processing Systems},
  year={2024}
}

@inproceedings{ge2025dmole,
  title={Dynamic Mixture of Curriculum LoRA Experts for Continual Multimodal Instruction Tuning},
  author={Ge, Chendi and Wang, Xin and Zhang, Zeyang and Chen, Hong and Fan, Jiapei and Huang, Longtao and Xue, Hui and Zhu, Wenwu},
  booktitle={Proceedings of the 42nd International Conference on Machine Learning},
  pages={19011--19033},
  volume={267},
  series={Proceedings of Machine Learning Research},
  publisher={PMLR},
  year={2025}
}

@inproceedings{zhang2025branchlora,
  title={Enhancing Multimodal Continual Instruction Tuning with BranchLoRA},
  author={Zhang, Duzhen and Ren, Yong and Li, Zhong-Zhi and Yu, Yahan and Dong, Jiahua and Li, Chenxing and Ji, Zhilong and Bai, Jinfeng},
  booktitle={Proceedings of the 63rd Annual Meeting of the Association for Computational Linguistics (Volume 1: Long Papers)},
  pages={5743--5756},
  year={2025},
  publisher={Association for Computational Linguistics},
  doi={10.18653/v1/2025.acl-long.287}
}

@inproceedings{wang2025smolora,
  title={SMoLoRA: Exploring and Defying Dual Catastrophic Forgetting in Continual Visual Instruction Tuning},
  author={Wang, Ziqi and Che, Chang and Wang, Qi and Li, Yangyang and Shi, Zenglin and Wang, Meng},
  booktitle={Proceedings of the IEEE/CVF International Conference on Computer Vision},
  pages={177--186},
  year={2025},
  doi={10.1109/ICCV51701.2025.00024}
}

@inproceedings{chee2025multimodal,
  title={Multi-Modal Continual Learning via Cross-Modality Adapters and Representation Alignment with Knowledge Preservation},
  author={Chee, Evelyn and Hsu, Wynne and Lee, Mong-Li},
  booktitle={European Conference on Artificial Intelligence},
  pages={1083--1090},
  year={2025},
  doi={10.3233/FAIA250918}
}

@article{zhao2026taskshared,
  title={Learning task-shared and specific knowledge via mixture-of-experts in generative model for continual learning},
  author={Zhao, Weinan and Ji, Yanling and Li, Yan and Cheng, De and Han, Junwei and Zhang, Dingwen},
  journal={Pattern Recognition},
  volume={179},
  pages={113504},
  year={2026},
  publisher={Elsevier},
  doi={10.1016/j.patcog.2026.113504}
}

@inproceedings{chen2026moblora,
  title={From Experts to Bases: Orthogonal Subspace Mixture for Continual Multimodal Instruction Tuning},
  author={Chen, Pei and Wang, Xilai and Shi, Qixu and Li, Zejian and Sun, Lingyun},
  booktitle={Proceedings of the 64th Annual Meeting of the Association for Computational Linguistics (Volume 1: Long Papers)},
  pages={10545--10561},
  year={2026},
  publisher={Association for Computational Linguistics},
  doi={10.18653/v1/2026.acl-long.481}
}

@inproceedings{hou2026passmoe,
  title={PASs-MoE: Mitigating Misaligned Co-drift among Router and Experts via Pathway Activation Subspaces for Continual Learning},
  author={Hou, Zhiyan and Guo, Haiyun and Ma, Haokai and Sun, Yandu and Yang, Yonghui and Wang, Jinqiao},
  booktitle={Proceedings of the 64th Annual Meeting of the Association for Computational Linguistics (Volume 1: Long Papers)},
  pages={31959--31972},
  year={2026},
  publisher={Association for Computational Linguistics}
}

@inproceedings{chen2025liveedit,
  title={Lifelong Knowledge Editing for Vision Language Models with Low-Rank Mixture-of-Experts},
  author={Chen, Qizhou and Wang, Chengyu and Wang, Dakan and Zhang, Taolin and Li, Wangyue and He, Xiaofeng},
  booktitle={Proceedings of the IEEE/CVF Conference on Computer Vision and Pattern Recognition},
  pages={9455--9466},
  year={2025},
  doi={10.1109/CVPR52734.2025.00883}
}

@article{chen2025multimelo,
  title={Multi-MELO: Unified multimodal model editing with dynamic LoRA},
  author={Chen, Qin and Yin, Jianghao and Yu, Lang and Zhou, Jie and He, Liang},
  journal={Expert Systems with Applications},
  volume={273},
  pages={126766},
  year={2025},
  publisher={Elsevier},
  doi={10.1016/j.eswa.2025.126766}
}

@inproceedings{zhuang2025seuf,
  title={SEUF: Is Unlearning One Expert Enough for Mixture-of-Experts LLMs?},
  author={Zhuang, Haomin and Zhang, Yihua and Guo, Kehan and Jia, Jinghan and Liu, Gaowen and Liu, Sijia and Zhang, Xiangliang},
  booktitle={Proceedings of the 63rd Annual Meeting of the Association for Computational Linguistics (Volume 1: Long Papers)},
  pages={8664--8678},
  year={2025},
  publisher={Association for Computational Linguistics},
  doi={10.18653/v1/2025.acl-long.424}
}

@inproceedings{zhao2026dymoe,
  title={On Token's Dilemma: Dynamic MoE with Drift-Aware Token Assignment for Continual Learning of Large Vision Language Models},
  author={Zhao, Chongyang and Li, Mingsong and Lu, Haodong and Gong, Dong},
  booktitle={Proceedings of the IEEE/CVF Conference on Computer Vision and Pattern Recognition},
  pages={3941--3952},
  year={2026}
}

@inproceedings{xie2026same,
  title={SAME: Stabilized Mixture-of-Experts for Multimodal Continual Instruction Tuning},
  author={Xie, Zhen-Hao and Tang, Jun-Tao and Shi, Yu-Cheng and Ye, Han-Jia and Zhan, De-Chuan and Zhou, Da-Wei},
  booktitle={Proceedings of the 43rd International Conference on Machine Learning},
  year={2026}
}

@inproceedings{tang2026cram,
  title={CRAM: Centroid-Routing and Adaptive MoE for Multimodal Continual Instruction Tuning},
  author={Tang, Jun-Tao and Xie, Zhen-Hao and Shi, Yu-Cheng and Zhou, Da-Wei},
  booktitle={Proceedings of the 2026 Conference on Empirical Methods in Natural Language Processing},
  year={2026}
}

@inproceedings{xu2026time,
  title={TIME: Tensor-Factorized Mixture-of-Experts with Intrinsic Routing for Lifelong Multimodal Knowledge Editing},
  author={Xu, Dexuan and Wang, Jieyi and Li, Shijie and Wang, Hanpin and Cao, Yongzhi and Huang, Yu},
  booktitle={Proceedings of the 43rd International Conference on Machine Learning},
  year={2026}
}

@inproceedings{wang2027exact,
  title={EXACT: Expert Attribution via Causal Tracing for Targeted Knowledge Unlearning in Mixture-of-Experts Language Models},
  author={Wang, Qicheng and Wang, Quanhe and Wang, Lu},
  booktitle={Advanced Intelligent Computing Technology and Applications},
  series={Lecture Notes in Computer Science},
  volume={16672},
  pages={170--181},
  publisher={Springer},
  year={2027},
  doi={10.1007/978-981-92-3520-9_14}
}

@inproceedings{yu2025proglora,
  title={Progressive LoRA for Multimodal Continual Instruction Tuning},
  author={Yu, Yahan and Zhang, Duzhen and Ren, Yong and Zhao, Xuanle and Chen, Xiuyi and Chu, Chenhui},
  booktitle={Findings of the Association for Computational Linguistics: ACL 2025},
  pages={2779--2796},
  year={2025},
  address={Vienna, Austria},
  publisher={Association for Computational Linguistics},
  doi={10.18653/v1/2025.findings-acl.143}
}

@article{cheng2026incremoe,
  title={Incremental mixture of experts: Continual learning for object detection in forestry scenarios},
  author={Cheng, Ximeng and Zhu, Qiaonan and Jia, Shukun and Cao, Yichao and Lu, Xiaobo},
  journal={Pattern Recognition},
  volume={175},
  pages={113001},
  year={2026},
  publisher={Elsevier},
  doi={10.1016/j.patcog.2025.113001}
}

@article{taskadaptive2026vlm,
  title={Task-adaptive continual learning of vision language models via prototype routing and prompt},
  author={Pan, Lei and Lu, Zhipeng and Zheng, Yuan and Yan, Chongyao and Wen, Haitao and Tao, Linwei and Xu, Chang},
  journal={Neurocomputing},
  volume={672},
  pages={132740},
  year={2026},
  publisher={Elsevier},
  doi={10.1016/j.neucom.2026.132740}
}

@inproceedings{alforhad2026moeconfidence,
  title={Mixing Expertise with Confidence: A Mixture of Experts Framework for Robust Multi-Modal Continual Learning},
  author={Al Forhad, Md Abdullah and Zhu, Yuansheng and Acharya, Abhinab and Liu, Xumin and Yu, Qi and Shi, Weishi},
  booktitle={Proceedings of the 43rd International Conference on Machine Learning},
  year={2026}
}

@inproceedings{mozaffari2026bayesianmoe,
  title={Calibrated Knowledge Aggregation in Bayesian Mixture-of-Experts for Continual VQA},
  author={Mozaffari, Mahsa and Sapkota, Hitesh and Kong, Yu and Liu, Xumin and Yu, Qi},
  booktitle={Proceedings of the 43rd International Conference on Machine Learning},
  volume={306},
  series={Proceedings of Machine Learning Research},
  publisher={PMLR},
  year={2026}
}

@inproceedings{fa2026rank1,
  title={Decomposing and Composing: Towards Efficient Vision-Language Continual Learning via Rank-1 Expert Pool in a Single LoRA},
  author={Fa, Zhan and Duan, Yue and Zhang, Jian and Qi, Lei and Yang, Wanqi and Shi, Yinghuan},
  booktitle={Proceedings of the AAAI Conference on Artificial Intelligence},
  volume={40},
  number={25},
  pages={20985--20993},
  year={2026},
  doi={10.1609/aaai.v40i25.39240}
}

@article{abdullakutty2026breastmoe,
  author  = {Faseela Abdullakutty and Younes Akbari and Sumaya Ali S. A. Al-Maadeed and Rafif Al-Saady and Hadi Mohamad Abu Rasheed},
  title   = {Attention-enhanced Mixture of Experts for multimodal breast cancer detection},
  journal = {Biomedical Signal Processing and Control},
  volume  = {118},
  pages   = {109576},
  year    = {2026},
  url     = {https://dblp.org/rec/journals/bspc/AbdullakuttyAAAR26}
}

@inproceedings{gao2024euar,
  author    = {Zixian Gao and Disen Hu and Xun Jiang and Huimin Lu and Heng Tao Shen and Xing Xu},
  title     = {Enhanced Experts with Uncertainty-Aware Routing for Multimodal Sentiment Analysis},
  booktitle = {Proceedings of the 32nd ACM International Conference on Multimedia},
  pages     = {9650--9659},
  year      = {2024},
  doi       = {10.1145/3664647.3680949},
  url       = {https://dblp.org/rec/conf/mm/GaoH00SX24}
}

@article{li2026pestmoe,
  author  = {Xuechen Li and Lang Qiao and Junxiong Zhou and Vera Krischik and Ce Yang},
  title   = {{Pest-MoE}: A mixture-of-experts framework for stage-aware multimodal few-shot pest recognition},
  journal = {Artificial Intelligence in Agriculture},
  year    = {2026},
  doi     = {10.1016/j.aiia.2026.07.009},
  url     = {https://doi.org/10.1016/j.aiia.2026.07.009}
}

@article{liang2026mamoe,
  author  = {Yongzhe Liang and Ruihong Jiang and Bisheng Wei and Wen Tian and Zhangfeng Ma},
  title   = {{MAMoE}: A Multi-Agent Mixture-of-Experts Framework for {LLM}-Assisted 3D Object Reconstruction and Transmission},
  journal = {IEEE Transactions on Network Science and Engineering},
  volume  = {13},
  pages   = {3862--3878},
  year    = {2026},
  url     = {https://dblp.org/rec/journals/tnse/LiangJWTM26}
}

@article{liao2026mgmtmoe,
  author  = {Yuehui Liao and Yun Zheng and Jingyu Zhu and Yu Chen and Feng Gao and Yaning Feng and Weiji Yang and Guang Yang and Xiaobo Lai and Panfei Li},
  title   = {Self-attention-based mixture-of-experts framework for non-invasive prediction of {MGMT} promoter methylation in glioblastoma using multi-modal {MRI}},
  journal = {Displays},
  volume  = {92},
  pages   = {103358},
  year    = {2026},
  url     = {https://dblp.org/rec/journals/displays/LiaoZZCGFYYLL26}
}

@article{lin2026multisensorfault,
  author  = {Tantao Lin and Zhijun Ren and Kai Huang and Hamid Reza Karimi and Yongsheng Zhu and Kai Feng and Jun Hong},
  title   = {A {MoE-LLM}-based multisensor flexible fusion fault diagnosis method for rotating machinery},
  journal = {Advanced Engineering Informatics},
  volume  = {69},
  pages   = {104009},
  year    = {2026},
  doi     = {10.1016/j.aei.2025.104009},
  url     = {https://doi.org/10.1016/j.aei.2025.104009}
}

@article{liu2025m2cd,
  author  = {Ziyuan Liu and Jiawei Zhang and Wenyu Wang and Yuantao Gu},
  title   = {{$M^2$CD}: A Unified MultiModal Framework for Optical--{SAR} Change Detection With Mixture of Experts and Self-Distillation},
  journal = {IEEE Geoscience and Remote Sensing Letters},
  volume  = {22},
  pages   = {1--5},
  year    = {2025},
  doi     = {10.1109/LGRS.2025.3590959},
  url     = {https://dblp.org/rec/journals/lgrs/LiuZWG25}
}

@inproceedings{park2025mome,
  author    = {Konyul Park and Yecheol Kim and Daehun Kim and Jun Won Choi},
  title     = {Resilient Sensor Fusion Under Adverse Sensor Failures via Multi-Modal Expert Fusion},
  booktitle = {Proceedings of the IEEE/CVF Conference on Computer Vision and Pattern Recognition},
  pages     = {6720--6729},
  year      = {2025},
  doi       = {10.1109/CVPR52734.2025.00630},
  url       = {https://dblp.org/rec/conf/cvpr/ParkKKC25}
}

@article{wang2026buildfunc,
  author  = {Ru Wang and Zhan Zhang and Daoyu Shu and Nan Jia and Fang Wan and Wenkai Hu and Xiaoling Chen and Zhenghong Peng},
  title   = {{BuildFunc-MoE}: An Adaptive Multimodal Mixture-of-Experts Network for Fine-Grained Building Function Identification},
  journal = {Remote Sensing},
  volume  = {18},
  number  = {1},
  pages   = {90},
  year    = {2026},
  doi     = {10.3390/rs18010090},
  url     = {https://doi.org/10.3390/rs18010090}
}

@inproceedings{willibald2025anomaly,
  author    = {Christoph Willibald and Daniel Sliwowski and Dongheui Lee},
  title     = {Multimodal Anomaly Detection with a Mixture-of-Experts},
  booktitle = {2025 IEEE/RSJ International Conference on Intelligent Robots and Systems},
  pages     = {20020--20027},
  year      = {2025},
  url       = {https://dblp.org/rec/conf/iros/WillibaldSL25}
}

@inproceedings{wu2025m4oe,
  author    = {Chenwei Wu and Zitao Shuai and Zhengxu Tang and Luning Wang and Liyue Shen},
  title     = {Dynamic Modeling of Patients, Modalities and Tasks via Multi-modal Multi-task Mixture of Experts},
  booktitle = {The Thirteenth International Conference on Learning Representations},
  year      = {2025},
  url       = {https://dblp.org/rec/conf/iclr/0008STWS25}
}

@article{xi2026semgait,
  author  = {Hao Xi and Peng Lu and Kai Ren and Yongqiang Li and Jinhao Fan and Chuanping Hu},
  title   = {Unlocking gait semantics: A multimodal lifelong gait recognition framework with decoupled features and attribute-driven mixture of experts},
  journal = {Expert Systems with Applications},
  volume  = {299},
  pages   = {129983},
  year    = {2026},
  doi     = {10.1016/j.eswa.2025.129983},
  url     = {https://dblp.org/rec/journals/eswa/XiLRLFH26}
}

@inproceedings{xu2023visualpromptmoe,
  author    = {Bo Xu and Shizhou Huang and Ming Du and Hongya Wang and Hui Song and Yanghua Xiao and Xin Lin},
  title     = {A Unified Visual Prompt Tuning Framework with Mixture-of-Experts for Multimodal Information Extraction},
  booktitle = {Database Systems for Advanced Applications},
  pages     = {544--554},
  year      = {2023},
  doi       = {10.1007/978-3-031-30675-4_40},
  url       = {https://dblp.org/rec/conf/dasfaa/XuHDWSXL23}
}

@article{zhang2026prismocc,
  author  = {Yujia Zhang and Hui Zhu and Chen Hua and Xinkai Kuang and Ziyu Chen and Chunmao Jiang},
  title   = {{PRISM-Occ}: Path-Routed Integrated Sparse Mixture-of-Experts for Multi-Modal {BEV} Occupancy Prediction},
  journal = {IEEE Signal Processing Letters},
  volume  = {33},
  pages   = {381--385},
  year    = {2026},
  url     = {https://dblp.org/rec/journals/spl/ZhangZHKCJ26}
}

@inproceedings{zhu2025quality,
  title={A quality-guided mixture of score-fusion experts framework for human recognition},
  author={Zhu, Jie and Su, Yiyang and Kim, Minchul and Jain, Anil and Liu, Xiaoming},
  booktitle={2025 IEEE/CVF International Conference on Computer Vision (ICCV)},
  pages={13076--13086},
  year={2025},
  organization={IEEE}
}

@article{luo2025breastmome,
  author  = {Luyang Luo and Mingxiang Wu and Mei Li and Yi Xin and Qiong Wang and Varut Vardhanabhuti and Winnie C. W. Chu and Zhenhui Li and Juan Zhou and Pranav Rajpurkar and Hao Chen},
  title   = {A large model for non-invasive and personalized management of breast cancer from multiparametric {MRI}},
  journal = {Nature Communications},
  volume  = {16},
  pages   = {3647},
  year    = {2025},
  doi     = {10.1038/s41467-025-58798-z},
  url     = {https://doi.org/10.1038/s41467-025-58798-z}
}

@article{chen2026comedtr,
  author  = {Qishen Chen and Wenxuan He and Xingyuan Chen and Chen Cheng and Minjie Bian and Huahu Xu},
  title   = {Large-small model collaboration for medical visual question answering with task aware mixture of experts and relation knowledge distillation},
  journal = {Image and Vision Computing},
  volume  = {165},
  pages   = {105820},
  year    = {2026},
  doi     = {10.1016/j.imavis.2025.105820},
  url     = {https://dblp.org/rec/journals/ivc/ChenHCCBX26}
}

@article{hashimoto2024lymphoma,
  author  = {Noriaki Hashimoto and Hiroyuki Hanada and Hiroaki Miyoshi and Miharu Nagaishi and Kensaku Sato and Hidekata Hontani and Koichi Ohshima and Ichiro Takeuchi},
  title   = {Multimodal Gated Mixture of Experts Using Whole Slide Image and Flow Cytometry for Multiple Instance Learning Classification of Lymphoma},
  journal = {Journal of Pathology Informatics},
  volume  = {15},
  pages   = {100359},
  year    = {2024},
  doi     = {10.1016/j.jpi.2023.100359},
  url     = {https://doi.org/10.1016/j.jpi.2023.100359}
}

@article{saueressig2026glioma,
  author  = {Camillo Saueressig and Daniel Scholz and Philipp Raffler and Claire Delbridge and Benedikt Wiestler and Peter Sch{"u}ffler},
  title   = {Multimodal fusion of pathology and radiology foundation models for {WHO} 2021 glioma subtyping},
  journal = {npj Precision Oncology},
  volume  = {10},
  pages   = {118},
  year    = {2026},
  doi     = {10.1038/s41698-026-01366-5},
  url     = {https://doi.org/10.1038/s41698-026-01366-5}
}

@inproceedings{xiong2024mome,
  author    = {Conghao Xiong and Hao Chen and Hao Zheng and Dong Wei and Yefeng Zheng and Joseph J. Y. Sung and Irwin King},
  title     = {{MoME}: Mixture of Multimodal Experts for Cancer Survival Prediction},
  booktitle = {Medical Image Computing and Computer Assisted Intervention -- MICCAI 2024},
  pages     = {318--328},
  year      = {2024},
  url       = {https://dblp.org/rec/conf/miccai/XiongCZWZSK24}
}

@article{zhang2026surmoe,
  author  = {Wei Zhang and Wenxin Xu and Tong Chen and Collin Sakal and Xinyue Li},
  title   = {Integrating images and genomics for multi-modal cancer survival analysis via mixture of experts},
  journal = {Information Fusion},
  volume  = {126},
  pages   = {103521},
  year    = {2026},
  doi     = {10.1016/j.inffus.2025.103521},
  url     = {https://dblp.org/rec/journals/inffus/ZhangXCSL26}
}

@inproceedings{raza2025neuromoe,
  author    = {Wajih Hassan Raza and Aamir Bader Shah and Yu Wen and Yidan Shen and Juan Diego Martinez Lemus and Mya Caryn Schiess and Timothy M. Ellmore and Renjie Hu and Xin Fu},
  title     = {{NeuroMoE}: A Transformer-Based Mixture-of-Experts Framework for Multi-Modal Neurological Disorder Classification},
  booktitle = {47th Annual International Conference of the IEEE Engineering in Medicine and Biology Society},
  year      = {2025},
  doi       = {10.1109/EMBC58623.2025.11254303},
  url       = {https://dblp.org/rec/journals/corr/abs-2506-14970}
}

@article{minoura2021scmm,
  author  = {Kodai Minoura and Ko Abe and Hyunha Nam and Hiroyoshi Nishikawa and Teppei Shimamura},
  title   = {A mixture-of-experts deep generative model for integrated analysis of single-cell multiomics data},
  journal = {Cell Reports Methods},
  volume  = {1},
  number  = {5},
  pages   = {100071},
  year    = {2021},
  doi     = {10.1016/j.crmeth.2021.100071},
  url     = {https://doi.org/10.1016/j.crmeth.2021.100071}
}

@article{bi2026ringmoe,
  author  = {Hanbo Bi and Yingchao Feng and Boyuan Tong and Mengyu Wang and Haichen Yu and Yongqiang Mao and Hao Chang and Wenhui Diao and Peijin Wang and Yue Yu and Hanyang Peng and Yehong Zhang and Kun Fu and Xian Sun},
  title   = {{RingMoE}: Mixture-of-Modality-Experts Multi-Modal Foundation Models for Universal Remote Sensing Image Interpretation},
  journal = {IEEE Transactions on Pattern Analysis and Machine Intelligence},
  volume  = {48},
  number  = {4},
  pages   = {4388--4405},
  year    = {2026},
  doi     = {10.1109/TPAMI.2025.3643453},
  url     = {https://dblp.org/rec/journals/pami/BiFTWYMCDWYPZFS26}
}

@inproceedings{li2026sm3det,
  author    = {Yuxuan Li and Xiang Li and Yunheng Li and Yicheng Zhang and Yimian Dai and Qibin Hou and Ming-Ming Cheng and Jian Yang},
  title     = {{SM3Det}: A Unified Model for Multi-Modal Remote Sensing Object Detection},
  booktitle = {Proceedings of the Fortieth AAAI Conference on Artificial Intelligence},
  pages     = {6717--6725},
  year      = {2026},
  doi       = {10.1609/aaai.v40i8.37603},
  url       = {https://dblp.org/rec/conf/aaai/LiLLZDHCY26}
}

@article{lin2025rsmoe,
  author  = {Hui Lin and Danfeng Hong and Shuhang Ge and Chuyao Luo and Kai Jiang and Hao Jin and Congcong Wen},
  title   = {{RS-MoE}: A Vision--Language Model With Mixture of Experts for Remote Sensing Image Captioning and Visual Question Answering},
  journal = {IEEE Transactions on Geoscience and Remote Sensing},
  volume  = {63},
  pages   = {1--18},
  year    = {2025},
  doi     = {10.1109/TGRS.2025.3547988},
  url     = {https://dblp.org/rec/journals/tgrs/LinHGLJJW25}
}

@inproceedings{liu2026skymoe,
  author    = {Jiaqi Liu and Ronghao Fu and Lang Sun and Haoran Liu and Xiao Yang and Weipeng Zhang and Xu Na and Zhuoran Duan and Bo Yang},
  title     = {{SkyMoE}: A Vision-Language Foundation Model for Enhancing Geospatial Interpretation with Mixture of Experts},
  booktitle = {Proceedings of the Fortieth AAAI Conference on Artificial Intelligence},
  pages     = {7168--7178},
  year      = {2026},
  doi       = {10.1609/aaai.v40i9.37653},
  url       = {https://dblp.org/rec/conf/aaai/LiuFSLYZNDY26}
}

@article{kong2025dualmoe,
  author  = {Yi Kong and Shaocai Yu and Yuhu Cheng and C. L. Philip Chen and Xuesong Wang},
  title   = {Joint Classification of Hyperspectral Images and {LiDAR} Data Based on Candidate Pseudo Labels Pruning and Dual Mixture of Experts},
  journal = {IEEE Transactions on Geoscience and Remote Sensing},
  volume  = {63},
  pages   = {1--12},
  year    = {2025},
  doi     = {10.1109/TGRS.2025.3543498},
  url     = {https://dblp.org/rec/journals/tgrs/KongYCCW25}
}

@article{liu2025mixturers,
  author  = {Yimei Liu and Changyuan Wu and Minglei Guan and Jingzhe Wang},
  title   = {{MixtureRS}: A Mixture of Expert Network Based Remote Sensing Land Classification},
  journal = {Remote Sensing},
  volume  = {17},
  number  = {14},
  pages   = {2494},
  year    = {2025},
  doi     = {10.3390/rs17142494},
  url     = {https://doi.org/10.3390/rs17142494}
}

@article{qi2026multimoe,
  author  = {Ziyue Qi and Qirui Guo and Libo Wang and Yicong Zhou},
  title   = {{MultiMoE}: Multimodal Mixture of Experts Adapter for Semantic Segmentation of Remote Sensing Images},
  journal = {IEEE Transactions on Geoscience and Remote Sensing},
  volume  = {64},
  pages   = {5626013},
  year    = {2026},
  doi     = {10.1109/TGRS.2026.3702194},
  url     = {https://doi.org/10.1109/TGRS.2026.3702194}
}

@article{liu2025priorenhanced,
  author  = {Kewei Liu and Dongliang Peng and Tao Li},
  title   = {Multimodal Remote Sensing Object Detection Based on Prior-Enhanced Mixture-of-Experts Fusion Network},
  journal = {IEEE Transactions on Geoscience and Remote Sensing},
  volume  = {63},
  pages   = {1--14},
  year    = {2025},
  doi     = {10.1109/TGRS.2025.3585634},
  url     = {https://dblp.org/rec/journals/tgrs/LiuPL25}
}

@article{khankeshizadeh2026dmoe,
  author  = {Seyed Ehsan Khankeshizadeh and Ali Mohammadzadeh and Ali Jamali and Sadegh Jamali},
  title   = {A Dual-Modal Mixture-of-Experts Attention {U-Net} ({DMoE-AttU-Net}) for Change Detection Using Heterogeneous Optical and {SAR} Remote Sensing Images},
  journal = {Remote Sensing},
  volume  = {18},
  number  = {10},
  pages   = {1508},
  year    = {2026},
  doi     = {10.3390/rs18101508},
  url     = {https://doi.org/10.3390/rs18101508}
}

@article{shu2026uniroute,
  author  = {Qing-Ling Shu and Sibao Chen and Wei Lu and Zhihui You and Chengzhuang Liu},
  title   = {{UniRoute}: Unified Routing Mixture-of-Experts for Modality-Adaptive Remote Sensing Change Detection},
  journal = {CoRR},
  volume  = {abs/2601.14797},
  year    = {2026},
  doi     = {10.48550/arXiv.2601.14797},
  url     = {https://dblp.org/rec/journals/corr/abs-2601-14797}
}

@article{gao2025mamol,
  author  = {Qinghao Gao and Jiahui Qu and Yunsong Li and Wenqian Dong},
  title   = {Rethinking Efficient Mixture-of-Experts for Remote Sensing Modality-Missing Classification},
  journal = {CoRR},
  volume  = {abs/2511.11460},
  year    = {2025},
  doi     = {10.48550/arXiv.2511.11460},
  url     = {https://dblp.org/rec/journals/corr/abs-2511-11460}
}

@article{liu2026damoe,
  author  = {Chuang Liu and Jianhua Guo and Yingdong Pi and Xiao Wu and Zhiqi Zhang and Ru Chen and Xinyi Wang and Mi Wang},
  title   = {Prior-Guided Multi-Domain Mixture-of-Experts for Multimodal Earth Observation Data Gaps},
  journal = {ISPRS Journal of Photogrammetry and Remote Sensing},
  volume  = {239},
  pages   = {876--892},
  year    = {2026},
  doi     = {10.1016/j.isprsjprs.2026.06.034},
  url     = {https://doi.org/10.1016/j.isprsjprs.2026.06.034}
}

@inproceedings{li2024lite,
  author    = {Haoran Li and Junqi Liu and Zexian Wang and Shiyuan Luo and Xiaowei Jia and Huaxiu Yao},
  title     = {{LITE}: Modeling Environmental Ecosystems with Multimodal Large Language Models},
  booktitle = {Proceedings of the First Conference on Language Modeling},
  year      = {2024},
  url       = {https://openreview.net/forum?id=DRffhKBVlE}
}

@article{alizadehsani2026ecological,
  author  = {Zakieh Alizadehsani and Oliver Hensel and Abozar Nasirahmadi},
  title   = {Transforming Multimodal Data into Decision Assets: A Vision-Language Model for Ecological Data Using Cross-Modal Alignment},
  journal = {Environmental Modelling \& Software},
  pages   = {107157},
  year    = {2026},
  doi     = {10.1016/j.envsoft.2026.107157},
  url     = {https://doi.org/10.1016/j.envsoft.2026.107157}
}

@article{guo2026qwindustry,
  author  = {Xinyue Guo and Yi Li and Songyi Cui and Yuming Xu and Ray Y. Zhong},
  title   = {Knowledge-guided multimodal {LLM} with mixture-of-experts for interpretable defect detection and fault diagnosis: An industrial {AI} copilot},
  journal = {Journal of Manufacturing Systems},
  volume  = {85},
  pages   = {662--677},
  year    = {2026},
  doi     = {10.1016/j.jmsy.2026.02.017},
  url     = {https://doi.org/10.1016/j.jmsy.2026.02.017}
}

@article{li2026textguided,
  author  = {Zewen Li and Shuo Ye and Zitong Yu and Weicheng Xie and Linlin Shen},
  title   = {Text-Guided Multimodal Unified Industrial Anomaly Detection},
  journal = {CoRR},
  volume  = {abs/2604.22899},
  year    = {2026},
  doi     = {10.48550/arXiv.2604.22899},
  url     = {https://dblp.org/rec/journals/corr/abs-2604-22899}
}

@article{lin2026awmoe,
  author  = {Hongwei Lin and Xun Huang and Chenglu Wen and Cheng Wang},
  title   = {{AW-MoE}: All-Weather Mixture of Experts for Robust Multi-Modal 3D Object Detection},
  journal = {CoRR},
  volume  = {abs/2603.16261},
  year    = {2026},
  doi     = {10.48550/arXiv.2603.16261},
  url     = {https://dblp.org/rec/journals/corr/abs-2603-16261}
}

@inproceedings{yang2026drivemoe,
  author    = {Zhenjie Yang and Yilin Chai and Xiaosong Jia and Qifeng Li and Yuqian Shao and Xuekai Zhu and Haisheng Su and Junchi Yan},
  title     = {{DriveMoE}: Mixture-of-Experts for Vision-Language-Action Model in End-to-End Autonomous Driving},
  booktitle = {Proceedings of the IEEE/CVF Conference on Computer Vision and Pattern Recognition},
  pages     = {10678--10688},
  year      = {2026},
  url       = {https://openaccess.thecvf.com/content/CVPR2026/html/Yang_DriveMoE_Mixture-of-Experts_for_Vision-Language-Action_Model_in_End-to-End_Autonomous_Driving_CVPR_2026_paper.html}
}

@inproceedings{li2026forcevla2,
  author    = {Yang Li and Zhaxizhuoma and Hongru Jiang and Junjie Xia and Hongquan Zhang and Jinda Du and Yunsong Zhou and Jia Zeng and Ce Hao and Jieji Ren and Qiaojun Yu and Cewu Lu and Yu Qiao and Jiangmiao Pang},
  title     = {{ForceVLA2}: Unleashing Hybrid Force-Position Control with Force Awareness for Contact-Rich Manipulation},
  booktitle = {Proceedings of the IEEE/CVF Conference on Computer Vision and Pattern Recognition},
  pages     = {8911--8920},
  year      = {2026},
  url       = {https://openaccess.thecvf.com/content/CVPR2026/html/Li_ForceVLA2_Unleashing_Hybrid_Force-Position_Control_with_Force_Awareness_for_Contact-Rich_CVPR_2026_paper.html}
}

@article{tang2026modevla,
  author  = {Tutian Tang and Xingyu Ji and Wanli Xing and Ce Hao and Wenqiang Xu and Lin Shao and Cewu Lu and Qiaojun Yu and Jiangmiao Pang and Kaifeng Zhang},
  title   = {Towards Human-Like Manipulation through {RL}-Augmented Teleoperation and Mixture-of-Dexterous-Experts {VLA}},
  journal = {CoRR},
  volume  = {abs/2603.08122},
  year    = {2026},
  doi     = {10.48550/arXiv.2603.08122},
  url     = {https://dblp.org/rec/journals/corr/abs-2603-08122}
}

@inproceedings{cappellazzo2025momeavsr,
  author    = {Umberto Cappellazzo and Minsu Kim and Pingchuan Ma and Honglie Chen and Xubo Liu and Stavros Petridis and Maja Pantic},
  title     = {{MoME}: Mixture of Matryoshka Experts for Audio-Visual Speech Recognition},
  booktitle = {Advances in Neural Information Processing Systems},
  volume    = {38},
  year      = {2025},
  url       = {https://proceedings.neurips.cc/paper_files/paper/2025/hash/392aae924264f2c56d1895b232bb46b6-Abstract-Conference.html}
}

@article{dutta2026mixture,
  author  = {Soumya Dutta and Smruthi Balaji and Sriram Ganapathy},
  title   = {A Mixture-of-Experts Model for Multimodal Emotion Recognition in Conversations},
  journal = {Computer Speech \& Language},
  volume  = {100},
  pages   = {101965},
  year    = {2026},
  doi     = {10.1016/j.csl.2026.101965},
  url     = {https://dblp.org/rec/journals/csl/DuttaBG26}
}

@article{huang2025pamoe,
  author  = {Changqin Huang and Zhenheng Lin and Zhongmei Han and Qionghao Huang and Fan Jiang and Xiaodi Huang},
  title   = {{PAMoE-MSA}: Polarity-aware mixture of experts network for multimodal sentiment analysis},
  journal = {International Journal of Multimedia Information Retrieval},
  volume  = {14},
  number  = {1},
  pages   = {7},
  year    = {2025},
  doi     = {10.1007/s13735-025-00362-y},
  url     = {https://dblp.org/rec/journals/ijmir/HuangLHHJH25}
}

@article{wang2026clipmoe,
  author  = {Jun Wang and Fengjun Zhou and Xueqiang Gao and Hong Wang and Zhiquan Feng and Maozhen Li},
  title   = {Fine-tuning {CLIP} with mixture of experts and cross-modal alignment via contrastive learning for multimodal sentiment analysis},
  journal = {Neurocomputing},
  volume  = {670},
  pages   = {132417},
  year    = {2026},
  doi     = {10.1016/j.neucom.2025.132417},
  url     = {https://dblp.org/rec/journals/ijon/WangZGWFL26}
}

@inproceedings{shou2026c2moe,
  author    = {Yuntao Shou and Tao Meng and Wei Ai and Keqin Li},
  title     = {{C$^{2}$MOE}: Consistency and Complementarity-guided Mixture of Experts for Incomplete Multimodal Emotion Learning},
  booktitle = {Proceedings of the 2026 International Conference on Multimedia Retrieval},
  pages     = {1055--1063},
  year      = {2026},
  doi       = {10.1145/3805622.3810726},
  url       = {https://dblp.org/rec/conf/mir/ShouMAL26}
}

@inproceedings{hu2025multilevel,
  author    = {Zhiwei Hu and V{\'i}ctor Guti{\'e}rrez-Basulto and Zhiliang Xiang and Ru Li and Jeff Z. Pan},
  title     = {Multi-level Mixture of Experts for Multimodal Entity Linking},
  booktitle = {Proceedings of the 31st ACM SIGKDD Conference on Knowledge Discovery and Data Mining},
  pages     = {979--990},
  year      = {2025},
  doi       = {10.1145/3711896.3737060},
  url       = {https://dblp.org/rec/conf/kdd/HuGX0P25}
}

@inproceedings{liu2025mimoefnd,
  author    = {Yifan Liu and Yaokun Liu and Zelin Li and Ruichen Yao and Yang Zhang and Dong Wang},
  title     = {Modality Interactive Mixture-of-Experts for Fake News Detection},
  booktitle = {Proceedings of the ACM Web Conference 2025},
  pages     = {5139--5150},
  year      = {2025},
  doi       = {10.1145/3696410.3714522},
  url       = {https://dblp.org/rec/conf/www/LiuLLYZ025}
}

@inproceedings{zhang2025hm4sr,
  author    = {Shengzhe Zhang and Liyi Chen and Dazhong Shen and Chao Wang and Hui Xiong},
  title     = {Hierarchical Time-Aware Mixture of Experts for Multi-Modal Sequential Recommendation},
  booktitle = {Proceedings of the ACM Web Conference 2025},
  pages     = {3672--3682},
  year      = {2025},
  doi       = {10.1145/3696410.3714676},
  url       = {https://dblp.org/rec/conf/www/Zhang0S0025}
}

@article{zhang2025docrouter,
  author  = {Jinxu Zhang and Yu Zhang},
  title   = {{DocRouter}: Prompt guided vision transformer and Mixture of Experts connector for document understanding},
  journal = {Information Fusion},
  volume  = {122},
  pages   = {103206},
  year    = {2025},
  doi     = {10.1016/j.inffus.2025.103206},
  url     = {https://doi.org/10.1016/j.inffus.2025.103206}
}

@article{zhang2026isac,
  author  = {Kai Zhang and Wentao Yu and Hengtao He and Shenghui Song and Jun Zhang and Khaled B. Letaief},
  title   = {Multimodal Mixture-of-Experts for {ISAC} in Low-Altitude Wireless Networks},
  journal = {IEEE Journal on Selected Topics in Signal Processing},
  pages   = {1--17},
  year    = {2026},
  doi     = {10.1109/JSTSP.2026.3696558},
  url     = {https://dblp.org/rec/journals/corr/abs-2512-01750}
}

@article{lin2026timi,
  author  = {Jiafeng Lin and Yuxuan Wang and Huakun Luo and Zhongyi Pei and Jianmin Wang},
  title   = {{TiMi}: Empower Time Series Transformers with Multimodal Mixture of Experts},
  journal = {CoRR},
  volume  = {abs/2602.21693},
  year    = {2026},
  doi     = {10.48550/arXiv.2602.21693},
  url     = {https://dblp.org/rec/journals/corr/abs-2602-21693},
  note    = {Preprint}
}

@article{zhang2026modulated,
  author  = {Lige Zhang and Ali Maatouk and Jialin Chen and Leandros Tassiulas and Rex Ying},
  title   = {Multi-Modal Time Series Prediction via Mixture of Modulated Experts},
  journal = {CoRR},
  volume  = {abs/2601.21547},
  year    = {2026},
  doi     = {10.48550/arXiv.2601.21547},
  url     = {https://dblp.org/rec/journals/corr/abs-2601-21547},
  note    = {Preprint}
}

@article{liu2025llmoetrading,
  author  = {Kuan{-}Ming Liu and Ming{-}Chih Lo},
  title   = {{LLM}-Based Routing in Mixture of Experts: A Novel Framework for Trading},
  journal = {CoRR},
  volume  = {abs/2501.09636},
  year    = {2025},
  doi     = {10.48550/arXiv.2501.09636},
  url     = {https://dblp.org/rec/journals/corr/abs-2501-09636},
  note    = {Preprint}
}

@article{soares2025molmoe,
  author  = {Eduardo E. M. Soares and Victor Yukio Shirasuna and Emilio Vital Brazil and Indra Priyadarsini and Seiji Takeda},
  title   = {Multi-view mixture-of-experts for predicting molecular properties using {SMILES}, {SELFIES}, and graph-based representations},
  journal = {Machine Learning: Science and Technology},
  volume  = {6},
  number  = {2},
  pages   = {025070},
  year    = {2025},
  doi     = {10.1088/2632-2153/ade4ef},
  url     = {https://dblp.org/rec/journals/mlst/SoaresSBPT25}
}

@inproceedings{gurban2008dynamic,
  author    = {Mihai Gurban and Jean-Philippe Thiran and Thomas Drugman and Thierry Dutoit},
  title     = {Dynamic Modality Weighting for Multi-Stream HMMs in Audio-Visual Speech Recognition},
  booktitle = {Proceedings of the 10th International Conference on Multimodal Interfaces},
  year      = {2008}
}

@article{dehghanghanatkaman2026dynamic,
  author  = {Amin Dehghanghanatkaman},
  title   = {Dynamic Context-Aware Multi-Modal Deep Learning for Longitudinal Prediction of Parkinson's Disease Progression},
  journal = {Scientific Reports},
  volume  = {16},
  pages   = {2104},
  year    = {2026}
}

@misc{zhang2026regularizing,
  author        = {Zhen Zhang and Jielei Chu and Bin Liu and Tianrui Li},
  title         = {Regularizing Modality Contribution Drift in Multimodal Continual Learning},
  year          = {2026},
  eprint        = {2607.27260},
  archivePrefix = {arXiv},
  primaryClass  = {cs.LG}
}

@InProceedings{sun2025flexireid,
  title     = {{FlexiReID}: Adaptive Mixture of Expert for Multi-Modal Person Re-Identification},
  author    = {Sun, Zhen and Tan, Lei and Shen, Yunhang and Cai, Chengmao
               and Sun, Xing and Dai, Pingyang and Cao, Liujuan and Ji, Rongrong},
  booktitle = {Proceedings of the 42nd International Conference on Machine Learning},
  series    = {Proceedings of Machine Learning Research},
  volume    = {267},
  pages     = {57680--57693},
  publisher = {PMLR},
  year      = {2025},
  url       = {https://proceedings.mlr.press/v267/sun25q.html}
}

@InProceedings{feng2025mfrnet,
  title     = {Multi-Modal Object Re-identification via Sparse Mixture-of-Experts},
  author    = {Feng, Yingying and Li, Jie and Xie, Chi and Tan, Lei and Jiayi, Ji},
  booktitle = {Proceedings of the 42nd International Conference on Machine Learning},
  series    = {Proceedings of Machine Learning Research},
  volume    = {267},
  pages     = {16796--16807},
  publisher = {PMLR},
  year      = {2025},
  url       = {https://proceedings.mlr.press/v267/feng25i.html}
}

@article{wang2025demo,
  title   = {{DeMo}: Decoupled Feature-Based Mixture of Experts for Multi-Modal Object Re-Identification},
  author  = {Wang, Yuhao and Liu, Yang and Zheng, Aihua and Zhang, Pingping},
  journal = {Proceedings of the AAAI Conference on Artificial Intelligence},
  volume  = {39},
  number  = {8},
  pages   = {8141--8149},
  year    = {2025},
  doi     = {10.1609/aaai.v39i8.32878}
}

@article{ren2025mosce,
  title   = {{MoSCE-ReID}: Mixture of Semantic Clustering Experts for Person Re-Identification},
  author  = {Ren, Kai and Hu, Chuanping and Xi, Hao and Li, Yongqiang
             and Fan, Jinhao and Liu, Lihua},
  journal = {Neurocomputing},
  volume  = {626},
  pages   = {129587},
  year    = {2025},
  doi     = {10.1016/j.neucom.2025.129587}
}

@InProceedings{hong2026dualmoe,
  title     = {Unified Multimodal Visual Tracking with Dual Mixture-of-Experts},
  author    = {Hong, Lingyi and Li, Jinglun and Zhou, Xinyu and Jiang, Kaixun
               and Guo, Pinxue and Chen, Zhaoyu and Li, Runze
               and Sheng, Xingdong and Zhang, Wenqiang},
  booktitle = {Proceedings of the 43rd International Conference on Machine Learning},
  series    = {Proceedings of Machine Learning Research},
  volume    = {306},
  pages     = {43649--43661},
  publisher = {PMLR},
  year      = {2026},
  url       = {https://proceedings.mlr.press/v306/hong26g.html}
}

@article{zhu2023amc,
  title   = {{AMC}: Adaptive Multi-expert Collaborative Network for Text-guided Image Retrieval},
  author  = {Zhu, Hongguang and Wei, Yunchao and Zhao, Yao and Zhang, Chunjie and Huang, Shujuan},
  journal = {ACM Transactions on Multimedia Computing, Communications, and Applications},
  volume  = {19},
  number  = {6},
  pages   = {1--22},
  year    = {2023},
  doi     = {10.1145/3584703}
}

@article{zhai2025mmes,
  title   = {Composed Image Retrieval by Multimodal Mixture-of-Expert Synergy},
  author  = {Zhai, Wenzhe and Gao, Mingliang and Jeon, Gwanggil and Zhou, Qiang and Camacho, David},
  journal = {Image and Vision Computing},
  volume  = {161},
  pages   = {105634},
  year    = {2025},
  doi     = {10.1016/j.imavis.2025.105634}
}

@article{li2026damoecrossmodal,
  title   = {Discrepancy-aware Contrastive Learning with Mixture of Experts for Cross-modal Image-text Semantic Alignment},
  author  = {Li, Xin and He, Xin and Tang, Gaigai and Qi, Kaiyuan and Su, Guangfeng and Zhang, Huiyun},
  journal = {Neurocomputing},
  volume  = {660},
  pages   = {131887},
  year    = {2026},
  doi     = {10.1016/j.neucom.2025.131887}
}

@inproceedings{zhang2026adafuse,
  title     = {{ADaFuSE}: Adaptive Diffusion-generated Image and Text Fusion for Interactive Text-to-Image Retrieval},
  author    = {Zhang, Zhuocheng and Zhang, Xingwu and Liang, Kangheng and Li, Guanxuan and McCreadie, Richard and Long, Zijun},
  booktitle = {Proceedings of the 49th International ACM SIGIR Conference on Research and Development in Information Retrieval},
  pages     = {4349--4354},
  year      = {2026},
  publisher = {ACM},
  doi       = {10.1145/3805712.3809836}
}

@inproceedings{peng2026retrievalexperts,
  title     = {Mixture-of-Retrieval Experts for Reasoning-Guided Multimodal Knowledge Exploitation},
  author    = {Peng, Chunyi and Xu, Zhipeng and Liu, Zhenghao and Li, Yishan and Yan, Yukun and Wang, Shuo and Gu, Yu and Yu, Minghe and Yu, Ge and Sun, Maosong},
  booktitle = {Proceedings of the 49th International ACM SIGIR Conference on Research and Development in Information Retrieval},
  pages     = {1440--1450},
  year      = {2026},
  publisher = {ACM},
  doi       = {10.1145/3805712.3809588}
}

@inproceedings{wang2022multimodalretrieval,
  title     = {Named Entity and Relation Extraction with Multi-Modal Retrieval},
  author    = {Wang, Xinyu and Cai, Jiong and Jiang, Yong and Xie, Pengjun and Tu, Kewei and Lu, Wei},
  booktitle = {Findings of the Association for Computational Linguistics: EMNLP 2022},
  pages     = {5925--5936},
  year      = {2022},
  publisher = {Association for Computational Linguistics},
  doi       = {10.18653/v1/2022.findings-emnlp.437},
  url       = {https://aclanthology.org/2022.findings-emnlp.437/}
}

@inproceedings{bian2023m3srec,
  title     = {Multi-modal Mixture of Experts Represetation Learning for Sequential Recommendation},
  author    = {Bian, Shuqing and Pan, Xingyu and Zhao, Wayne Xin and Wang, Jinpeng and Wang, Chuyuan and Wen, Ji-Rong},
  booktitle = {Proceedings of the 32nd ACM International Conference on Information and Knowledge Management},
  pages     = {110--119},
  year      = {2023},
  publisher = {ACM},
  doi       = {10.1145/3583780.3614978}
}

@inproceedings{qu2025xsmoe,
  title     = {Efficient Multimodal Streaming Recommendation via Expandable Side Mixture-of-Experts},
  author    = {Qu, Yunke and Qu, Liang and Chen, Tong and Nguyen, Quoc Viet Hung and Yin, Hongzhi},
  booktitle = {Proceedings of the 34th ACM International Conference on Information and Knowledge Management},
  pages     = {2460--2470},
  year      = {2025},
  publisher = {ACM},
  doi       = {10.1145/3746252.3761390}
}

@inproceedings{guo2025m3rec,
  title     = {$M^{3}$Rec: Selective State Space Models with Mixture-of-Modality Experts for Multi-Modal Sequential Recommendation},
  author    = {Guo, Xu and Zhang, Tong and Xue, Yufei and Wang, Chenxu and Wang, Fuyun and Cui, Zhen},
  booktitle = {Proceedings of the IEEE International Conference on Acoustics, Speech and Signal Processing},
  pages     = {1--5},
  year      = {2025},
  doi       = {10.1109/ICASSP49660.2025.10887582}
}

@article{zheng2026mamoe4rec,
  title   = {{MaMoE4Rec}: Multimodal Recommendation with Hop-Aware Graph Modeling and Mixture-of-Experts Fusion},
  author  = {Zheng, Sirui and Liu, Jin and Huang, Bo and Tang, Yongqiang and You, Lan and Fujita, Hamido},
  journal = {Expert Systems with Applications},
  volume  = {316},
  pages   = {131852},
  year    = {2026},
  doi     = {10.1016/j.eswa.2026.131852}
}

@article{xu2026cammsr,
  title   = {{CAMMSR}: Category-Guided Attentive Mixture of Experts for Multimodal Sequential Recommendation},
  author  = {Xu, Jinfeng and Chen, Zheyu and Yang, Shuo and Li, Jinze and Wang, Hewei and Li, Yijie and Tang, Jianheng and Liu, Yunhuai and Ngai, Edith C. H.},
  journal = {CoRR},
  volume  = {abs/2603.04320},
  year    = {2026},
  eprint  = {2603.04320},
  archivePrefix = {arXiv}
}

@article{liu2025casemcom,
  author  = {Guangyuan Liu and Yijing Liu and Jiacheng Wang and Hongyang Du and
             Dusit Niyato and Jiawen Kang and Zehui Xiong and Abbas Jamalipour},
  title   = {Context-Aware Semantic Communication for the Wireless Networks},
  journal = {IEEE Wireless Communications},
  volume  = {32},
  number  = {5},
  pages   = {20--28},
  year    = {2025},
  doi     = {10.1109/MWC.001.2500043},
  url     = {https://dblp.org/rec/journals/wc/LiuLWDNKXJ25}
}

@article{li2026cogmoe,
  author  = {Xingpeng Li and Enwen Hu and Siyuan Jin and Baoding Zhou and
             Jingrong Liu},
  title   = {{CoGMoE}: Sparse and Specialized Framework for Multi-Agent
             Collaborative Perception via Graph Mixture-of-Experts},
  journal = {Knowledge-Based Systems},
  volume  = {337},
  pages   = {115329},
  year    = {2026},
  doi     = {10.1016/j.knosys.2026.115329},
  url     = {https://dblp.org/rec/journals/kbs/LiHJZL26}
}

@inproceedings{koval2025mseitt,
  author    = {Ross Koval and Nicholas Andrews and Xifeng Yan},
  title     = {Multimodal Language Models for Financial Forecasting from
               Interleaved Sequences of Text and Time Series},
  booktitle = {Proceedings of the 14th International Joint Conference on
               Natural Language Processing and the 4th Conference of the
               Asia-Pacific Chapter of the Association for Computational
               Linguistics},
  pages     = {987--1001},
  year      = {2025},
  doi       = {10.18653/v1/2025.ijcnlp-long.54},
  url       = {https://aclanthology.org/2025.ijcnlp-long.54/}
}

@article{erni2026solar,
  author  = {Sarvarbek Erniyazov and Mohammed Abdul Jaleel and
             Chang Gyoon Lim and Seung Bum Ha},
  title   = {A Multimodal Fusion Framework for Solar Forecasting using
             Dynamic {GNNs} and Mixture of Experts},
  journal = {KSII Transactions on Internet and Information Systems},
  volume  = {20},
  number  = {3},
  pages   = {1337--1360},
  year    = {2026},
  doi     = {10.3837/tiis.2026.03.012},
  url     = {https://dblp.org/rec/journals/itiis/ErniyazovJLH26}
}

@article{wang2026mmpg,
  author  = {Yusong Wang and Jialun Shen and Zhihao Wu and Yicheng Xu and
             Shiyin Tan and Mingkun Xu and Changshuo Wang and Zixing Song
             and Prayag Tiwari},
  title   = {{MMPG}: {MoE}-Based Adaptive Multi-Perspective Graph Fusion
             for Protein Representation Learning},
  journal = {Proceedings of the AAAI Conference on Artificial Intelligence},
  volume  = {40},
  number  = {2},
  pages   = {1240--1248},
  year    = {2026},
  doi     = {10.1609/aaai.v40i2.37096},
  url     = {https://dblp.org/rec/conf/aaai/WangSWXTXWST26}
}

@article{zeng2026esslmmoe,
  author  = {Min Zeng and Xu Liu and Yiming Li and Qianpei Liu and Wenkang Wang
             and Fuhao Zhang and Fei Guo and Min Li},
  title   = {{EssLM-MoE}: A Mixture-of-Experts-Enhanced Framework for Protein
             Essentiality Prediction Using Fused Protein Language Models},
  journal = {Neurocomputing},
  volume  = {699},
  pages   = {134404},
  year    = {2026},
  doi     = {10.1016/j.neucom.2026.134404},
  url     = {https://dblp.org/rec/journals/ijon/ZengLLLWZGL26}
}

@article{wu2026mera,
  author  = {Jiayang Wu and Jiale Zhou and Rubo Wang and Xingyi Zhang and
             Xun Lin and Tianxu Lv and Leong Hou U and Yefeng Zheng},
  title   = {Multimodal Mixture-of-Experts with Retrieval Augmentation for
             Protein Active Site Identification},
  journal = {Proceedings of the AAAI Conference on Artificial Intelligence},
  volume  = {40},
  number  = {32},
  pages   = {26913--26921},
  year    = {2026},
  doi     = {10.1609/aaai.v40i32.39903},
  url     = {https://dblp.org/rec/conf/aaai/WuZWZLLUZ26}
}

@article{dai2026mdmdta,
  author  = {Yang Dai and Xiaoyu Tan and Haoyu Wang and Gengchen Ma and
             Yujie Xiong and Xihe Qiu},
  title   = {{MDM-DTA}: Message Passing Neural Network with Molecular
             Descriptors and Mixture of Experts for Drug-Target Affinity
             Prediction},
  journal = {Computer Methods and Programs in Biomedicine},
  volume  = {274},
  pages   = {109163},
  year    = {2026},
  doi     = {10.1016/j.cmpb.2025.109163},
  url     = {https://dblp.org/rec/journals/cmpb/DaiTWMXQ26}
}

@article{kang2025mfdta,
  author  = {Yanlei Kang and Haoyu Zhuang and Yunliang Jiang and Zhong Li},
  title   = {{MF-DTA}: Predicting Drug-Target Affinity with Multi-Modal
             Feature Fusion Model},
  journal = {Journal of Biomedical Informatics},
  volume  = {171},
  pages   = {104926},
  year    = {2025},
  doi     = {10.1016/j.jbi.2025.104926},
  url     = {https://dblp.org/rec/journals/jbi/KangZJL25}
}

\end{document}